\documentclass{article}

\usepackage[numbers,sort&compress]{natbib}

\usepackage[preprint]{neurips_2021}

\usepackage[breaklinks,colorlinks,
linkcolor=red,
anchorcolor=blue,
citecolor=blue
]{hyperref}
\usepackage[utf8]{inputenc} 
\usepackage[T1]{fontenc}    
\usepackage{url}            
\usepackage{booktabs}       
\usepackage{amsfonts}       
\usepackage{nicefrac}       
\usepackage{microtype}      
\usepackage{graphicx}
\graphicspath{{figures/}}
\usepackage{longfbox}
\usepackage{amsmath}
\usepackage{amsfonts}
\usepackage{amssymb}
\usepackage{amsmath}   
\usepackage{mathtools}
\usepackage{amsthm}
\usepackage[algo2e,algoruled,boxed,lined,norelsize]{algorithm2e} 
\usepackage{subcaption}
\usepackage{wrapfig}
\usepackage{multirow}
\usepackage{makecell}
\usepackage{parnotes}
\usepackage{float}
\usepackage[makeroom]{cancel}
\usepackage{chngcntr}
\usepackage{apptools}
\usepackage{enumitem}
\AtAppendix{\counterwithin{lemma}{section}}

\makeatletter
\newcommand{\mathleft}{\@fleqntrue\@mathmargin\parindent}
\newcommand{\mathcenter}{\@fleqnfalse\@mathmargin\parindent}

\makeatother
\DeclarePairedDelimiter{\norm}{\lVert}{\rVert}
 
\DeclareMathOperator{\Tr}{Tr}
\DeclareMathOperator{\col}{col}

\newtheorem{theorem}{Theorem}

\newtheorem{lemma}{Lemma}

\title{Safe Pontryagin Differentiable Programming}

\author{
	Wanxin Jin\\
	University of Pennsylvania\\
	\texttt{wanxinjin@gmail.com}
	\And
	Shaoshuai Mou \\
	Purdue University\\
	\texttt{mous@purdue.edu}
	\And
	George J. Pappas \\
	University of Pennsylvania\\
	\texttt{pappasg@seas.upenn.edu} 
}

\begin{document}

\maketitle
	
\vspace{-10pt}
\begin{abstract}
	
We propose a Safe Pontryagin Differentiable Programming (Safe PDP) methodology, which establishes a theoretical and algorithmic  framework to solve a broad class of safety-critical learning and control tasks---problems that require the guarantee of safety constraint satisfaction at any stage of the learning and control progress.  In the spirit of interior-point methods,   Safe PDP handles different types of system  constraints on states and inputs by incorporating them into the cost or loss through barrier functions. We prove three fundamentals  of the proposed  Safe PDP:  first, both the  solution and its gradient in the backward pass can be approximated by solving their  more efficient unconstrained counterparts;  second,   the approximation for both the  solution and its gradient can be controlled for arbitrary accuracy by a  barrier parameter;   and third,   importantly, all intermediate results throughout the approximation and optimization  strictly respect the  constraints,  thus guaranteeing safety throughout the entire learning and control process. We demonstrate the capabilities of   Safe PDP in solving various safety-critical tasks,  including safe policy optimization, safe motion planning, and learning MPCs from demonstrations, on different challenging systems such as 6-DoF maneuvering quadrotor and 6-DoF rocket powered landing.
	

\end{abstract}

\vspace{-10pt}

\section{Introduction}
\vspace{-5pt}

Safety is usually a  priority in the deployment of a learning or control algorithm to real-world systems. For a physical system (agent), safety is normally given in various constraints on system states and inputs, which must not be violated by the algorithm at \emph{any stage} of the learning and control process, otherwise will cause irrevocable or unacceptable failure/damage. Those systems are  referred to as \emph{safety-critical}. The constraints in  a safety-critical system can  include the {immediate} ones, which are directly  imposed on  the system state and input  at certain or all-time instances, and the {long-term} ones, which are defined on the  trajectory of system states and inputs over a long period.

Compared to the abundant results that  focus on   system optimality  \cite{mnih2015human,silver2016mastering,lillicrap2015continuous},  systematic and principled treatments for safety-critical learning and control problems  seem largely insufficient, particularly in the following  gaps (detailed in Section \ref{relatedwork}). First, existing safety strategies are either too conservative, which may restrict the  task performance, or violation-tolerable, which only pursues the near-constraint guarantee and thus are not strictly  constraint-respecting. Second, a systematic safety paradigm   capable  of handling different types of  constraints,   including system state and input (or mixed), immediate, or/and long-term constraints, is still lacked. Third,  some  existing safety strategies suffer from huge  computational- and data- complexity,  difficult to be  integrated into any differentiable programming frameworks to solve  large-scale  learning and continuous control tasks.

To address the above research gaps, this paper aims to develop a  \emph{safe differentiable programming} framework  with the following  key  capabilities.  First, the framework provides a systematic treatment for \emph{different types of  constraints} in a safety-critical problem; second, it attains \emph{provable safety- and accuracy- guarantees} throughout the  learning and control process;  third, it is flexible   to perform \emph{safe learning} of any unknown aspects of a constrained decision-making system, including  policy, dynamics, state and input constraints, and control cost;  finally, it can be integrated to any \emph{differentiable programming framework} to efficiently solve large-scale safe learning and control tasks.

\vspace{-5pt}

\subsection{Related Work} \label{relatedwork}
\vspace{-8pt}

In  machine learning and control fields,  safety has been defined  by different criteria, such as worst-case \cite{nilim2005robust,heger1994consideration}, risk-sensitive \cite{howard1972risk},  ergodicity \cite{moldovan2012safe},  robust \cite{madry2017towards,zhou1998essentials}, etc., most of which  are formulated by directly altering an objective function \cite{garcia2015comprehensive}. In this paper, we  focus only on constrained learning and control problems, where constraints are \emph{explicitly formulated and must be satisfied}. We categorize existing techniques into   \emph{at-convergence safety} methods,  which only concern    constraint satisfaction at  convergence, or  \emph{in-progress safety} methods, which attempt to ensure  constraint satisfaction during the entire optimization process.

\textbf{At-convergence safety methods}.  \,\,  In   reinforcement learning (RL), a constrained agent is typically formulated as a Constrained Markov Decision Process (CMDP) \cite{altman1999constrained}, seeking a policy that  not only optimizes a  reward but also satisfies an upper bound for a  cumulative cost. A  common  strategy \cite{altman1998constrained,yu2019convergent,chow2017risk,bhatnagar2012online,ding2020natural,calvo2021state} to solve CMDPs is to use the primal-dual method, by establishing the unconstrained Lagrangian and performing  saddle-point optimization. In deep learning, the primal-dual method has been recently used  \cite{nandwani2019primal} to train deep neural networks with constraints. In  control,  the primal-dual method has  been used to solve constrained optimal control (constrained trajectory) problems \cite{bergounioux1999primal,kirchner2018primal,howell2019altro}.
While proved to satisfy constraints  at  convergence \cite{du2019linear,jin2020local}, the  primal-dual type methods  cannot  guarantee   constraint satisfaction  during optimization, as  shown in  \cite{chow2019lyapunov,yu2019convergent}, thus are not suitable for safety-critical tasks.

\textbf{In-progress safety methods.} \,\, To  enforce  safety during  training,  \cite{achiam2017constrained} and     \cite{yang2020projection} solve CMDPs by introducing additional constraints into the Trust Region Policy Optimization (TRPO) \cite{schulman2015trust}. Since these methods  only obtain the `near constraint' guarantee,     constraint violation is not fully eliminated. Another line of constrained RL \cite{chow2018lyapunov,chow2019lyapunov, perkins2002lyapunov,berkenkamp2017safe}  leverages  the  Lyapunov theory \cite{lyapunov1992general} to bound  behavior of an  agent.  But how to choose a valid Lyapunov function for general tasks is still an open problem to date \cite{giesl2015review}, particularly for  constrained RL, since it requires a Lyapunov function  to  be  consistent with the   constraints and  to permit   optimal policies \cite{chow2018lyapunov}. Some other work also attempts to  handle immediate  constraints --- the constraints imposed on agent state and input   at any time. In \cite{turchetta2016safe},   a safe exploration scheme is proposed  to produce a safe reachable region; and it only considers finite state space.    \cite{wachi2020safe}  develops a  method that  learns safety constraints and then optimizes a reward within  the certified safe region; the method defines  constraints  purely on agent state and thus may not be readily applicable to mixed state-input   constraints.


In control, in-progress safety can be  achieved via two model-based frameworks:  reachability theory \cite{bansal2017hamilton} and control barrier functions \cite{wieland2007constructive,ames2016control}.   Safe control based on   reachability theory \cite{bansal2017hamilton,fisac2018general,herbert2017fastrack,chen2015safe}  explicitly considers  adversarial factors and seeks   a strategy that maintains the  constraints despite the  adversarial factors. This process typically requires solving the Hamilton-Jacobi-Isaacs equations \cite{evans1984differential}, which  become computationally difficult for high-dimensional systems \cite{bansal2017hamilton}.   Control barrier functions  \cite{wieland2007constructive,ames2016control}  constrain  a system only on  safety boundaries, making it  a less-conservative  strategy for  safety-critical  tasks \cite{ames2019control,choi2020reinforcement,cheng2019end}. Most of the methods  consider  affine dynamics  and directly use the given constraint function  as a  control barrier function. Such a choice could be problematic when a system  is uncontrollable at the boundary of the sublevel set. Thus, how to find a valid control barrier function is still  an ongoing research topic \cite{robey2020learning,robey2021learning, jin2020neural}. The above two control safety frameworks  favorably focus on  pure state constraints  and cannot be readily  extended to other constraints, such as  mixed state-input constraints or the  cumulative constraints defined on the system trajectory.

\textbf{Interior-point methods and control.} \,\, Interior-point methods (IPMs) \cite{fiacco1990nonlinear,nesterov1994interior,forsgren2002interior} solve constrained optimization  by sequentially finding  solutions to  
 unconstrained problems with the objective combining the original objective and a  barrier that prevents from leaving the feasible regions. IPMs have been used for \emph{constrained}  linear quadratic regular (LQR) control in  \cite{lim1996linearly,wright1991structured,wright1993interior,rao1998application,hansson1998robust,hansson2000primal, feller2016relaxed}. While IPMs for nonlinear constrained optimal control  are studied in \cite{laurent2007interior,hauser2006barrier,malisani2016interior,domahidi2012efficient,pavlov2021interior}, they mostly focus on developing  algorithms to solve the unconstrained approximation (from the perturbed KKT conditions) and lack of performance analysis. Most recently, \cite{usmanova2020safe} uses the IPM  to develop a  zero-th order non-convex optimization method; and \cite{liu2020ipo}  uses IPMs to solve reinforcement learning  with only cumulative constraints. Despite the promise of the trend,  the theoretical results and systematic algorithms regarding the \emph{differentiability of  general constrained   control  systems based on IPMs}  have not been    studied and established.

\textbf{Differentiable projection layer.} \,\, In machine learning, a recent line of  work considers embedding a differentiable projection layer  \cite{donti2020enforcing,chen2021enforcing,pham2018optlayer} into a general training process to ensure safety. Particularly,  \cite{pham2018optlayer} and \cite{chen2021enforcing} enforce safety by constructing a dedicated projection layer, which projects the unsafe actions outputted from a neural policy into  a safe region (satisfying safety constraints). This projection layer is a differentiable convex  layer \cite{amos2018differentiable,amos2017optnet}, which can be trained end-to-end. In \cite{donti2020enforcing}, safety is defined as robustness in the case of the worst adversarial disturbance, and  the set of robust policies is solved by classic robust control  (solving LMIs). An RL neural policy with a differentiable projection layer is learned such that the action from the neural policy lies in the robust policy set. Different from the above work,
Safe PDP does not enforce safety by  projection; instead, Safe PDP uses \emph{barrier functions} to guarantee safety constraint satisfaction. More importantly, we have shown,  in both theory and experiments, that \emph{with barrier functions,  differentiability can also be attained}.

\textbf{Sensitivity analysis and differentiable MPCs.} \,\,
Other  work related to Safe PDP includes the recent  results  for sensitivity analysis \cite{andersson2018sensitivity}, which focuses on  differentiation of a solution to a general nonlinear program, and differentiable MPCs \cite{amos2018differentiable}, which is based on differentiable quadratic programming. In long-horizon control settings, directly applying \cite{andersson2018sensitivity}
and \cite{amos2018differentiable} can be inefficient: the complexity of \cite{andersson2018sensitivity, amos2018differentiable} for  differentiating a solution to a general optimal control system is at least $\mathcal{O}(T^2)$ ($T$ is the time horizon) due to computing the inverse of  Jacobian  of the  KKT conditions. Since an optimal control system has more sparse structures than general nonlinear or quadratic programs, by exploiting those structures and proposing the  \emph{Auxiliary Control System}, Safe PDP   enjoys the complexity of $\mathcal{O}(T)$ for differentiating  a solution to a general control system. Such advantages have been  discussed and shown in the foundational PDP work \cite{jin2019pontryagin} and will also be shown later (in Section \ref{discussion.section}) in this paper.


\vspace{-5pt}

\subsection{Paper Contributions}
\vspace{-5pt}
We propose a safe differentiable programming methodology  named as \emph{Safe Pontryagin Differentiable Programming} (Safe PDP). Safe PDP provides a systematic treatment of different types of system constraints, including  state and inputs (or mixed), immediate, and long-term constraints, with provable safety- and performance-guarantee.   Safe PDP is also a unified differentiable
programming framework, which can be used to efficiently solve a broad class of safety-critical learning and control tasks.

In the spirit of  interior-point methods, Safe PDP incorporates different types of system constraints into   control cost and loss through  barrier functions, approximating a constrained control system and task using their unconstrained counterparts. Contributions of  Safe PDP are  theoretical and algorithmic. Theoretically, we prove in Theorem \ref{theorem2} and Theorem \ref{theorem3} that  (I)  not only a solution
but also the gradient of the solution  can be safely approximated  by solving a more efficient unconstrained counterpart;
(II) any intermediate results  throughout the approximation and optimization are strictly safe,  that is, never violating the original system/task constraints; 
and (III)  the approximations for both solution and its gradient can be controlled for arbitrary accuracy by  barrier parameters.  Arithmetically, (IV) we  prove  in Theorem \ref{theorem1} that if a constrained control system is differentiable, the gradient of its  trajectory is a globally unique solution to an Auxiliary Control System \cite{jin2019pontryagin}, which can be solved efficiently with the complexity of only  $\mathcal{O}(T)$, $T$ is control horizon;  (V)  in Section  \ref{section.applications}, we  experimentally demonstrate the capability of  Safe PDP  for efficiently solving various safety-critical learning and control problems, including  safe neural policy optimization, safe motion planning,   learning MPCs from demonstrations.

\section{Safe PDP Problem  Formulation}

\vspace{-5pt}

Consider a class of  constrained optimal control systems (models) $\boldsymbol{\Sigma}(\boldsymbol{\theta})$, which are parameterized by a tunable parameter $\boldsymbol{\theta}\in\mathbb{R}^r$ in its control cost function, dynamics, initial condition, and constraints:

	\begin{longfbox}[padding-top=-3pt,margin-top=-5pt, padding-bottom=0pt, margin-bottom=5pt]
		\mathleft
		\begin{equation}\label{equ_oc}
		\boldsymbol{\Sigma}(\boldsymbol{\theta}):\quad
		\begin{aligned}
		\emph{control cost:} &\quad J(\boldsymbol{\theta})=\sum\nolimits_{t=0}^{T{-}1}c_t(\boldsymbol{x}_t,\boldsymbol{u}_t, {\boldsymbol{\theta}})+c_T(\boldsymbol{x}_T,{\boldsymbol{\theta}})\\
		\text{subject to}&\\
		\emph{dynamics:} &\quad \boldsymbol{x}_{t+1}=\boldsymbol{f}(\boldsymbol{x}_{t},\boldsymbol{u}_{t}, {\boldsymbol{\theta}}) \quad \text{with}\quad \boldsymbol{x}_{0}=\boldsymbol{x}_0(\boldsymbol{\theta}),  \quad \forall t, \\
		\emph{terminal constraints:}& \quad
		\boldsymbol{g}_T(\boldsymbol{x}_T,\boldsymbol{\theta})\leq  \boldsymbol{0}, \quad
		\boldsymbol{h}_T(\boldsymbol{x}_T,\boldsymbol{\theta})= \boldsymbol{0},\\
		\emph{path constraints:}& \quad \boldsymbol{g}_t(\boldsymbol{x}_t,\boldsymbol{u}_t,\boldsymbol{\theta})\leq \boldsymbol{0}, \quad \boldsymbol{h}_t(\boldsymbol{x}_t,\boldsymbol{u}_t,\boldsymbol{\theta})= \boldsymbol{0},  \quad \forall t.
		\end{aligned}
		\end{equation}	\mathcenter
	\end{longfbox}
Here, $\boldsymbol{x}_t\in\mathbb{R}^n$ is the system state; $\boldsymbol{u}_t\in\mathbb{R}^m$ is the control input; ${c}_t:\mathbb{R}^n\times\mathbb{R}^m\times\mathbb{R}^r\rightarrow\mathbb{R}$ and $c_T:\mathbb{R}^n\times\mathbb{R}^r\rightarrow\mathbb{R}$ are the stage  and final costs, respectively;  $\boldsymbol{f}:\mathbb{R}^n\times\mathbb{R}^m\times\mathbb{R}^r\rightarrow\mathbb{R}^n$ is the dynamics  with initial state $\boldsymbol{x}_0= \boldsymbol{x}_0(\boldsymbol{\theta})\in\mathbb{R}^{n}$;  $t=0,1,...,T$ is the time step with $T$  the time horizon;   $\boldsymbol{g}_T:\mathbb{R}^n\times\mathbb{R}^r\rightarrow\mathbb{R}^{q_{\text{T}}}$ and  $\boldsymbol{h}_T:\mathbb{R}^n\times\mathbb{R}^r\rightarrow\mathbb{R}^{s_{\text{T}}}$ are the final inequality and equality constraints, respectively;  $\boldsymbol{g}_t:\mathbb{R}^n\times\mathbb{R}^m\times\mathbb{R}^r\rightarrow\mathbb{R}^{q_t}$ and $\boldsymbol{h}_t:\mathbb{R}^n\times\mathbb{R}^m\times\mathbb{R}^r\rightarrow\mathbb{R}^{s_t}$ are the  immediate inequality and equality constraints at time $t$, respectively. 
All  inequalities (here and below) are entry-wise.   We consider that all functions  in $\boldsymbol{\Sigma}(\boldsymbol{\theta})$ are three-times continuously differentiable (i.e., ${C}^3$) with respect to  its arguments. Although we here have parameterized all  aspects of  
$\boldsymbol{\Sigma}(\boldsymbol{\theta})$, for a specific application (see Section \ref{section.applications}), one  only needs to parameterize the unknown aspects in $\boldsymbol{\Sigma}(\boldsymbol{\theta})$  and keep others given. Any  unknown aspects in $\boldsymbol{\Sigma}(\boldsymbol{\theta})$ can be implemented by differentiable neural networks. 
For  a given  $\boldsymbol{\theta}$,  $\boldsymbol{\Sigma}(\boldsymbol{\theta})$ produces a  trajectory $\boldsymbol{\xi}_{\boldsymbol{\theta}}=\{\boldsymbol{x}_{0:T}^{\boldsymbol{\theta}},\boldsymbol{u}_{0:T-1}^{\boldsymbol{\theta}}\}$ by solving the following Problem \ref{equ_traj}:
\begin{equation}\label{equ_traj}
\begin{aligned}
\small\boldsymbol{\xi}_{\boldsymbol{\theta}}=\{\boldsymbol{x}_{0:T}^{\boldsymbol{\theta}},\boldsymbol{u}_{0:T-1}^{\boldsymbol{\theta}}\}\in \arg&\min\nolimits_{\{\boldsymbol{x}_{0:T},\boldsymbol{u}_{0:T-1}\}} \quad J(\boldsymbol{\theta})\\
&\,\,\text{subject to}\quad 
\begin{aligned}[t]
&\small\boldsymbol{x}_{t+1}=\boldsymbol{f}(\boldsymbol{x}_{t},\boldsymbol{u}_{t}, {\boldsymbol{\theta}}) \quad \text{with}\quad \boldsymbol{x}_{0}=\boldsymbol{x}_0(\boldsymbol{\theta}), \quad \forall t,\\
&\small
\boldsymbol{g}_T(\boldsymbol{x}_T,\boldsymbol{\theta})\leq \boldsymbol{0}\quad \text{and}\quad
\boldsymbol{h}_T(\boldsymbol{x}_T,\boldsymbol{\theta})=\boldsymbol{0},\\
&\small \boldsymbol{g}_t(\boldsymbol{x}_t,\boldsymbol{u}_t,\boldsymbol{\theta})\leq \boldsymbol{0} \quad \text{and}\quad \boldsymbol{h}_t(\boldsymbol{x}_t,\boldsymbol{u}_t,\boldsymbol{\theta})= \boldsymbol{0}, \quad \forall t.
\end{aligned}
\end{aligned}\tag*{B($\boldsymbol{\theta}$)}
\end{equation}
Here, we use  $\in$ since  $\boldsymbol{\xi}_{\boldsymbol{\theta}}$ to Problem \ref{equ_traj} may not be unique in general,  thus constituting a solution set $\{\boldsymbol{\xi}_{\boldsymbol{\theta}}\}$. We will  discuss the existence and uniqueness of $\{\boldsymbol{\xi}_{\boldsymbol{\theta}}\}$  in Section \ref{section.diffcpmp_auxsys}.

For a specific  task, we aim to find a specific model $\boldsymbol{\Sigma}(\boldsymbol{\theta}^*)$, i.e, searching for a specific $\boldsymbol{\theta}^*$, such that its    trajectory  $\boldsymbol{\xi}_{\boldsymbol{\theta}^*}$
from \textcolor{red}{B($\boldsymbol{\theta}^*$)} meets the following two \emph{given} requirements. 
First, $\boldsymbol{\xi}_{\boldsymbol{\theta}^*}$  minimizes a \emph{given task loss} $\ell(\boldsymbol{\xi}_{\boldsymbol{\theta}}, \boldsymbol{\theta})$; 
and second, $\boldsymbol{\xi}_{\boldsymbol{\theta}^*}$ satisfies the  \emph{given task constraints} ${R}_i(\boldsymbol{\xi}_{\boldsymbol{\theta}}, \boldsymbol{\theta})\leq {0}$, $i=1,2,...,l$.  Note that, we  need to distinguish between the two types of objectives: task loss  $\ell(\boldsymbol{\xi}_{\boldsymbol{\theta}}, \boldsymbol{\theta})$  and control cost $J(\boldsymbol{\theta})$,  and also  the two types of constraints: task constraints ${R}_i(\boldsymbol{\xi}_{\boldsymbol{\theta}}, \boldsymbol{\theta})$ and model constraints $\boldsymbol{g}_t(\boldsymbol{\theta})$. In fact,  $J(\boldsymbol{\theta})$ and  $\boldsymbol{g}_t(\boldsymbol{\theta})$ in $\boldsymbol{\Sigma}(\boldsymbol{\theta})$ are  \emph{unknown and parameterized by $\boldsymbol{\theta}$} and can represent the   unknown inherent aspects of a physical  agent, while  $\ell(\boldsymbol{\xi}_{\boldsymbol{\theta}}, \boldsymbol{\theta})$  and ${R}_i(\boldsymbol{\xi}_{\boldsymbol{\theta}}, \boldsymbol{\theta})$ are \emph{given  and known} depending on the specific task (they also explicitly depend on  $\boldsymbol{\theta}$ since  $\boldsymbol{\theta}$ needs to be
regularized in some learning cases).  Assume  $\ell(\boldsymbol{\xi}_{\boldsymbol{\theta}}, \boldsymbol{\theta})$ and   ${R}_i(\boldsymbol{\xi}_{\boldsymbol{\theta}}, \boldsymbol{\theta})$ are both twice-continuously differentiable. The  problem of searching for $\boldsymbol{\theta}^*$ can be formally written as:
\begin{equation}\label{equ_problem}
\begin{aligned}
\boldsymbol{\theta}^*\,\,=\,\,\arg\min_{\boldsymbol{\theta}} \quad & \ell(\boldsymbol{\xi}_{\boldsymbol{\theta}},\boldsymbol{\theta}) \quad \\ \text{subject to}\quad 
&  R_i(\boldsymbol{\xi}_{\boldsymbol{\theta}}, \boldsymbol{\theta})\leq 0, \quad i=1,2,..., l,\\
& \boldsymbol{\xi}_{\boldsymbol{\theta}} \,\, \text{solves  Problem \ref{equ_traj}\,\,.}
\end{aligned}\tag*{P}
\end{equation}
For a specific learning and control task, one only needs to specify the  details of $\boldsymbol{\Sigma}(\boldsymbol{\theta})$
and give a  task loss  $\ell(\boldsymbol{\xi}_{\boldsymbol{\theta}},\boldsymbol{\theta})$ and  constraints  $R_i(\boldsymbol{\xi}_{\boldsymbol{\theta}}, \boldsymbol{\theta})\leq {0}$.
Section \ref{section.applications} will give representative examples.

\vspace{-3pt}

\section{Challenges to Solve Problem \ref{equ_problem}} \label{section.challenge}

\vspace{-5pt}

Problem  \ref{equ_problem} belongs to  bi-level optimization \cite{sinha2017review}---each time  $\boldsymbol{\theta}$ is updated in the outer-level (including task loss $\ell$ and task constraint $R_i$) of Problem \ref{equ_problem},  the corresponding trajectory  $\boldsymbol{\xi}_{\boldsymbol{\theta}}$ needs  to be solved from the inner-level  Problem \ref{equ_traj}.
Similar to PDP \cite{jin2019pontryagin},   one could approach  Problem \ref{equ_problem} using gradient-based methods by ignoring the process of solving inner-level  Problem \ref{equ_traj} and just viewing $\boldsymbol{\xi}_{\boldsymbol{\theta}}$ as  an explicit differentiable  function of $\boldsymbol{\theta}$. Then, based on interior-point methods \cite{fiacco1990nonlinear}, one can introduce a logarithmic barrier function for each task constraint, $-\ln\big({-}R_i\big)$, and a barrier parameter $\epsilon>0$. This leads to solving the following unconstrained Problem \ref{equ_problem_penalty} sequentially
\begin{equation}\label{equ_problem_penalty}
\boldsymbol{\theta}^*(\epsilon)= \arg\min_{\boldsymbol{{\theta}}}\,\, \ell(\boldsymbol{\xi}_{\boldsymbol{\theta}},\boldsymbol{\theta})-\epsilon\sum\nolimits_{i=1}^{l}\ln\big({-}R_i(\boldsymbol{\xi}_{\boldsymbol{\theta}},\boldsymbol{\theta})\big)
\tag*{SP(${\epsilon}$)}
\end{equation}
for a fixed $\epsilon$. By controlling  $\epsilon\rightarrow0$, $\boldsymbol{\theta}^*(\epsilon)$ is expected to converge to the solution $\boldsymbol{\theta}^*$ to Problem \ref{equ_problem}. Although  plausible, the above process has the following technical challenges  to be addressed:

\vspace{-5pt}
\begin{itemize}[leftmargin=0.60cm]
			\setlength\itemsep{-0.015em}
	\item[(1)] As $\boldsymbol{\xi}_{\boldsymbol{\theta}}$ is a solution to the constrained optimal control Problem \ref{equ_traj}, is $\boldsymbol{\xi}_{\boldsymbol{\theta}}$  differentiable? Does the auxiliary control system \cite{jin2019pontryagin} exist for solving $\frac{\partial\boldsymbol{\xi}_{\boldsymbol{{\theta}}}}{\partial \boldsymbol{{\theta}}}$?
	\item[(2)]  Since we want to obtain $\boldsymbol{\xi}_{\boldsymbol{\theta}}$ and  $\frac{\partial\boldsymbol{\xi}_{\boldsymbol{{\theta}}}}{\partial \boldsymbol{{\theta}}}$ at as low  cost as possible, instead of solving  the {constrained} Problem \ref{equ_traj}, can we use an \emph{unconstrained} system to approximate both $\boldsymbol{\xi}_{\boldsymbol{\theta}}$ and  $\frac{\partial\boldsymbol{\xi}_{\boldsymbol{{\theta}}}}{\partial \boldsymbol{{\theta}}}$? Importantly, can  the accuracy of the approximations for  $\boldsymbol{\xi}_{\boldsymbol{\theta}}$ and  $\frac{\partial\boldsymbol{\xi}_{\boldsymbol{{\theta}}}}{\partial \boldsymbol{{\theta}}}$  be arbitrarily and safely controlled?
	
	\item[(3)]  {Can we guarantee that the  approximation for    $\boldsymbol{\xi}_{\boldsymbol{\theta}}$ is safe in a sense that the  approximation  always respects the system original inequality constraints $\boldsymbol{g}_t\leq 0$ and $\boldsymbol{g}_T\leq 0$}?  
	
	\smallskip
	\item[(4)] 
With the safe approximations  for both $\boldsymbol{\xi}_{\boldsymbol{\theta}}$ and $\frac{\partial \boldsymbol{\xi}_{\boldsymbol{\theta}}}{\partial \boldsymbol{\theta}}$, can  accuracy of the solution to  the   outer-level unconstrained optimization \ref{equ_problem_penalty}  be arbitrarily controlled towards $\boldsymbol{\theta}^*$? 
 	\smallskip
 \item[(5)]With the safe approximations for both $\boldsymbol{\xi}_{\boldsymbol{\theta}}$ and $\frac{\partial \boldsymbol{\xi}_{\boldsymbol{\theta}}}{\partial \boldsymbol{\theta}}$, 
 can we guarantee  the safety of the outer-level inequality constraints $R_i\leq 0$ during the  optimization for the outer-level  \ref{equ_problem_penalty}?	

\end{itemize}

The following paper will address the above challenges. For  reference, we  give  a quick overview: Challenge (1) will be addressed in Section \ref{section.diffcpmp_auxsys} and the result is in Theorem~\ref{theorem1}; Challenges (2) and (3) will be addressed in Section \ref{keysection} and the result is in Theorem~\ref{theorem2}; Challenges (4) and (5) will be addressed in Section \ref{section.safeP}  and the result is in Theorem \ref{theorem3}; and  Section \ref{section.applications}  gives some representative applications.

\section{Differentiability for $\boldsymbol{\Sigma}(\boldsymbol{\theta})$ and its Auxiliary Control System}\label{section.diffcpmp_auxsys}

\vspace{-5pt}

\subsection{Differentiability of $\boldsymbol{\xi}_{\boldsymbol{{\theta}}}$}

\vspace{-3pt}

For the constrained optimal control system $\boldsymbol{\Sigma}(\boldsymbol{\theta})$ in (\ref{equ_oc}), we define the following  Hamiltonian $L_t$ for  $t={0,1,...,T{-}1}$ and $L_T$, respectively,
\begin{subequations}\label{Hamiltonian}
	\begin{align}
		L_t&=c_t(\boldsymbol{x}_t,\boldsymbol{u}_t,\boldsymbol{\theta})+\boldsymbol{\lambda}^\prime_{t{+}1}\boldsymbol{f}(\boldsymbol{x}_t,\boldsymbol{u}_t,\boldsymbol{\theta})+\boldsymbol{v}_t^\prime \boldsymbol{g}_t(\boldsymbol{x}_t,\boldsymbol{u}_t,\boldsymbol{\theta})+\boldsymbol{w}_t^\prime \boldsymbol{h}_t(\boldsymbol{x}_t,\boldsymbol{u}_t,\boldsymbol{\theta}),\\
		L_T&=c_T(\boldsymbol{x}_T,\boldsymbol{\theta})+\boldsymbol{v}_T^\prime\boldsymbol{g}_T(\boldsymbol{x}_T,\boldsymbol{\theta})+\boldsymbol{w}_T^\prime\boldsymbol{h}_T(\boldsymbol{x}_T,\boldsymbol{\theta}),
	\end{align}
\end{subequations}
where $\boldsymbol{\lambda}_t\in\mathbb{R}^n$ is the costate, $\boldsymbol{v}_t\in\mathbb{R}^{q_t}$ and $\boldsymbol{w}_t\in\mathbb{R}^{s_t}$  are   multipliers for the inequality and equality constraints, respectively. The well-known second-order  condition  for $\boldsymbol{\xi}_{\boldsymbol{\theta}}$ to be \emph{a local isolated (locally unique) minimizing trajectory} to $\boldsymbol{\Sigma}(\boldsymbol{\theta})$ in  Problem \ref{equ_traj} has been well-established in \cite{pearson1966discrete}. For  completeness, we   present it in Lemma \ref{Lemma1} in Appendix \ref{appendix.secondordercondition}.  Lemma \ref{Lemma1}  states that there exist costate sequence $\boldsymbol{\lambda}_{1:T}^{\boldsymbol{\theta}}$, and multiplier sequences $\boldsymbol{v}_{0:T}^{\boldsymbol{\theta}}$ and $\boldsymbol{w}_{0:T}^{\boldsymbol{\theta}}$, such that ($\boldsymbol{\xi}_{\boldsymbol{{\theta}}}$, $ \boldsymbol{\lambda}_{1:T}^{{\boldsymbol{{\theta}}}}$, $\boldsymbol{v}_{0:T}^{{\boldsymbol{{\theta}}}}$, $\boldsymbol{w}_{0:T}^{{\boldsymbol{{\theta}}}}$)
satisfies the well-known Constrained Pontryagin Minimum Principle (C-PMP)  given in (\ref{CPMP}) in Lemma \ref{Lemma1}.  Based on the above, one can have the following result for the differentiability of $\boldsymbol{\xi}_{\boldsymbol{{\theta}}}$.

\begin{lemma}[Differentiability of $\boldsymbol{\xi}_{\boldsymbol{{\theta}}}$] \label{result1}
	Given a fixed ${\boldsymbol{\bar\theta}}$, assume the following conditions hold for   $\boldsymbol{\Sigma}({{
			\boldsymbol{\bar\theta}}})$:
	\vspace{-5pt}
	
	\begin{itemize}
		\setlength\itemsep{-0.02em}
		\item[(i)] the second-order condition (Lemma \ref{Lemma1})  is satisfied for $\boldsymbol{\Sigma}({{
				\boldsymbol{\bar\theta}}})$;
		\item[(ii)] the gradients of all binding constraints  at $\boldsymbol{\xi}_{{\boldsymbol{\bar{\theta}}}}$   are linearly independent (binding constraints include all equality constraints and all active inequality constraints);
		\item[(iii)] strict complementarity holds at $\boldsymbol{\xi}_{{\boldsymbol{\bar{\theta}}}}$, i.e.,  active inequality constraint has    positive multiplier.
	\end{itemize}
	
	\vspace{-5pt}
	Then, for all  $\boldsymbol{\theta}$ in a neighborhood of ${\boldsymbol{\bar\theta}}$, there exists  a unique once-continuously differentiable function ($\boldsymbol{\xi}_{\boldsymbol{{\theta}}}$,$ \boldsymbol{\lambda}_{1:T}^{{\boldsymbol{{\theta}}}}$,$\boldsymbol{v}_{0:T}^{{\boldsymbol{{\theta}}}}$,$\boldsymbol{w}_{0:T}^{{\boldsymbol{{\theta}}}}$) 
	that satisfies the second-order condition  (Lemma \ref{Lemma1}) for  the  constrained optimal control system  $\boldsymbol{\Sigma}(\boldsymbol{\theta})$ with $\big(\boldsymbol{\xi}_{{\boldsymbol{{\theta}}}},\boldsymbol{\lambda}_{1:T}^{{\boldsymbol{{\theta}}}},\boldsymbol{v}_{0:T}^{{\boldsymbol{{\theta}}}},\boldsymbol{w}_{0:T}^{{\boldsymbol{{\theta}}}}\big)=\big(\boldsymbol{\xi}_{{\boldsymbol{\bar{\theta}}}},\boldsymbol{\lambda}_{1:T}^{{\boldsymbol{\bar{\theta}}}},\boldsymbol{v}_{0:T}^{{\boldsymbol{\bar{\theta}}}},\boldsymbol{w}_{0:T}^{\bar{\boldsymbol{{\theta}}}}\big)$ at $\boldsymbol{\theta}{=}{\boldsymbol{\bar\theta}}$.
	Hence, $\boldsymbol{\xi}_{\boldsymbol{{\theta}}}$ is a local isolated minimizing trajectory to  $\boldsymbol{\Sigma}(\boldsymbol{\theta})$. Further, for  all    $\boldsymbol{\theta}$ near ${\boldsymbol{\bar\theta}}$, the strict complementarity  is preserved, and the linear independence of the gradients of all binding constraints  at $\boldsymbol{\xi}_{{\boldsymbol{{\theta}}}}$   hold.	
\end{lemma}

The proof of Lemma \ref{result1} can directly follow  the well-known first-order sensitivity result in Theorem~2.1 in \cite{fiacco1976sensitivity}. Here,
 conditions (i)-(iii) are the sufficient conditions to guarantee the applicability of  the well-known implicit function theorem \cite{rudin1976principles} to the C-PMP. Condition (ii) is well-known and serves as a sufficient condition for the constraint qualification  to establish the C-PMP (see Corollary 3, pp. 22, \cite{fiacco1990nonlinear}). Condition (iii) is necessary  to ensure that the Jacobian matrix in the implicit function theorem is invertible, and it also  leads to the persistence of strict complementarity,  saying that the inactive inequalities remain inactive and active ones remain active and there is no `switching' between them near $\boldsymbol{\bar{\theta}}$. Both our practice and previous works  \cite{kolstad1990derivative,fiacco1976sensitivity,amos2017optnet,amos2018differentiable,jin2019pontryagin}  show that  the  conditions (i)-(iii) are very mild and the differentiability of $\boldsymbol{\xi}_{\boldsymbol{{\theta}}}$ can be attained almost everywhere in the space of $\boldsymbol{\theta}$.

\vspace{-5pt}

\subsection{Auxiliary Control System to Solve $\frac{\partial\boldsymbol{\xi}_{\boldsymbol{\theta}}}{\partial\boldsymbol{\theta}}$ }\label{section.auxsys}

\vspace{-5pt}

If the conditions (i)-(iii) in Lemma \ref{result1} for  differentiability of $\boldsymbol{\xi}_{\boldsymbol{{\theta}}}$  hold, we next show that $\frac{\partial\boldsymbol{\xi}_{\boldsymbol{\theta}}}{\partial\boldsymbol{\theta}}$ can also be efficiently solved by  an auxiliary control system, which is originally proposed in the foundational work  \cite{jin2019pontryagin}.
First, we define the  new state and  input  (matrix) variables
$
X_t\in\mathbb{R}^{n\times r}$ and $U_t\in\mathbb{R}^{m\times r},
$
respectively. Then, we introduce the following \emph{auxiliary control system},

\begin{longfbox}[padding-top=-3pt,margin-top=-4pt, padding-bottom=0pt, margin-bottom=5pt]
	\mathleft
	\begin{equation}\label{equ_aux}
	\boldsymbol{\overline\Sigma}(\boldsymbol{\xi}_{\boldsymbol{\theta}}):\,\,\,
	\begin{aligned}
	\emph{control cost:} &\quad \bar{J}=\small\Tr\sum_{t=0}^{T{-}1}
	\left(\frac{1}{2}\begin{bmatrix}
	X_t\\
	U_t
	\end{bmatrix}^\prime\begin{bmatrix}
	L_t^{xx}& L_t^{xu}\\
	L_t^{ux}& L_t^{uu}
	\end{bmatrix}\begin{bmatrix}
	X_t\\
	U_t
	\end{bmatrix}+
	\begin{bmatrix}
	L_t^{x\theta}\\
	L_t^{u\theta}
	\end{bmatrix}^\prime\begin{bmatrix}
	X_t\\
	U_t
	\end{bmatrix}
	\right)
	\\
	&\qquad\quad+\small\Tr\left(\frac{1}{2}X_T^\prime L_T^{xx}X_T+(L_T^{x\theta})^\prime X_T\right)\\
	\text{subject to}&\\
	\emph{dynamics:} &\quad \small{X}_{t+1}=F_t^xX_t
	+F_t^uU_{t}+F_t^\theta\quad \text{with}\quad {X}_{0}=X_{0}^{\boldsymbol{\theta}} \\
	\emph{terminal constraint:}& \quad\small
	\bar{G}_T^xX_T+\bar{G}_T^\theta = \boldsymbol{0}, \quad\quad 
	H_T^xX_T
	+H_T^\theta= \boldsymbol{0},\\
	\emph{path constraint:}& \quad \small	\bar{G}_t^xX_t+\bar{G}_t^uU_t+\bar{G}_t^\theta = \boldsymbol{0}, \quad\quad 
	H_t^xX_t
	+H_t^uU_t
	+H_t^\theta= \boldsymbol{0}. 
	\end{aligned}
	\end{equation}	\mathcenter
\end{longfbox}
Here,  $L_t^x$ and $L_t^{xx}$ denote the first- and second- order derivatives, respectively, of the Hamiltonian $L_t$ in (\ref{Hamiltonian}) with respect to $\boldsymbol{x}$; $F_t^x$, $H_t^x$, $\bar{{G}}_t$ denote the first-order derivatives  of $\boldsymbol{f}_t$, $\boldsymbol{h}_t$, $\boldsymbol{\bar{g}}_t$ with respect to $\boldsymbol{x}$,  respectively,  where $\boldsymbol{\bar{g}}_t$ is the  vector of stacking all active inequality constraints in $\boldsymbol{g}_t$; and the similar   convention applies to the other notations. All  derivative  matrices defining ${\boldsymbol{\overline\Sigma}}(\boldsymbol{\xi}_{\boldsymbol{\theta}})$ are evaluated at $\big(\boldsymbol{\xi}_{{\boldsymbol{{\theta}}}},\boldsymbol{\lambda}_{1:T}^{{\boldsymbol{{\theta}}}},\boldsymbol{v}_{0:T}^{{\boldsymbol{{\theta}}}},\boldsymbol{w}_{0:T}^{{\boldsymbol{{\theta}}}}\big)$, where  $\boldsymbol{\lambda}_{1:T}^{{\boldsymbol{{\theta}}}}$, $\boldsymbol{v}_{0:T}^{{\boldsymbol{{\theta}}}}$, and $\boldsymbol{w}_{0:T}^{{\boldsymbol{{\theta}}}}$  are usually the byproducts of a constrained optimal control solver \cite{andersson2019casadi} or can be easily solved from the C-PMP given $\boldsymbol{\xi}_{\boldsymbol{\theta}}$, as done in \cite{jin2019pontryagin}.
We note that  ${\boldsymbol{\overline\Sigma}}(\boldsymbol{\xi}_{\boldsymbol{\theta}})$ is a \emph{Equality-constrained Linear Quadratic Regulator (LQR)}  system, as its control cost function is quadratic and dynamics and constraints are linear.
For the above ${\boldsymbol{\overline\Sigma}}(\boldsymbol{\xi}_{\boldsymbol{\theta}})$, we have the following important result \emph{without additional assumptions}.

\begin{theorem}[$\frac{\partial\boldsymbol{\xi}_{\boldsymbol{\theta}}}{\partial\boldsymbol{\theta}}$ is a globally unique minmizing trajectory  to  $\boldsymbol{\overline\Sigma}(\boldsymbol{\xi}_{\boldsymbol{\theta}})$]\label{theorem1} 
	
	Let  the  conditions (i)-(iii)~in Lemma~\ref{result1}  for   differentiability of  $\boldsymbol{\xi}_{\boldsymbol{{\theta}}}$   hold. Then, the auxiliary control system $\overline{\boldsymbol{\Sigma}}(\boldsymbol{\xi}_{\boldsymbol{\theta}})$  in (\ref{equ_aux}) has a globally unique  minimizing trajectory, denoted as $\left\{
		X^{{\boldsymbol{{\theta}}}}_{0:T}, U^{{\boldsymbol{{\theta}}}}_{0:T-1}
		\right\}$, which is exactly $\frac{\partial\boldsymbol{\xi}_{{\boldsymbol{{\theta}}}} }{\partial \boldsymbol{\theta}}$, i.e.,
		\begin{equation}\label{difftraj}
		\small\left\{
		X^{{\boldsymbol{{\theta}}}}_{0:T}, U^{{\boldsymbol{{\theta}}}}_{0:T-1}
		\right\}=\frac{\partial\boldsymbol{\xi}_{{\boldsymbol{{\theta}}}} }{\partial \boldsymbol{\theta}} \quad
		\text{with}
		\quad X^{{\boldsymbol{{\theta}}}}_{t}=\frac{\partial\boldsymbol{x}_t^{{\boldsymbol{{\theta}}}} }{\partial \boldsymbol{\theta}}    \quad
		\text{and}
		\quad U^{{\boldsymbol{{\theta}}}}_{t}= \frac{\partial\boldsymbol{u}_t^{{\boldsymbol{{\theta}}}} }{\partial \boldsymbol{\theta}}.
		\end{equation} 
\end{theorem}
The proof of the above theorem is  in Appendix \ref{pf_theorem1}. Theorem \ref{theorem1} states that as long as the conditions (i)-(iii) in Lemma \ref{result1} for  differentiability of  $\boldsymbol{\xi}_{\boldsymbol{{\theta}}}$  are satisfied, without additional assumptions, the  auxiliary control system $\boldsymbol{\overline\Sigma}(\boldsymbol{\xi}_{\boldsymbol{\theta}})$ always has a globally unique  minimizing trajectory, which is exactly  $\frac{\partial\boldsymbol{\xi}_{{\boldsymbol{{\theta}}}} }{\partial \boldsymbol{\theta}}$.
Thus, obtaining  $\frac{\partial\boldsymbol{\xi}_{{\boldsymbol{{\theta}}}} }{\partial \boldsymbol{\theta}}$ is equivalent to solving  $\boldsymbol{\overline\Sigma}(\boldsymbol{\xi}_{\boldsymbol{\theta}})$, which be efficiently done thanks to the recent development of the equality-constrained LQR algorithms \cite{sideris2010riccati,yang2020equality,laine2019efficient}, all of which have a complexity of $\mathcal{O }(T)$. The  algorithm that implements Theorem \ref{theorem1} is  given in Algorithm \ref{alg_theorem1} in Appendix \ref{appendix.algorithm1}.

\vspace{-2pt}

\section{Safe Unconstrained  Approximations for $\boldsymbol{\xi}_{\boldsymbol{\theta}}$ and  $\frac{\partial\boldsymbol{\xi}_{\boldsymbol{\theta}}}{\partial\boldsymbol{\theta}}$} \label{keysection}

\vspace{-5pt}

From Section \ref{section.diffcpmp_auxsys}, we know that one can solve the constrained system  $\boldsymbol{\Sigma}(\boldsymbol{\theta})$ to obtain $\boldsymbol{\xi}_{\boldsymbol{\theta}}$ and solve its auxiliary control system $\boldsymbol{\overline\Sigma}(\boldsymbol{\xi}_{\boldsymbol{\theta}})$ to obtain  $\frac{\partial\boldsymbol{\xi}_{\boldsymbol{\theta}}}{\partial\boldsymbol{\theta}}$. Although theoretically appealing, there are several  difficulties in implementation. First, solving  a constrained optimal control Problem \ref{equ_traj} is not as easy as solving an unconstrained optimal control, for which many  trajectory optimization algorithms, e.g.,  iLQR \cite{li2004iterative} and DDP \cite{jacobson1970differential}, are available. Second, 
 establishing  $\boldsymbol{\overline\Sigma}(\boldsymbol{\xi}_{\boldsymbol{\theta}})$  requires the values of the  multipliers  $\boldsymbol{{v}}_{0:T}^{{\boldsymbol{{\theta}}}}$ and 
$\boldsymbol{w}_{0:T}^{{\boldsymbol{{\theta}}}}$. And third, to construct $\boldsymbol{\overline\Sigma}(\boldsymbol{\xi}_{\boldsymbol{\theta}})$, one also needs to identify all   active inequality constraints $\boldsymbol{\bar{g}}_t$, which can be numerically difficult due to numerical error (we will show this in later experiments). All those difficulties  motivate us to develop a more efficient paradigm to obtain both $\boldsymbol{\xi}_{\boldsymbol{\theta}}$ and  $\frac{\partial\boldsymbol{\xi}_{\boldsymbol{\theta}}}{\partial\boldsymbol{\theta}}$, which is the goal  of this section.

To  proceed, we first convert the constrained  system  $\boldsymbol{\Sigma}(\boldsymbol{\theta})$  to an unconstrained  system ${\boldsymbol{\Sigma}}(\boldsymbol{\theta},\gamma)$ by adding all  constraints to its control cost  via  barrier functions. Here, we use  quadratic barrier function for each equality constraint and logarithm barrier functions for each inequality constraint; and  all barrier functions are associated with the same barrier parameter  $\gamma>0$. This leads to ${\boldsymbol{\Sigma}}(\boldsymbol{\theta},\gamma)$ to be

{
	\begin{longfbox}[padding-top=-3pt,margin-top=-4pt, padding-bottom=3pt, margin-bottom=5pt]
		\mathleft
		\begin{equation}\label{equ_oc_penality}
		{\boldsymbol{\Sigma}}(\boldsymbol{\theta},\gamma):\qquad
		\begin{aligned}
		\emph{control cost:} & \quad J(\boldsymbol{\theta},\gamma)=\small\sum_{t=0}^{T{-}1}
		\Big(
		c_t(\boldsymbol{x}_t,\boldsymbol{u}_t,{\boldsymbol{\theta}})
		{-}\gamma\sum_{i=1}^{q_t}\ln\big({-}g_{t,i}(\boldsymbol{x}_t,\boldsymbol{u}_t,{\boldsymbol{\theta}})\big)
		{+}\\[-3pt]
		&\qquad\qquad\qquad\quad\small\frac{1}{2\gamma}\sum_{i=1}^{s_t}\big(h_{t,i}(\boldsymbol{x}_t,\boldsymbol{u}_t, {\boldsymbol{\theta}})\big)^2
		\Big)
		+c_T(\boldsymbol{x}_T,{\boldsymbol{\theta}}) -\\[-3pt]
		&\qquad\qquad\quad\,\,\small\gamma\sum_{i=1}^{q_T}\ln\big({-}g_{T,i}(\boldsymbol{x}_T,{\boldsymbol{\theta}})\big){+}
		\frac{1}{2\gamma}\sum_{i=1}^{s_T}\big(h_{T,i}(\boldsymbol{x}_T, {\boldsymbol{\theta}})\big)^2,
		\\[2pt]
			\text{subject to}&\\
		\emph{dynamics:} & \quad \boldsymbol{x}_{t+1}=\boldsymbol{f}(\boldsymbol{x}_{t},\boldsymbol{u}_{t}, {\boldsymbol{\theta}}) \quad \text{with}\quad \boldsymbol{x}_{0}=\boldsymbol{x}_0(\boldsymbol{\theta}),  \quad \forall t.
		\end{aligned}
		\end{equation}	\mathcenter
	\end{longfbox}
}
The  trajectory $\small\boldsymbol{\xi}_{(\boldsymbol{\theta},\gamma)}=\left\{\boldsymbol{x}_{0:T}^{(\boldsymbol{\theta},\gamma)}, \boldsymbol{u}_{0:T-1}^{(\boldsymbol{\theta},\gamma)}\right\}$ produced by the  above unconstrained  system  ${\boldsymbol{\Sigma}}(\boldsymbol{\theta},\gamma)$ is
\begin{equation}\label{pdpapprx_oc}
\begin{aligned}
\boldsymbol{\xi}_{(\boldsymbol{\theta},\gamma)}=\left\{\boldsymbol{x}_{0:T}^{(\boldsymbol{\theta},\gamma)}, \boldsymbol{u}_{0:T-1}^{(\boldsymbol{\theta},\gamma)}\right\}\in\arg&\min\nolimits_{\{\boldsymbol{x}_{0:T},\boldsymbol{u}_{0:T-1}\}} \quad J(\boldsymbol{\theta},\gamma)\\
&\,\,\text{s.t.}\quad \boldsymbol{x}_{t+1}=\boldsymbol{f}(\boldsymbol{x}_{t},\boldsymbol{u}_{t}, {\boldsymbol{\theta}}) \quad \text{with}\quad \boldsymbol{x}_{0}=\boldsymbol{x}_0(\boldsymbol{\theta}),
\end{aligned}\tag*{SB(${\boldsymbol{\theta},\gamma}$)}
\end{equation}
that is, $\boldsymbol{\xi}_{(\boldsymbol{\theta},\gamma)}$ is minimizing the new control cost $J(\boldsymbol{\theta},\gamma)$ subject to only dynamics. Then we have the following important result about the  safe unconstrained  approximation for  $\boldsymbol{\xi}_{\boldsymbol{\theta}}$ and  $\frac{\partial\boldsymbol{\xi}_{\boldsymbol{\theta}}}{\partial\boldsymbol{\theta}}$ using ${\boldsymbol{\Sigma}}(\boldsymbol{\theta},\gamma)$.

\vspace{2pt}
\begin{theorem}\label{theorem2}
	Let   conditions (i)-(iii)   in Lemma \ref{result1}  for   differentiability of  $\boldsymbol{\xi}_{\boldsymbol{{\theta}}}$ hold. For any small $\gamma>0$, 
	\begin{itemize}
		\setlength\itemsep{-0.02em}
		\item[(a)]  there exists a  local isolated minimizing trajectory $\boldsymbol{\xi}_{(\boldsymbol{\theta},\gamma)}$ that solves Problem \ref{pdpapprx_oc}, and  ${\boldsymbol{\Sigma}}(\boldsymbol{\theta},\gamma)$ is well-defined at $\boldsymbol{\xi}_{(\boldsymbol{\theta},\gamma)}$, i.e., $\boldsymbol{g}_{t}\big(\boldsymbol{x}_t^{(\boldsymbol{\theta},\gamma)},\boldsymbol{u}_t^{(\boldsymbol{\theta},\gamma)},\boldsymbol{\theta}\big){<} \boldsymbol{0}$ and 
		$\boldsymbol{g}_{T}\big(\boldsymbol{x}_T^{(\boldsymbol{\theta},\gamma)},\boldsymbol{\theta}\big){<} \boldsymbol{0}$;
		\item[(b)]  $\boldsymbol{\xi}_{(\boldsymbol{\theta},\gamma)}$ is once-continuously differentiable with respect to $(\boldsymbol{\theta},\gamma)$, and 
		\begin{align}\label{theorem2assertion}
		&\boldsymbol{\xi}_{(\boldsymbol{\theta},\gamma)}\rightarrow\boldsymbol{\xi}_{\boldsymbol{\theta}}\quad \text{and} \quad \small\frac{\partial \boldsymbol{\xi}_{(\boldsymbol{\theta},\gamma)}}{\partial \boldsymbol{\theta}}\normalsize\rightarrow\small\frac{\partial \boldsymbol{\xi}_{\boldsymbol{\theta}}}{\partial \boldsymbol{\theta}} \quad  \text{as}\quad \gamma \rightarrow 0;
		\end{align}
		\vspace{-8pt}
		\item[(c)] the trajectory derivative $\frac{\partial \boldsymbol{\xi}_{(\boldsymbol{\theta},\gamma)}}{\partial \boldsymbol{\theta}}$  is a  globally unique minimizing trajectory to  the  auxiliary control system ${\boldsymbol{\overline\Sigma}}\big(\boldsymbol{\xi}_{(\boldsymbol{\theta},\gamma)}\big)$ corresponding to $\boldsymbol{\Sigma}(\boldsymbol{\theta},\gamma)$.
	\end{itemize}
\end{theorem}

\begin{wrapfigure}[15]{r}{0pt}
	\raisebox{0pt}[\dimexpr\height-1.2\baselineskip]{\includegraphics[width=5.6cm]{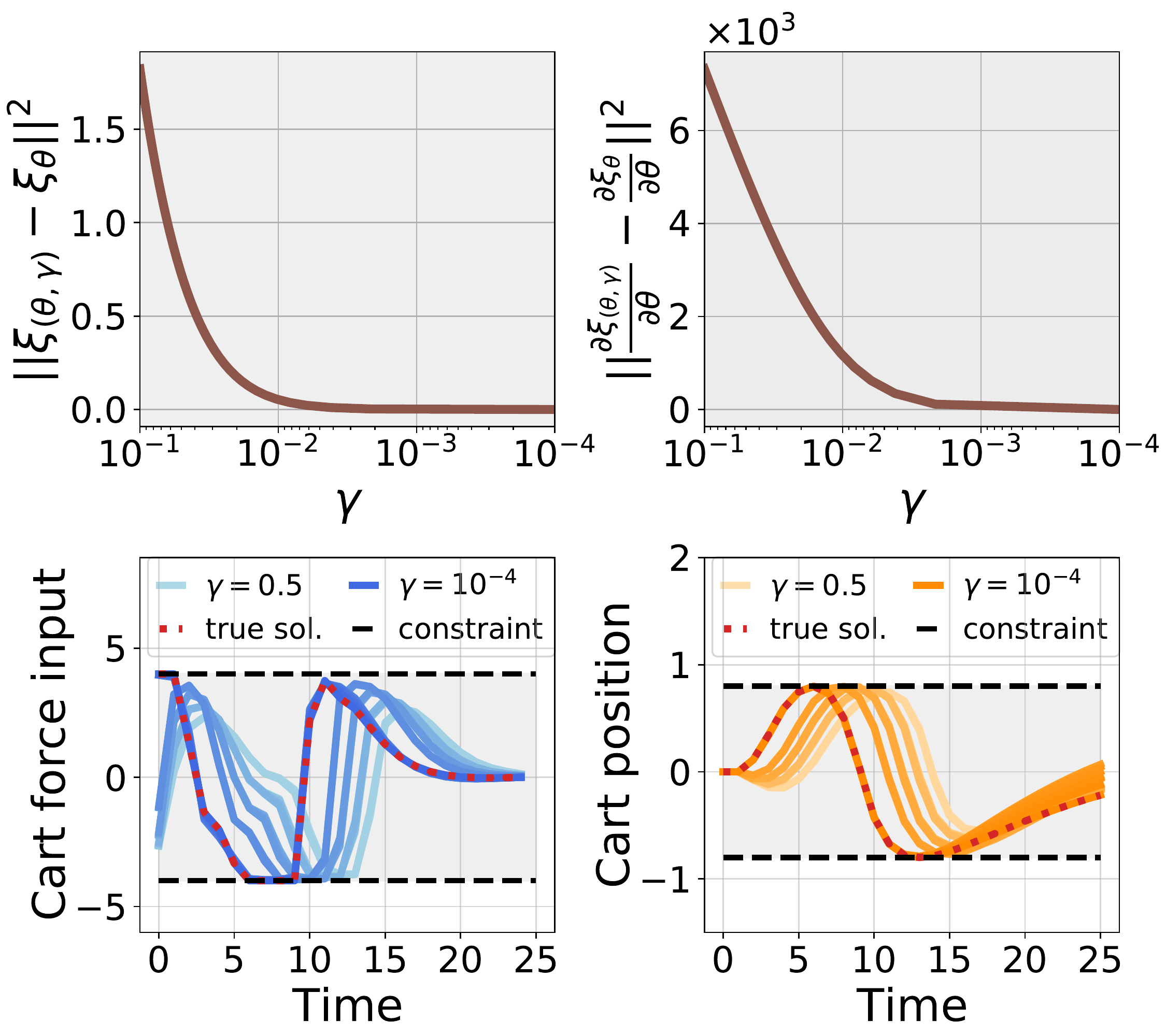}}
	\vspace{-3pt}
	\caption{ $\small\boldsymbol{\xi}_{(\boldsymbol{\theta},\gamma)}$ and  $\small\frac{\partial \boldsymbol{\xi}_{(\boldsymbol{\theta},\gamma)}}{\partial \boldsymbol{\theta}}$  approximate  $\small\boldsymbol{\xi}_{\boldsymbol{\theta}}$ and  $\small\frac{\partial \boldsymbol{\xi}_{\boldsymbol{\theta}}}{\partial \boldsymbol{\theta}}$ under different  $\gamma>0$.}
	\label{figthem}
\end{wrapfigure}

The proof of the above theorem is given in Appendix \ref{pf_theorem2}.  
It is worth noting that the above assertions require no additional assumption except the same  conditions (i)-(iii) for  differentiability of $\boldsymbol{\xi}_{\boldsymbol{\theta}}$. We make the following comments on the above results, using an illustrative cartpole example in Fig. \ref{figthem}. 
 First,  assertion (b) states that by choosing a small $\gamma>0$, one can simply use  $\boldsymbol{\xi}_{(\boldsymbol{\theta},\gamma)}$ and   $\small\frac{\partial \boldsymbol{\xi}_{(\boldsymbol{\theta},\gamma)}}{\partial \boldsymbol{\theta}}$ of the \emph{unconstrained optimal control system}   ${\boldsymbol{\Sigma}}(\boldsymbol{\theta},\gamma)$ to approximate  $\boldsymbol{\xi}_{\boldsymbol{\theta}}$ and  $\small\frac{\partial \boldsymbol{\xi}_{\boldsymbol{\theta}}}{\partial \boldsymbol{\theta}}$  of the original constrained system $\boldsymbol{\Sigma}(\boldsymbol{\theta})$, respectively. Second, notably, assertion (b) also states that the above   approximations can be controlled for arbitrary accuracy by simply letting  $\gamma\rightarrow0$, as  illustrated in the upper panels in Fig. \ref{figthem}. Third, more importantly,  assertion (a) states that the above approximations are always safe in a sense that the  approximation $\boldsymbol{\xi}_{(\boldsymbol{\theta},\gamma)}$ with any small $\gamma>0$ is guaranteed to satisfy all inequality constraints in the original  $\boldsymbol{\Sigma}(\boldsymbol{\theta})$,  as illustrated in the bottom panels in Fig. \ref{figthem}. Finally, similar to  Theorem \ref{theorem1}, assertion (c) states that the  derivative  $\small\frac{\partial \boldsymbol{\xi}_{(\boldsymbol{\theta},\gamma)}}{\partial \boldsymbol{\theta}}$ for  ${\boldsymbol{\Sigma}}(\boldsymbol{\theta},\gamma)$ is a globally unique minimizing trajectory to its corresponding auxiliary control system $\small{\boldsymbol{\overline\Sigma}}\big(\boldsymbol{\xi}_{(\boldsymbol{\theta},\gamma)}\big)$, thus   PDP  \cite{jin2019pontryagin} directly applies here.

In addition to the theoretical importance of Theorem \ref{theorem2}, we also summarize its algorithmic advantage   compared to  directly handling the original constrained  system  $\boldsymbol{\Sigma}(\boldsymbol{\theta})$ and its auxiliary control system $\small\boldsymbol{\overline\Sigma}(\boldsymbol{\xi}_{\boldsymbol{\theta}})$. First, solving the unconstrained  ${\boldsymbol{\Sigma}}(\boldsymbol{\theta},\gamma)$ is  easier than solving the constrained   $\boldsymbol{\Sigma}(\boldsymbol{\theta})$ as more 
 off-the-shelf algorithms are available for unconstrained  trajectory optimization  than for constrained one. Second, when solving $\small\frac{\partial \boldsymbol{\xi}_{(\boldsymbol{\theta},\gamma)}}{\partial \boldsymbol{\theta}}$ using ${\boldsymbol{\overline\Sigma}}\big(\boldsymbol{\xi}_{(\boldsymbol{\theta},\gamma)}\big)$, there is no need to identify the inactive and active inequality constraints, as opposed to solving  $\small\frac{\partial \boldsymbol{\xi}_{\boldsymbol{\theta}}}{\partial \boldsymbol{\theta}}$ using $\small\boldsymbol{\overline\Sigma}(\boldsymbol{\xi}_{\boldsymbol{\theta}})$;   thus it is easier to implement and more numerically stable (we will show this later in experiments). Third,  in contrast to Theorem~\ref{theorem1}, the unconstrained ${\boldsymbol{\Sigma}}(\boldsymbol{\theta},\gamma)$ and  ${\boldsymbol{\overline\Sigma}}\big(\boldsymbol{\xi}_{(\boldsymbol{\theta},\gamma)}\big)$ avoid dealing with the  multipliers   $\boldsymbol{{v}}_{0:T}$ and 
$\boldsymbol{w}_{0:T}$.  Finally, by absorbing  hard inequality constraints  into the control cost  through barrier functions, ${\boldsymbol{\Sigma}}(\boldsymbol{\theta},\gamma)$ introduces the `softness' of constraints  and mitigates the discontinuous `switching' between  inactive/active inequalities over a large range of $\boldsymbol{\theta}$. This leads to a more numerically stable  algorithm,  as we will show in later experiments. Implementation of Theorem \ref{theorem2} is given in Algorithm \ref{alg_theorem2} in Appendix \ref{appendix.algorithm2}.

\section{Safe PDP to Solve Problem \ref{equ_problem}} \label{section.safeP}

\vspace{-5pt}
According to Theorem \ref{theorem2}, we  use the  safe unconstrained  approximation system ${\boldsymbol{\Sigma}}(\boldsymbol{\theta},\gamma)$ in (\ref{equ_oc_penality}) to replace the original inner-level constrained system ${\boldsymbol{\Sigma}}(\boldsymbol{\theta})$ in (\ref{equ_oc}). Then, we  give the following important result for solving Problem \ref{equ_problem}, which addresses the Challenges (4) and (5)  in Section~\ref{section.challenge}.

\begin{theorem}\label{theorem3}
	Consider all functions  defining the constrained optimal control system $\boldsymbol{\Sigma}(\boldsymbol{\theta})$ are at least three-times continuously differentiable,  and let  the conditions (i)-(iii)  in Lemma \ref{result1}  for   differentiability of  $\boldsymbol{\xi}_{\boldsymbol{\theta}}$  hold in a neighborhood of $\boldsymbol{\theta}^*$. Suppose that the second-order condition  for a  local isolated  minimizor  $\boldsymbol{\theta}^*$ to Problem \ref{equ_problem} is satisfied,  that the  gradients  $\nabla_{\boldsymbol{\theta}} R_i(\boldsymbol{\xi}_{\boldsymbol{\theta}^*},\boldsymbol{\theta}^*)$ of   all binding constraints $R_i(\boldsymbol{\xi}_{\boldsymbol{\theta}},\boldsymbol{\theta})=0$ are linearly independent at $\boldsymbol{\theta}^*$, and that  the strict complementary holds at $\boldsymbol{\theta}^*$. Then, for any  small $\epsilon>0$ and any small $\gamma>0$, the following outer-level unconstrained approximation
\begin{equation}\label{equ_problem_penalty_approx}
\boldsymbol{\theta}^*{(\epsilon, \gamma)}= \arg\min_{\boldsymbol{{\theta}}}\,\, \ell\big(\boldsymbol{\xi}_{(\boldsymbol{\theta}, \gamma)},\boldsymbol{\theta}\big)-\epsilon\sum\nolimits_{i=1}^{l}\ln\Big({-}R_i\big(\boldsymbol{\xi}_{(\boldsymbol{\theta}, \gamma)},\boldsymbol{\theta}\big)\Big)
\tag*{SP(${\epsilon},\gamma$)},
\end{equation}
	 with $\boldsymbol{\xi}_{(\boldsymbol{\theta},\gamma)}$ being the optimal trajectory  to the inner-level safe unconstrained approximation system $\boldsymbol{\Sigma}(\boldsymbol{\theta},\gamma)$ in (\ref{equ_oc_penality}), has the following assertions:
	\begin{itemize}
				\setlength\itemsep{-0.05em}
		\item[(a)] there  exists a local isolated minimizor $\boldsymbol{\theta}^*(\epsilon,\gamma)$ to the above   \ref{equ_problem_penalty_approx}, and the corresponding  trajectory $\boldsymbol{\xi}_{
			\left(\boldsymbol{\theta}^*(\epsilon,\gamma),\gamma\right)
			}$ from the inner-level approximation system  $\boldsymbol\Sigma{\big(\boldsymbol{\theta}^*(\epsilon,\gamma),\gamma\big)}$ is  safe with respect the original outer-level constraints,  i.e., $\small R_i\big(\boldsymbol{\xi}_{(
			\boldsymbol{\theta}^*(\epsilon,\gamma),\gamma
			)}, \boldsymbol{\theta}^*(\epsilon,\gamma)\big)<0$, $i=1,2,...,l$;
		\item[(b)] $\boldsymbol{\theta}^*(\epsilon,\gamma)$ is  once-continuously differentiable with respect to both $\epsilon$ and $\gamma$, and 
		\begin{equation}
		\boldsymbol{\theta}^*(\epsilon,\gamma)\rightarrow \boldsymbol{\theta}^* \qquad \text{as} \qquad (\epsilon, \gamma)\rightarrow (0,0);
		\end{equation}
		\item [(c)] for any $\boldsymbol{\theta}$ near  $\boldsymbol{\theta}^*(\epsilon,\gamma)$,   $\boldsymbol{\xi}_{(\boldsymbol{\theta},\gamma)}$ from the inner-level approximation system  $\boldsymbol\Sigma{(\boldsymbol{\theta},\gamma)}$ is safe with respect to the original outer-level constraints, i.e.,  $R_i\big(\boldsymbol{\xi}_{(\boldsymbol{\theta},\gamma)},\boldsymbol{\theta}\big)< 0$,  $i=1,2,...,l$.  
	\end{itemize}
\end{theorem}

The proof of the above theorem is given in Appendix \ref{pf_theorem3}. The above result  says that   instead of solving the original constrained Problem \ref{equ_problem} with the  inner-level constrained system  ${\boldsymbol{\Sigma}}(\boldsymbol{\theta})$ in (\ref{equ_oc}), one can solve an \emph{unconstrained approximation} Problem \ref{equ_problem_penalty_approx} with the inner-level \emph{safe unconstrained  approximation system} ${\boldsymbol{\Sigma}}(\boldsymbol{\theta},\gamma)$ in (\ref{equ_oc_penality}). Particularly, we make the following comments on the importance of the above theorem. 
First, claim (a) affirms that although the inner-level trajectory  $\boldsymbol{\xi}_{(\boldsymbol{\theta},\gamma)}$  is  an approximation (recall  Theorem \ref{theorem2}), the outer-level unconstrained Problem \ref{equ_problem_penalty_approx} always has a locally unique solution  $\boldsymbol{\theta}^*(\epsilon,\gamma)$; furthermore, at $\boldsymbol{\theta}^*(\epsilon,\gamma)$, the corresponding inner-level trajectory   $\boldsymbol{\xi}_{(\boldsymbol{\theta}^*({\epsilon,\gamma}),\gamma)}$ is safe with respect to the original outer-level constraints, i.e., $\small R_i\big(\boldsymbol{\xi}_{(
	\boldsymbol{\theta}^*(\epsilon,\gamma),\gamma
	)}, \boldsymbol{\theta}^*(\epsilon,\gamma)\big)<0$, $i=1,2,...,l$. 
Second, claim (b) asserts that the accuracy of the  solution  $\boldsymbol{\theta}^*(\epsilon,\gamma)$ to the outer-level approximation Problem \ref{equ_problem_penalty_approx} is controlled jointly by  the inner-level barrier parameter $\gamma$ and outer-level barrier parameter $\epsilon$: as both barrier parameters approach zero,  $\boldsymbol{\theta}^*(\epsilon,\gamma)$   is converging to the true solution  $\boldsymbol{\theta}^*$  to the original Problem \ref{equ_problem}. Third, claim (c) says that during the local search of the outer-level  solution $\boldsymbol{\theta}^*(\epsilon,\gamma)$, the corresponding inner-level  trajectory  $\boldsymbol{\xi}_{(\boldsymbol{\theta},\gamma)}$ is always safe with respect to the original outer-level constraints, i.e., $R_i\big(\boldsymbol{\xi}_{(\boldsymbol{\theta},\gamma)},\boldsymbol{\theta}\big)< 0$,  $i=1,2,...,l$.  
The above Theorem \ref{theorem3}, together with Theorem \ref{theorem2} provide the safety- and accuracy- guarantees for the whole  Safe PDP framework.
Then entire Safe PDP algorithm   is given  in Algorithm \ref{alg_theorem3} in Appendix \ref{appendix.algorithm3}.

\vspace{-3pt}

\section{Applications to Different Safety-Critical  Tasks}\label{section.applications}

\vspace{-5pt}

We apply   Safe PDP to solve some representative safety-critical learning/control tasks. For a specific task, one only needs to specify the  parameterization detail of  $\boldsymbol{\Sigma}(\boldsymbol{\theta})$, a task loss  $\ell(\boldsymbol{\xi}_{\boldsymbol{\theta}},\boldsymbol{\theta})$, and task    constraints $R_i(\boldsymbol{\xi}_{\boldsymbol{\theta}},\boldsymbol{\theta})$ in Problem \ref{equ_problem}. The  experiments  are performed on the  systems of different complexities in Table \ref{experimenttable}. All codes are available at \url{https://github.com/wanxinjin/Safe-PDP}.

\vspace{-10pt}
\begin{table}[h]
	\captionsetup[table]{skip=-5pt}  
	\caption{Experimental environments \cite{jin2019pontryagin} }
	\label{experimenttable}
	\centering
	\begin{tabular}{llll}
		\toprule
		\small System $\boldsymbol{\Sigma}(\boldsymbol{\theta})$    & \small Dynamics  $\small\boldsymbol{f}(\boldsymbol{\theta}_\text{dyn})$    & \small Control cost  $\small J(\boldsymbol{\theta}_{\text{obj}})$ & \small Constraints  $\small \boldsymbol{g}(\boldsymbol{\theta}_\text{cstr})$   \\[-2pt]
		\midrule
		\small Cartpole     & \small cart \& pole masses and  length & \small \multirow{4}{*}{
			\makecell{
				$
				c_t{=}\norm{\boldsymbol{u}}_2^2{+}$\\
				$\norm{\boldsymbol{\theta}_{\text{obj}}^\prime(\boldsymbol{x}{-}\boldsymbol{x}_{\text{goal}})}_2^2$,\\[3pt]
				$
				c_T{=}$\\
				$\norm{\boldsymbol{\theta}_{\text{obj}}^\prime(\boldsymbol{x}{-}\boldsymbol{x}_{\text{goal}})}_2^2$		
			}
		} & \small \multirow{4}{*}{
			\makecell{
				$
			\boldsymbol{g}_x(\boldsymbol{x}){\leq} X_{\max} $,\\[2.5pt] $
				\norm{\boldsymbol{u}}_{2/\infty}\leq U_{\max}$,\\[5pt]
				$\boldsymbol{\theta}_{\text{cstr}}{=}\{X_{\max},U_{\max}\}$\\
		} } \\
		
		\small Two-link Robot arm     & \small  length and mass  of links   & & \\
		\small 6-DoF  quadrotor & \small mass, wing length, inertia  & & \\
		\small 6-DoF rocket  landing & \small rocket mass,  length, inertia  & & \\[1pt]
		\bottomrule
	\end{tabular}
	
	\vspace{1pt}
	{\raggedright \small{Note that for each system,  $\boldsymbol{g}({\boldsymbol{\theta}_{\text{cstr}}})$ includes the    immediate constraints   on   system input $\boldsymbol{u}$ and state  $\boldsymbol{x}$ at any time instance;  $\boldsymbol{g}_x$ is known;  $\norm{\cdot}_{2/\infty}$ is the $2$ or $\infty$ norm; and time horizon $T$  is around $50$ for all systems. } \par}
\end{table}

\vspace{-2pt}

\textbf{Problem I: Safe Policy Optimization}  aims to find a  policy that minimizes a control cost $J$ subject to  constraints $\boldsymbol{g}$ while guaranteeing that \emph{any intermediate policy during optimization  should never violate  the constraints}. To apply Safe PDP to solve such a problem for the systems in Table \ref{experimenttable}, we set:

\begin{longfbox}[padding-top=-4pt,margin-top=-6pt, padding-bottom=-3pt, margin-bottom=-10pt]
		\mathleft
		\begin{equation}\label{sys.opt}
		\boldsymbol{\Sigma}(\boldsymbol{\theta}):\quad\qquad\qquad\qquad
		\begin{aligned}
		\emph{dynamics:} &\quad \boldsymbol{x}_{t+1}=\boldsymbol{f}(\boldsymbol{x}_{t},\boldsymbol{u}_{t}) \quad \text{with}\quad \boldsymbol{x}_{0}, \\
		\emph{policy:}& \quad
		\boldsymbol{u}_t=\boldsymbol{\pi}_t( \boldsymbol{x}_t,\boldsymbol{\theta}),
		\end{aligned}
		\end{equation}	\mathcenter
\end{longfbox}
where  dynamics $\boldsymbol{f}$ is learned from demonstrations in Problem III, and  $\boldsymbol{\pi}(\boldsymbol{\theta})$ is represented by a (deep) feedforward neural network (NN) with $\boldsymbol{\theta}$ the NN parameter. In Problem \ref{equ_problem}, the task loss $\ell(\boldsymbol{\xi}_{\boldsymbol{\theta}},\boldsymbol{\theta})$ is set as   $J(\boldsymbol{\theta}_{\text{obj}})$, and task constraints  $R_i(\boldsymbol{\xi}_{\boldsymbol{\theta}},\boldsymbol{\theta})$ as $\boldsymbol{g}(\boldsymbol{\theta}_\text{cstr})$, with both $\boldsymbol{\theta}_{\text{obj}}$ and $\boldsymbol{\theta}_\text{cstr}$ known. Then, safe policy optimization is to solve Problem \ref{equ_problem} using Safe PDP.   The results for the  robot arm and 6-DoF maneuvering quadrotor are  in Fig. \ref{figspo}, and the other  results and details are  in Appendix \ref{appendix.application.spo}.

\vspace{-8pt}
\begin{figure} [h]
	\begin{subfigure}{.182\textwidth}
		\centering
		\includegraphics[width=\linewidth]{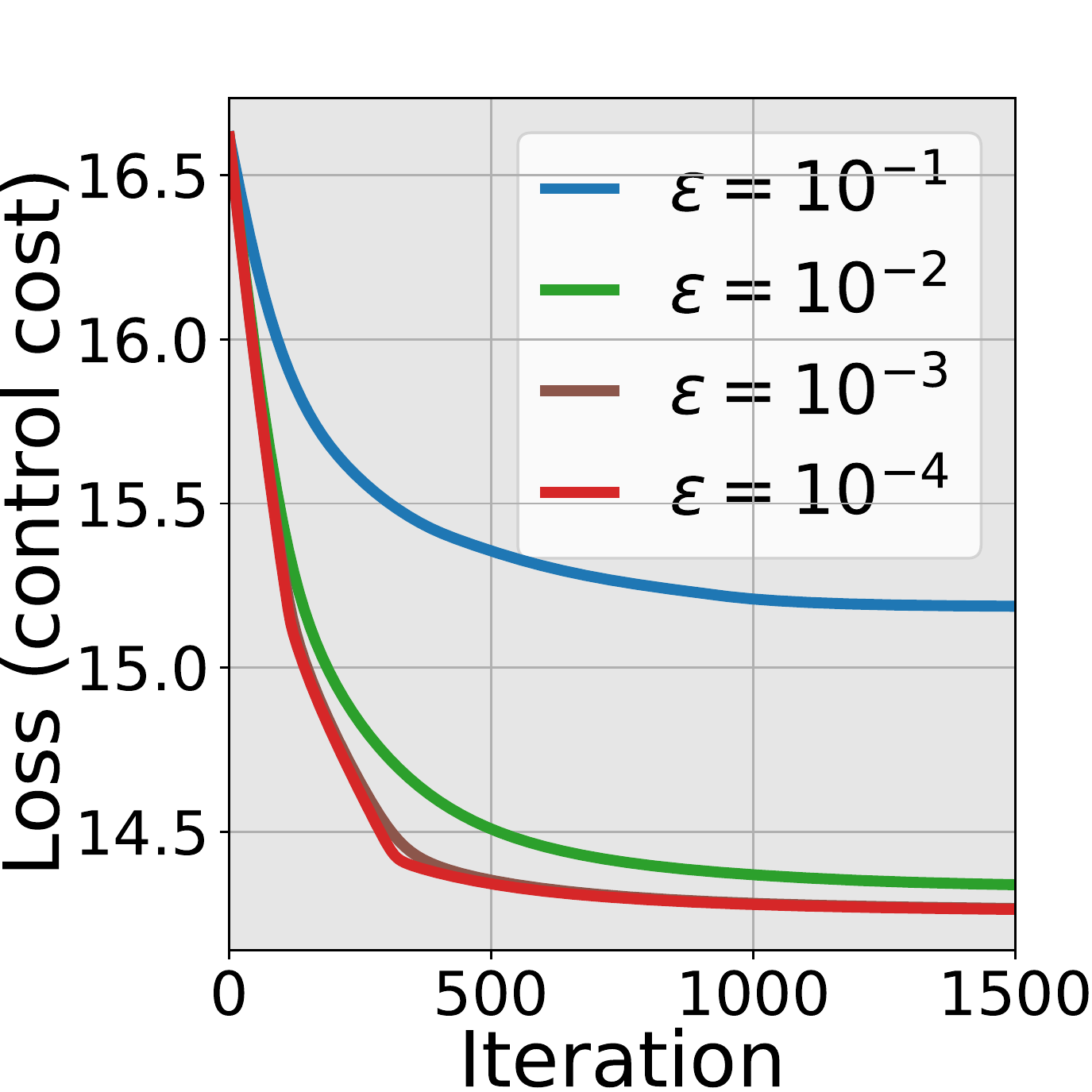}
		\caption{Robot-arm loss}
		\label{figspo.1}
	\end{subfigure}
	\begin{subfigure}{.31\textwidth}
		\centering
		\includegraphics[width=\linewidth]{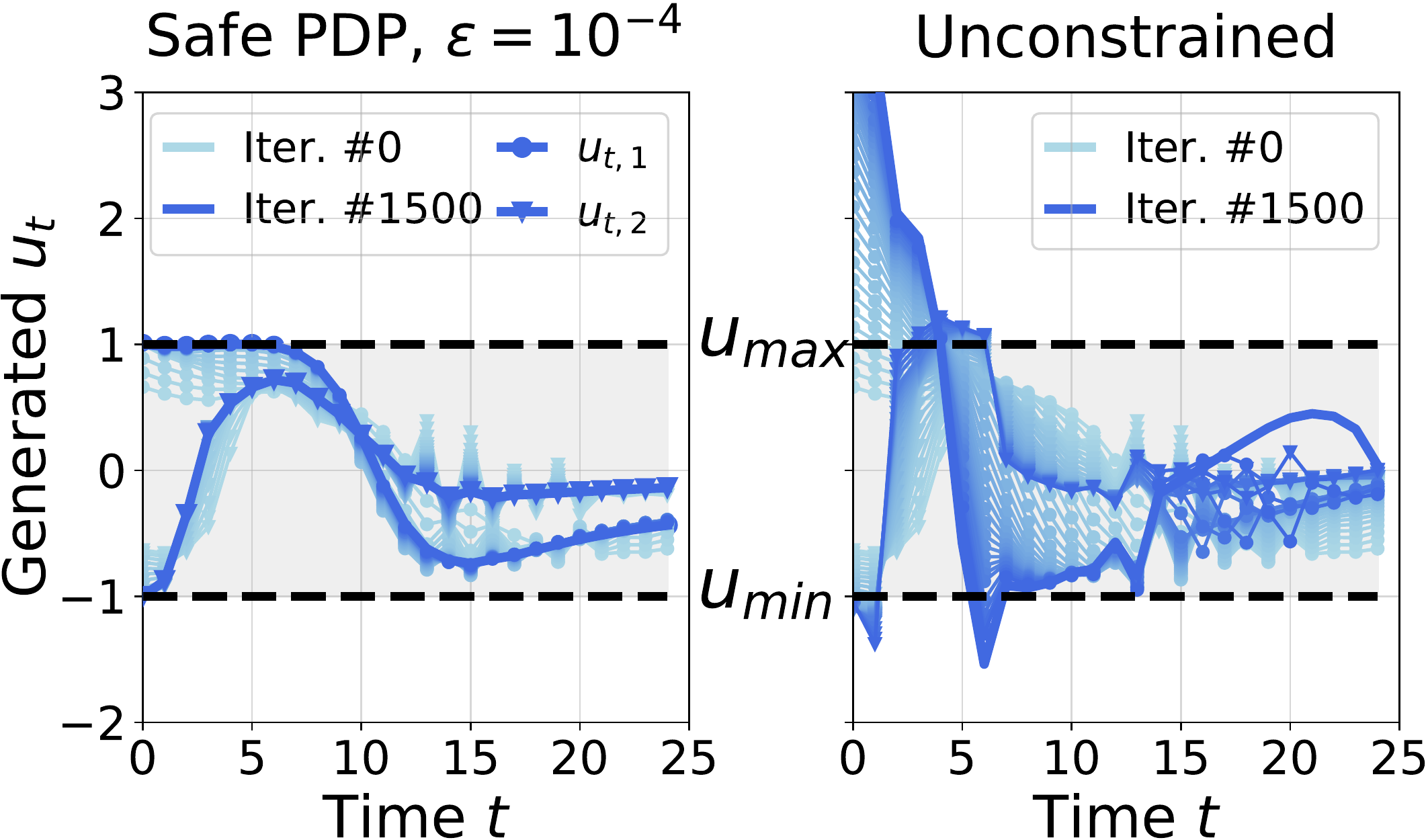}
		\caption{Constraint violation during opt.}
		\label{figspo.2}
	\end{subfigure} 
	\begin{subfigure}{.182\textwidth}
		\centering
		\includegraphics[width=\linewidth]{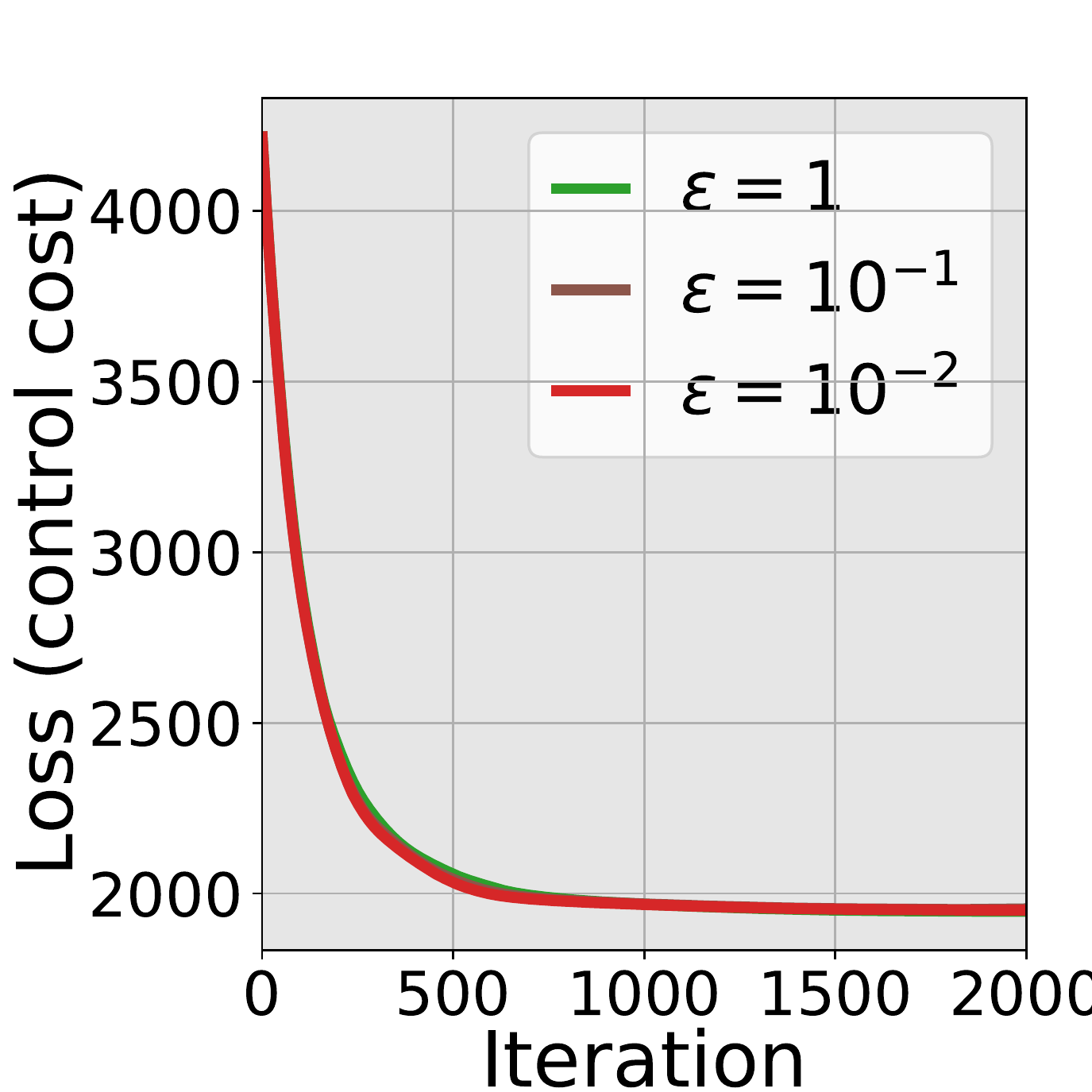}
		\caption{Quadrotor loss}
		\label{figspo.3}
		
	\end{subfigure}
	\begin{subfigure}{.31\textwidth}
		\centering
		\includegraphics[width=\linewidth]{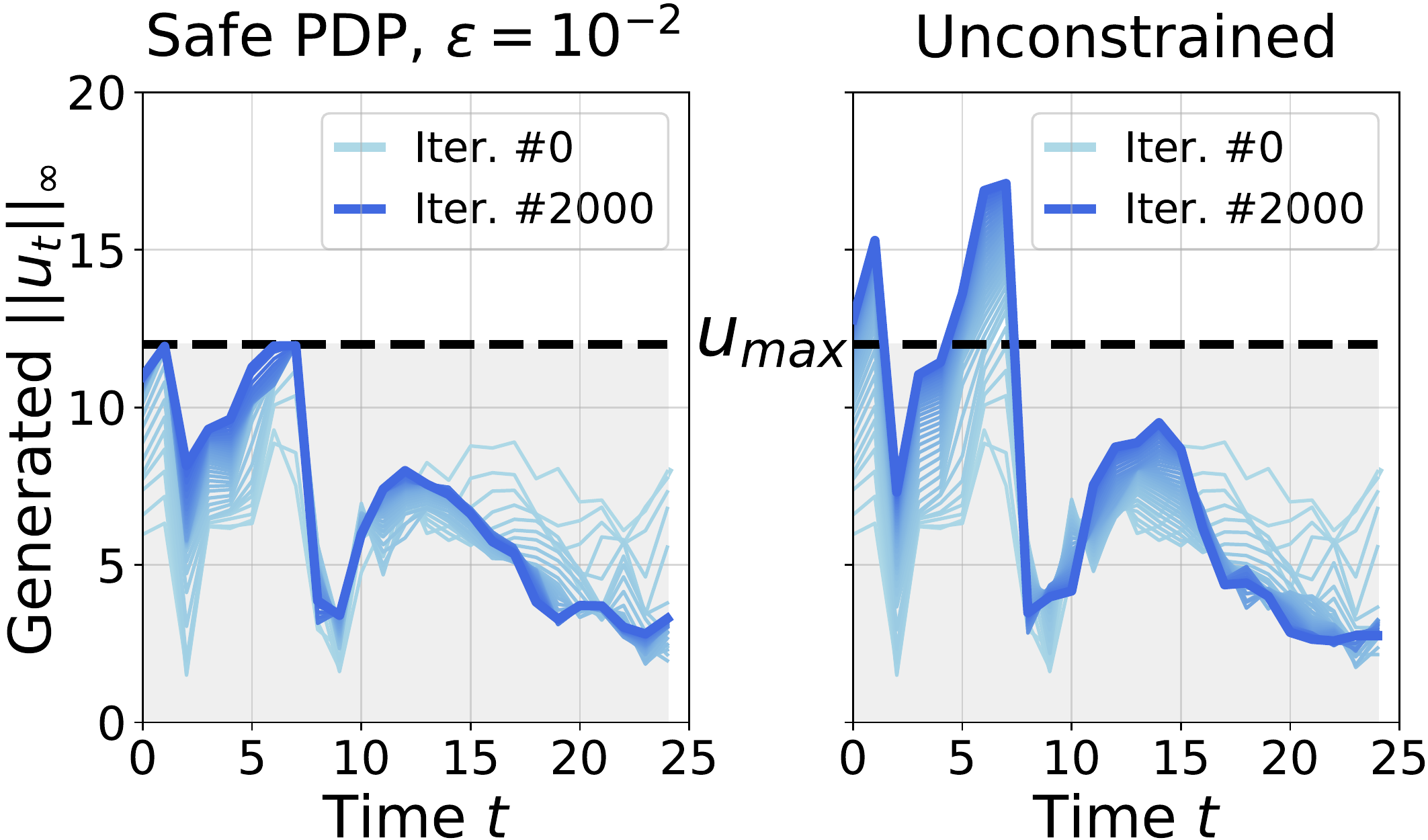}
		\caption{Constraint violation during opt.}
		\label{figspo.4}
	\end{subfigure}
	\caption{Safe neural policy optimization for robot-arm (a)-(b) and 6-DoF quadrotor (c)-(d).}
	\label{figspo}
	\vspace{-13pt}
\end{figure}

Fig. \ref{figspo.1} and \ref{figspo.3} plot    loss (control cost) versus gradient-descent iteration under different  $\epsilon$, showing that the NN policy achieves a good convergence when  $\epsilon\leq 10^{{-}2}$ (as asserted by Theorem \ref{theorem3}).  Fig. \ref{figspo.2} and \ref{figspo.4} show all indeterminate control trajectories generated from the NN policy during entire iterations; we  also mark the  constraints $U_{\max}$ and compare  with the unconstrained policy optimization under the same settings. The results confirm that  Safe PDP enables \emph{to achieve an optimal  policy while guaranteeing that any intermediate policy throughout optimization is safe}.

\textbf{Problem II: Safe Motion Planning} searches for a dynamics-feasible trajectory that optimizes a  criterion and avoids unsafe regions (obstacles), meanwhile  guaranteeing that any intermediate  motion trajectory during search must  avoid the unsafe regions. To apply  Safe PDP to solve such  problem, we specialize $	\boldsymbol{\Sigma}(\boldsymbol{\theta})$ as (\ref{sys.opt}) except that   policy here is $\boldsymbol{u}_t=\boldsymbol{u}(t,\boldsymbol{\theta})$, which is represented by  Lagrangian polynomial \cite{abramowitz1964handbook} with $\boldsymbol{\theta}$  the parameters (pivots). In Problem \ref{equ_problem},   task loss  is set as    $J(\boldsymbol{\theta}_{\text{obj}})$, and task constraints as $\boldsymbol{g}(\boldsymbol{\theta}_\text{cstr})$, with  $\boldsymbol{\theta}_{\text{obj}}$ and $\boldsymbol{\theta}_\text{cstr}$ known in Table \ref{experimenttable}.  The safe planning  results using Safe PDP for  cartpole and 6-DoF rocket landing   are  in Fig. \ref{figspo}, in comparison with ALTRO, a state-of-the-art constrained trajectory optimization method \cite{howell2019altro}. Other results and more  details are in Appendix \ref{appendix.application.splan}.

\vspace{-5pt}
\begin{figure} [h]
	\begin{subfigure}{.178\textwidth}
		\centering
		\includegraphics[width=\linewidth]{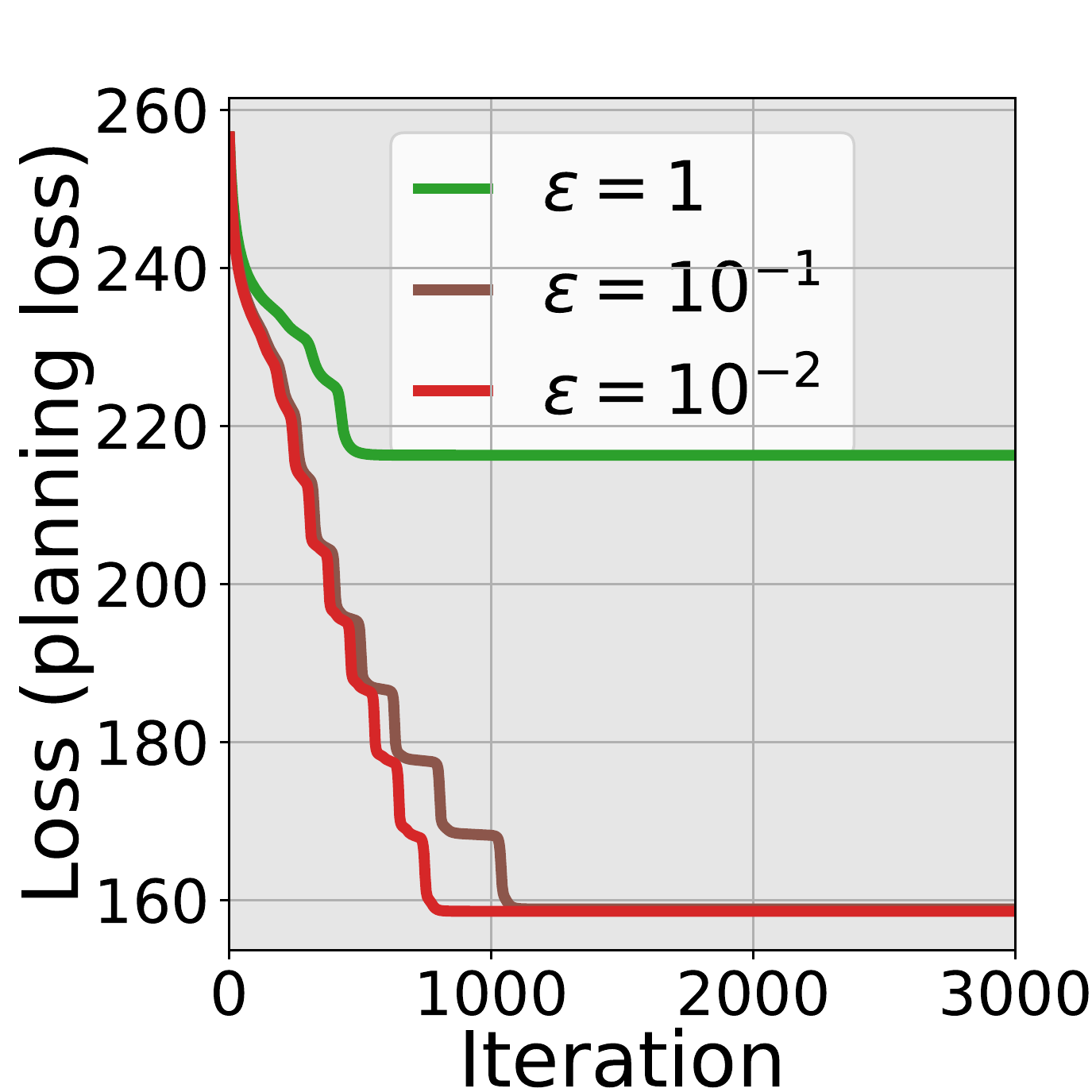}
		\caption{Cartpole loss}
		\label{figsplan.1}
	\end{subfigure}
	\begin{subfigure}{.31\textwidth}
		\centering
		\includegraphics[width=\linewidth]{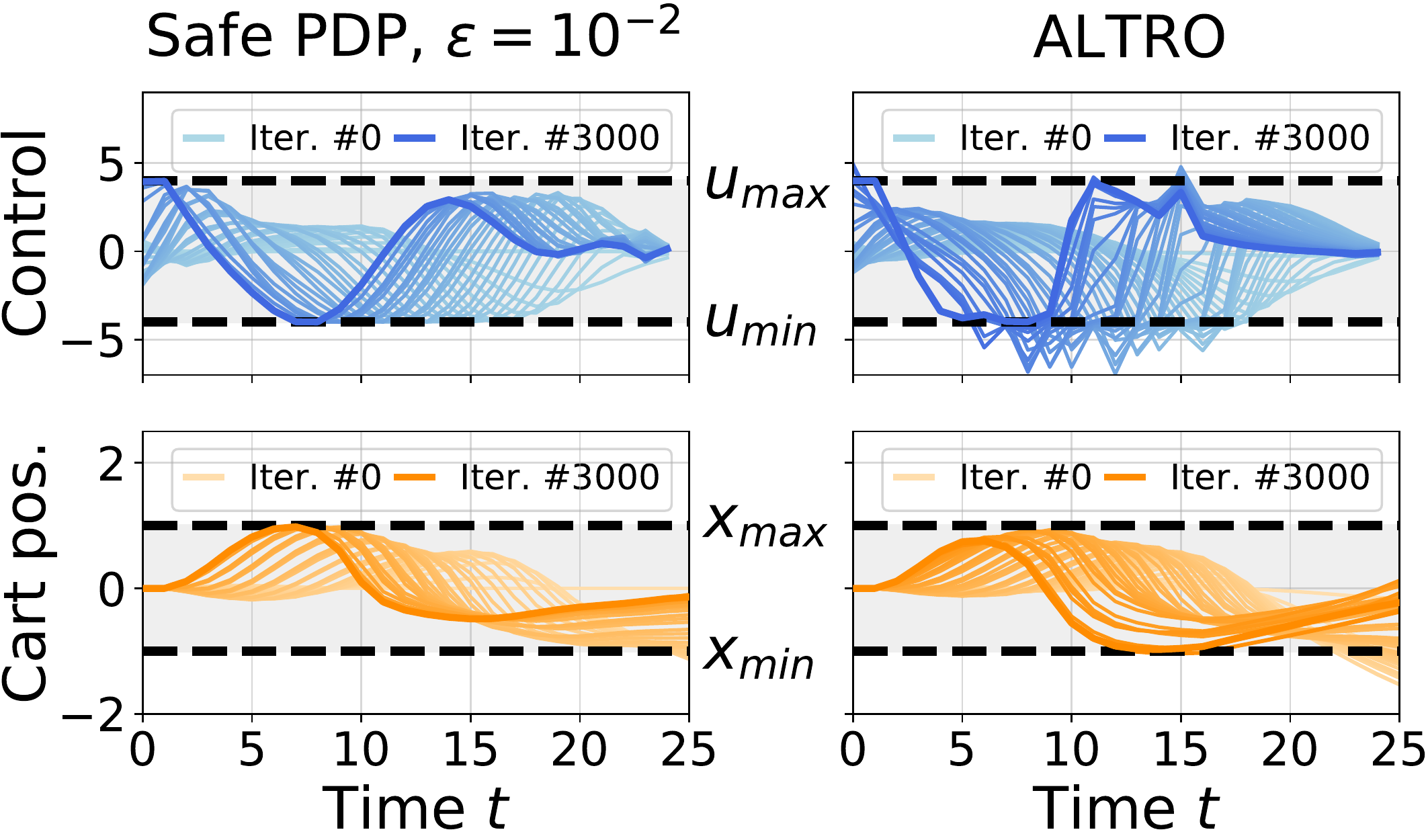}
		\caption{Constraint violation during opt.}
		\label{figsplan.2}
	\end{subfigure} 
	\hspace{0.5pt}
	\begin{subfigure}{.178\textwidth}
		\centering
		\includegraphics[width=\linewidth]{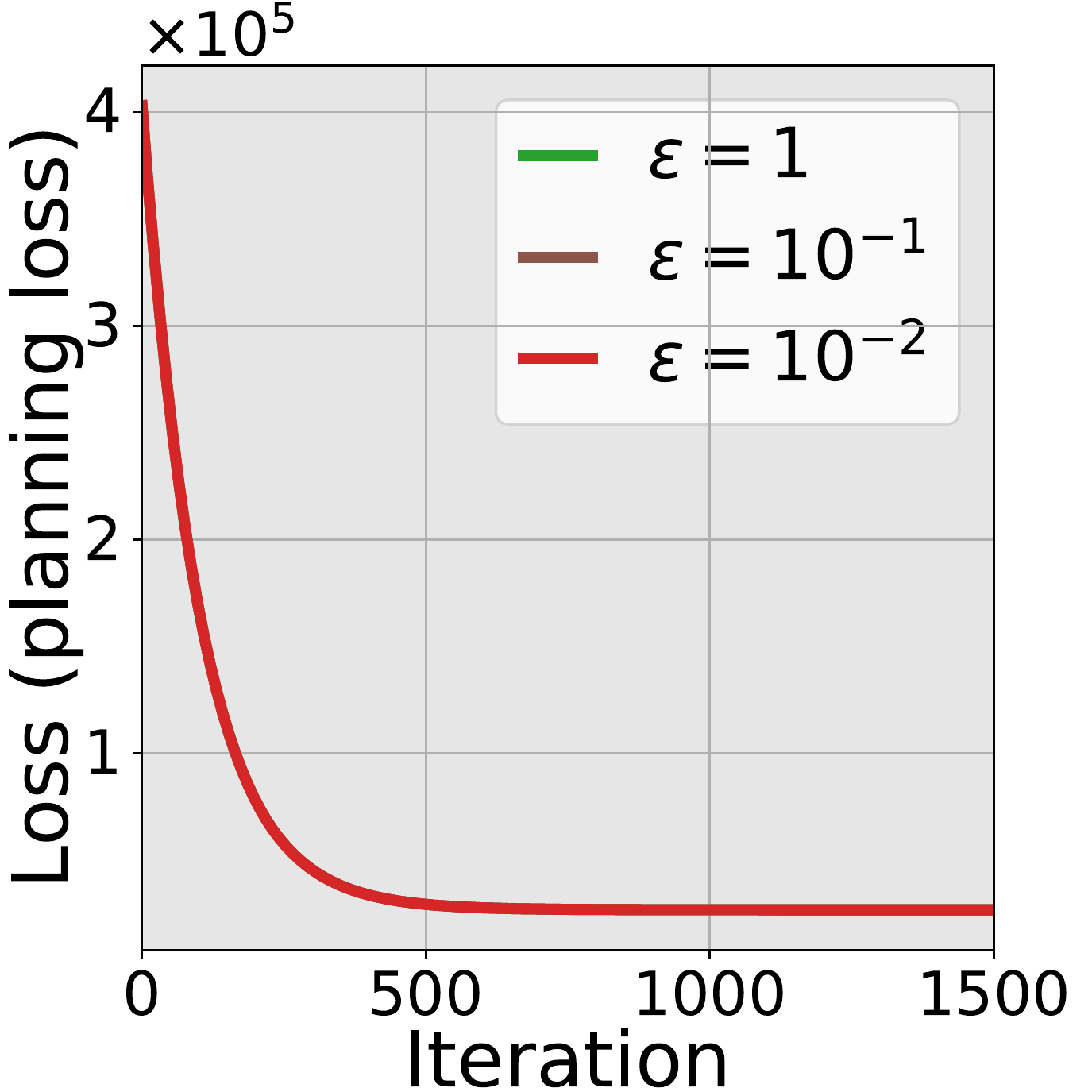}
		\caption{Rocket loss}
		\label{figsplan.3}
		
	\end{subfigure}
	\begin{subfigure}{.31\textwidth}
		\centering
		\includegraphics[width=\linewidth]{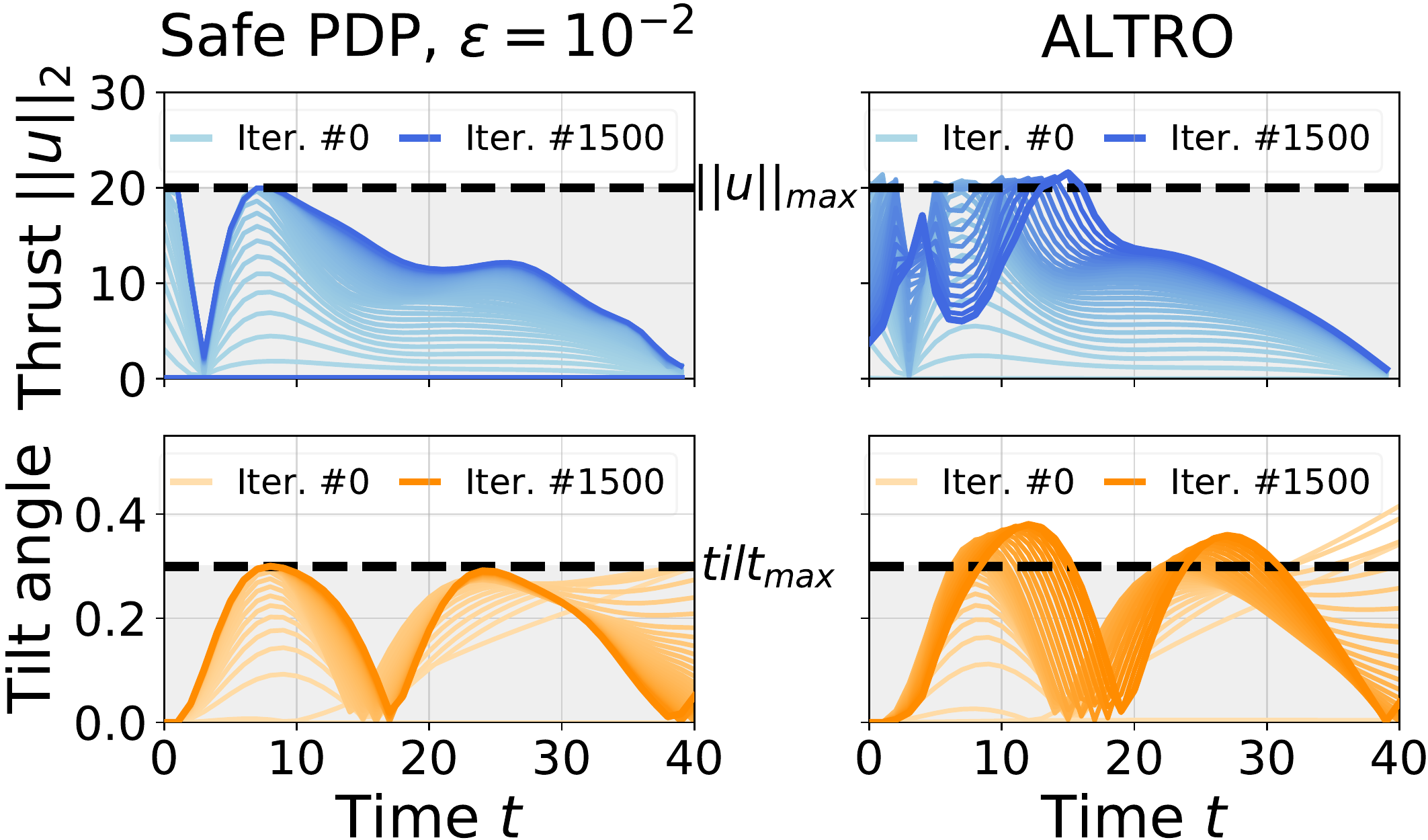}
		\caption{Constraint violation during opt.}
		\label{figsplan.4}
	\end{subfigure}
	\caption{Safe motion planning for cartpole (a)-(b) and 6-DoF rocket powered landing (c)-(d).}
	\label{figsplan}
\end{figure}
\vspace{-6pt}

Fig. \ref{figsplan.1} and \ref{figsplan.3} plot the task loss  versus gradient-descent iteration, showing that the trajectory achieves a good convergence with  $\epsilon\leq 10^{{-}2}$.  Fig. \ref{figsplan.2} and \ref{figsplan.4} show all intermediate motion trajectories during entire optimization, with  constraints marked. The results confirm that Safe PDP can find an optimal trajectory while always respecting constraints throughout planning process.

\textbf{Problem III: Learning MPC from Demonstrations.} Suppose  for all systems in Table \ref{experimenttable},   the control cost $J(\boldsymbol{\theta}_{\text{cost}})$,   dynamics $\boldsymbol{f}(\boldsymbol{\theta}_{\text{dyn}})$ and constraints $\boldsymbol{g}_t(\boldsymbol{\theta}_{\text{cstr}})$ are all  unknown and parameterized as in Table \ref{experimenttable}. We aim to jointly learn $\boldsymbol{\theta}=\{\boldsymbol{\theta}_{\text{cost}}, \boldsymbol{\theta}_{\text{dyn}}, \boldsymbol{\theta}_{\text{cstr}}\}$ from  demonstrations  $\small\boldsymbol{\xi}^\text{demo}=\{\boldsymbol{{x}}_{0:T}^\text{demo}, \boldsymbol{u}_{0:T-1}^{\text{demo}}\}$ of a true expert system.  In Problem \ref{equ_problem}, set $\small\boldsymbol{\Sigma}(\boldsymbol{\theta})$  as (\ref{equ_oc}), consisting of  $J(\boldsymbol{\theta}_{\text{cost}})$,    $\boldsymbol{f}(\boldsymbol{\theta}_{\text{dyn}})$, and  $\boldsymbol{g}_t(\boldsymbol{\theta}_{\text{cstr}})$ parameterized; set  task loss
$
\ell(\boldsymbol{\xi}_{\boldsymbol{\theta}}, \boldsymbol{\theta})=\norm{\boldsymbol{\xi}^\text{demo}{-} \boldsymbol{\xi}_{\boldsymbol{\theta}}}^2$, which quantifies the \emph{reproducing loss} between  $\boldsymbol{\xi}^{\text{demo}}$  and $\boldsymbol{\xi}_{\boldsymbol{\theta}}$; and there is no task constraints. By solving Problem \ref{equ_problem}, we can learn $\boldsymbol{\Sigma}(\boldsymbol{\theta})$ such that its reproduced   $\boldsymbol{\xi}_{\theta}$ is closest to given $\boldsymbol{\xi}^\text{demo}
$. The  demonstrations $\boldsymbol{\xi}^\text{demo}
$ here  are generated with $\boldsymbol{\theta}$ known (two episode trajectories for each system with time horizon $T{=}50$). The plots of the loss versus gradient-descent iteration  are  in Fig. \ref{figsioc}, and more details and results  are in Appendix \ref{appendix.application.mpc}.

In Fig. \ref{figsioc.1}-\ref{figsioc.4}, for each system, we use three strategies to obtain $ \boldsymbol{\xi}_{\boldsymbol{\theta}}$ and $\small\frac{\partial \boldsymbol{\xi}_{\boldsymbol{\theta}}}{\partial \boldsymbol{\theta}}$ for $\boldsymbol{\Sigma}(\boldsymbol{\theta})$: (A) use a solver \cite{andersson2019casadi} to obtain $ \boldsymbol{\xi}_{\boldsymbol{\theta}}$ and use Theorem \ref{theorem1} to obtain  $\small\frac{\partial \boldsymbol{\xi}_{\boldsymbol{\theta}}}{\partial \boldsymbol{\theta}}$;  (B) use Theorem \ref{theorem2} to approximate  both  $ \small\boldsymbol{\xi}_{\boldsymbol{\theta}}$ and $\small\frac{\partial \boldsymbol{\xi}_{\boldsymbol{\theta}}}{\partial \boldsymbol{\theta}}$ by  $\small \boldsymbol{\xi}_{(\boldsymbol{\theta},\gamma)}$ and $\small\frac{\partial \boldsymbol{\xi}_{(\boldsymbol{\theta}, \gamma)}}{\partial \boldsymbol{\theta}}$, respectively, $\gamma{=}10^{{-}2}$; and  (C) use a solver to  obtain $ \boldsymbol{\xi}_{\boldsymbol{\theta}}$ and Theorem~\ref{theorem2} only for  $\small\frac{\partial \boldsymbol{\xi}_{\boldsymbol{\theta}}}{\partial \boldsymbol{\theta}}$. Fig. \ref{figsioc.1}-\ref{figsioc.4} show that for Strategies (B) and (C),
 the reproducing loss quickly converges to zeros, indicating that the  dynamics, constraints, and control cost are successfully learned to reproduce the  demonstrations. Fig. \ref{figsioc.1}-\ref{figsioc.4}  also show  numerical instability for strategy (A); this is due to the discontinuous `switching' of  active inequalities between iterations, and also the error in correctly identifying   active inequalities  (we identify them by checking  $g_{t,i}>-\delta$ with  $\delta>0$ a small threshold), as analyzed in Section \ref{keysection}. More analysis is given in Appendix \ref{appendix.application.mpc}. Note that we are not aware of any existing methods that can handle jointly learning of cost, dynamics, and constraints here, and thus we have not given benchmark comparison. 
 Fig. \ref{figsioc.5} gives  timing results of Safe PDP.

\vspace{-8pt}
\begin{figure} [h]
	\begin{subfigure}{.19\textwidth}
		\centering
		\includegraphics[width=\linewidth]{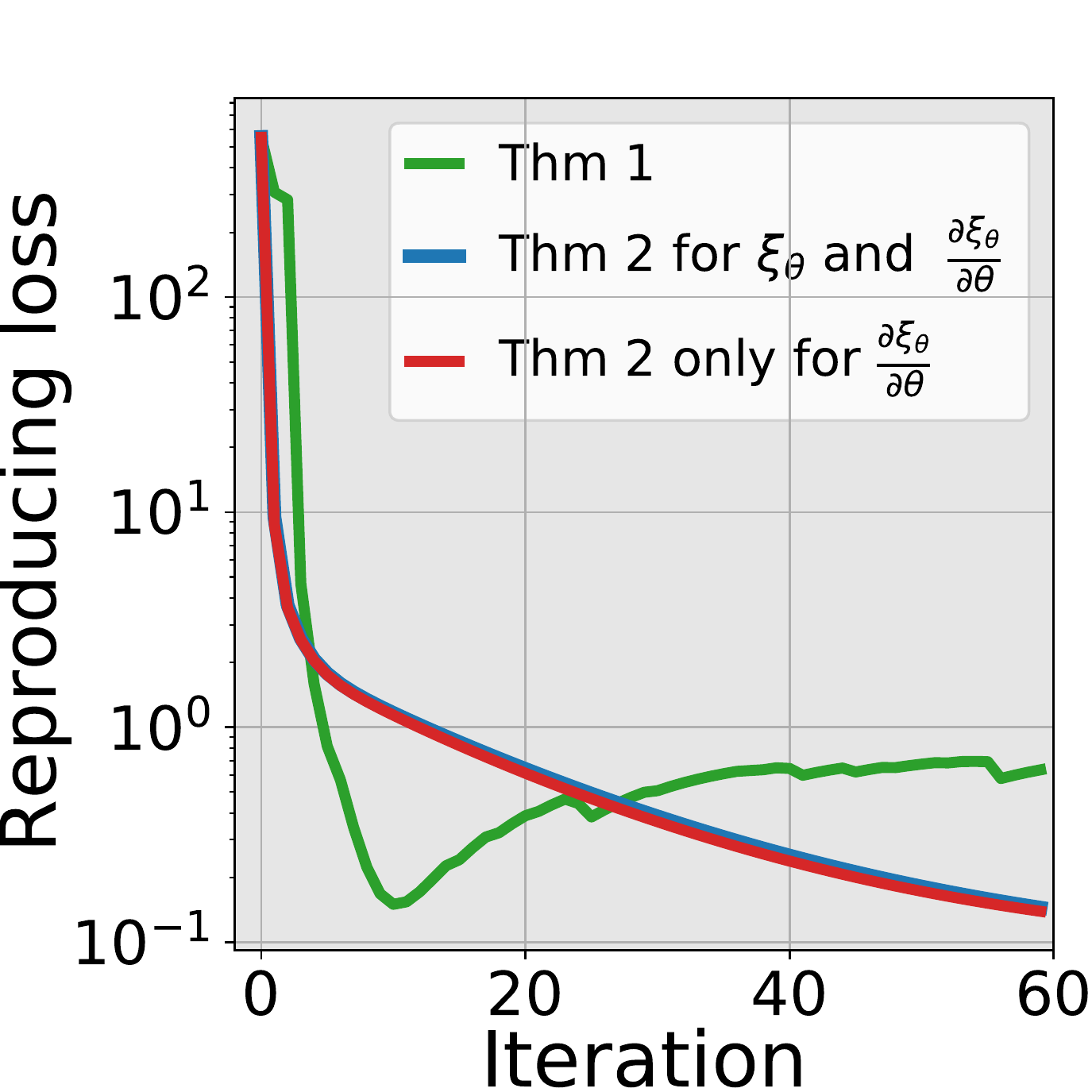}
		\caption{Cartpole (log-y)}
		\label{figsioc.1}
	\end{subfigure}
	\begin{subfigure}{.19\textwidth}
		\centering
		\includegraphics[width=\linewidth]{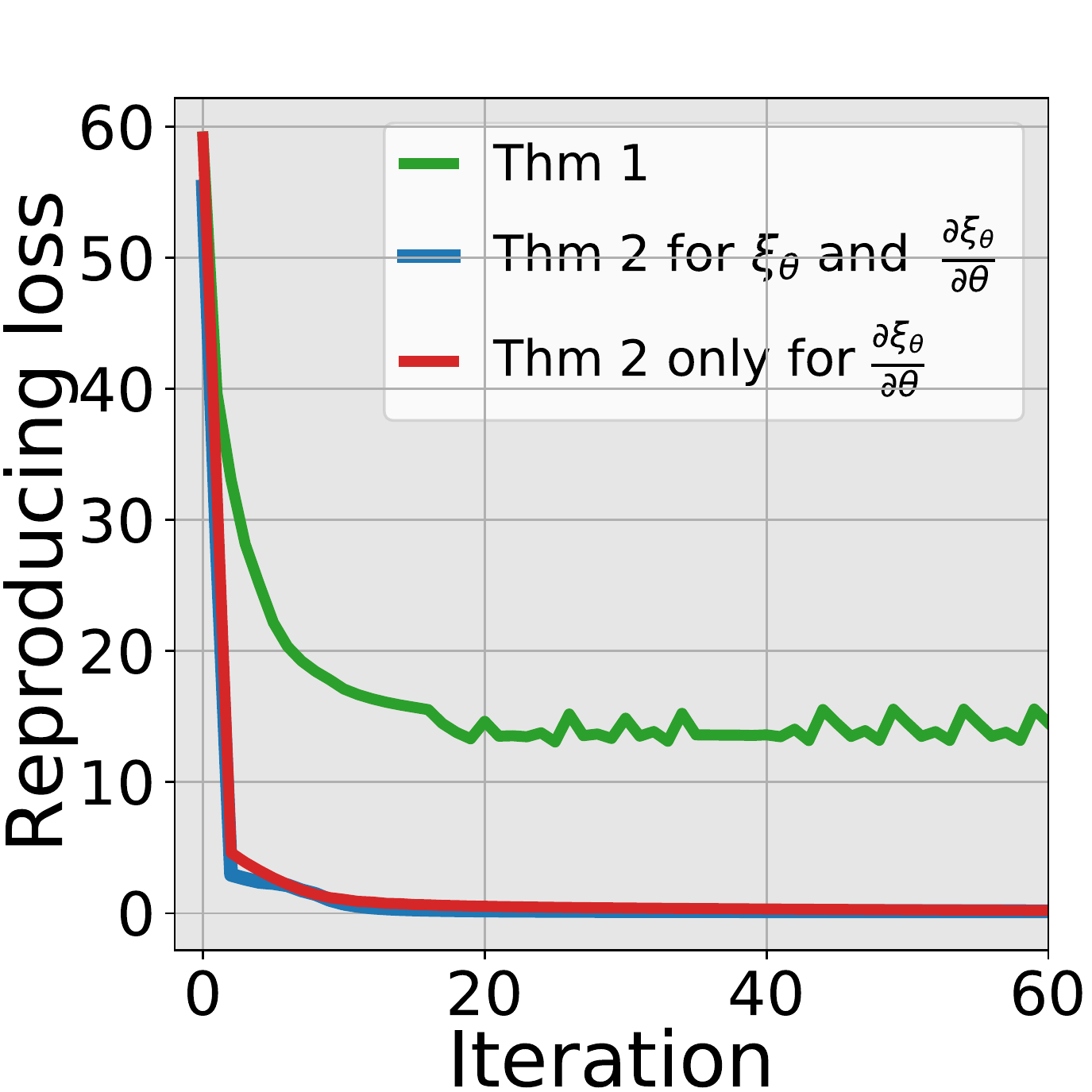}
		\caption{Robot-arm}
		\label{figsioc.2}
	\end{subfigure}
	\begin{subfigure}{.19\textwidth}
		\centering
		\includegraphics[width=\linewidth]{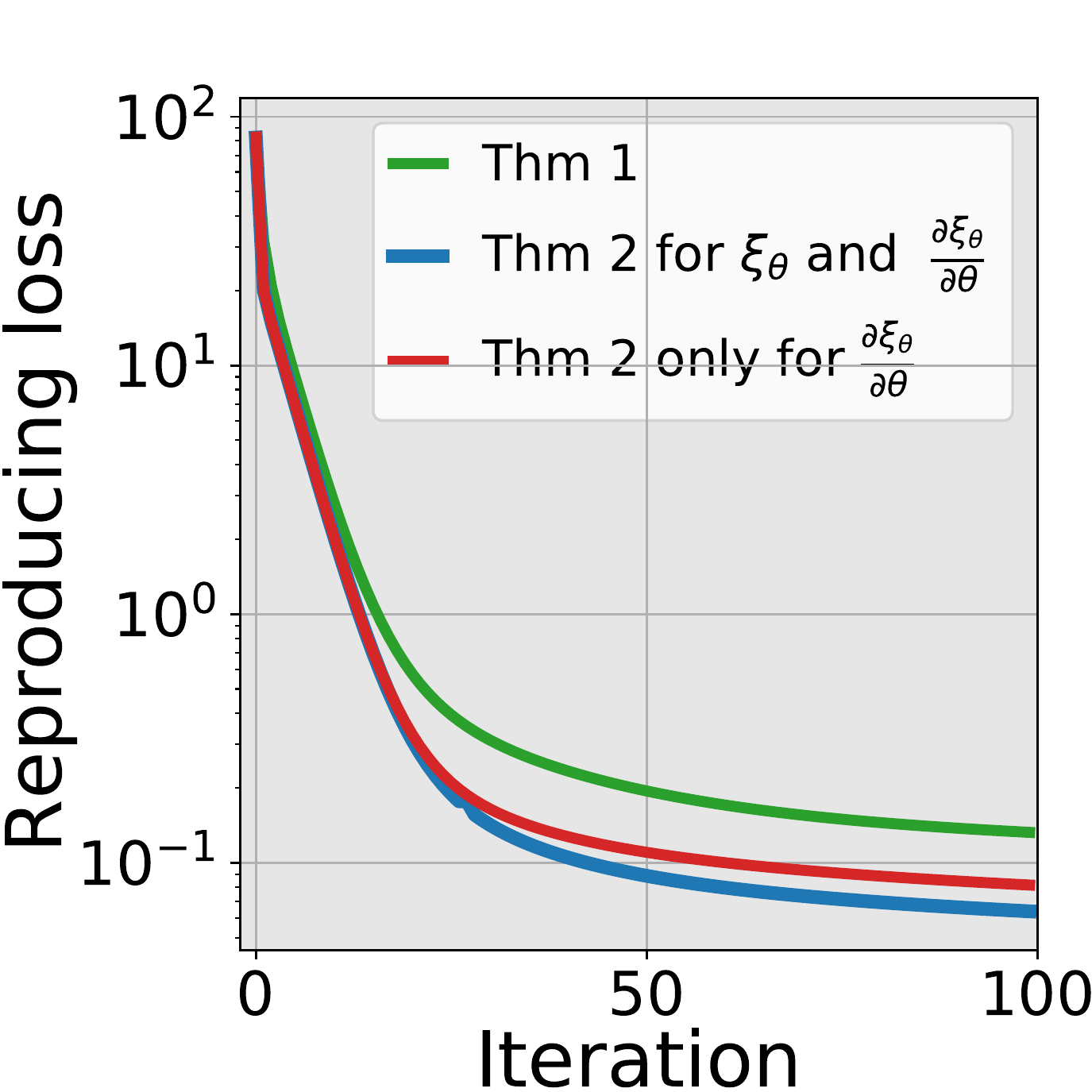}
		\caption{Quadrotor (log-y)}
		\label{figsioc.3}
	\end{subfigure}
	\begin{subfigure}{.19\textwidth}
		\centering
		\includegraphics[width=\linewidth]{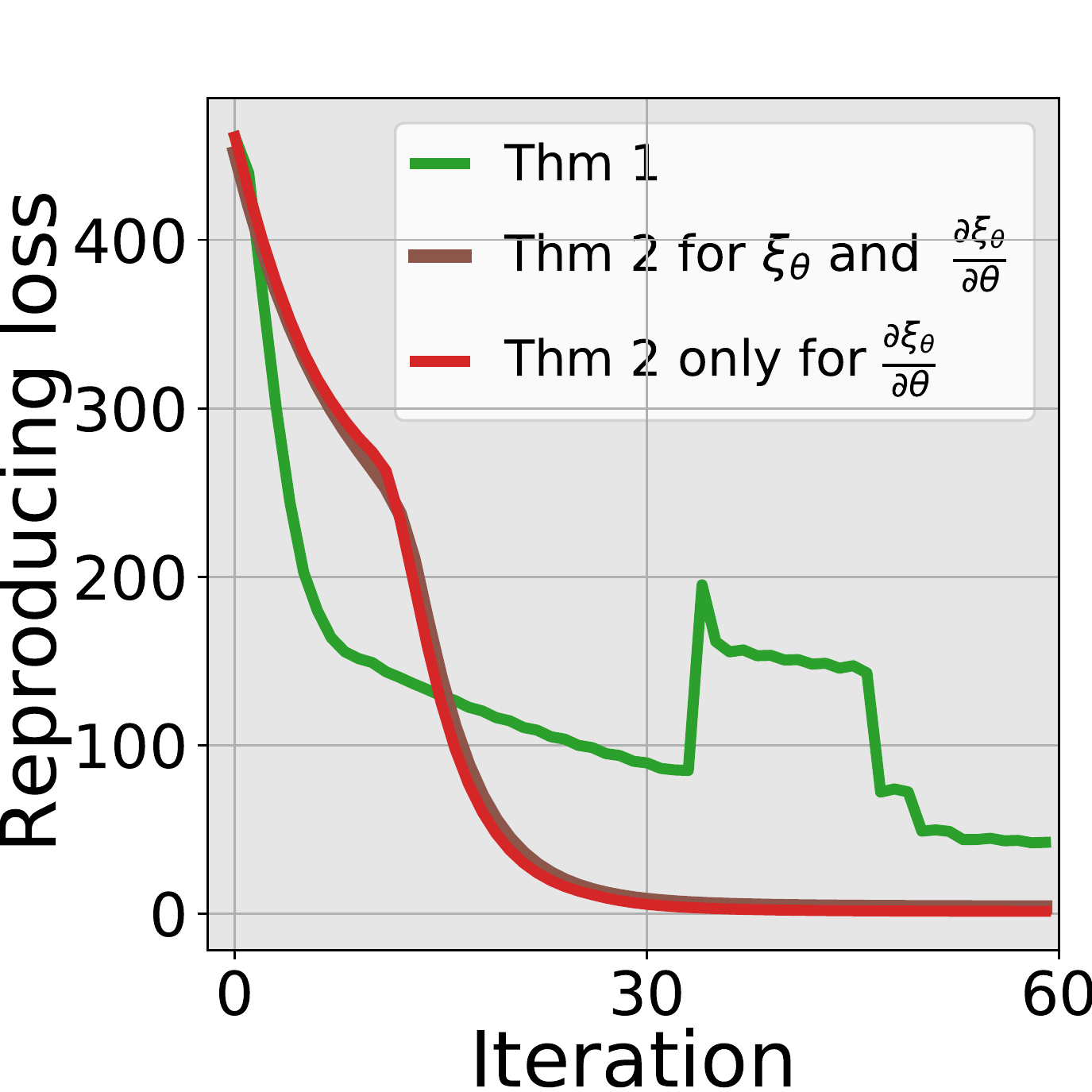}
		\caption{Rocket}
		\label{figsioc.4}
	\end{subfigure}
	\begin{subfigure}{.19\textwidth}
	\centering
	\includegraphics[width=\linewidth]{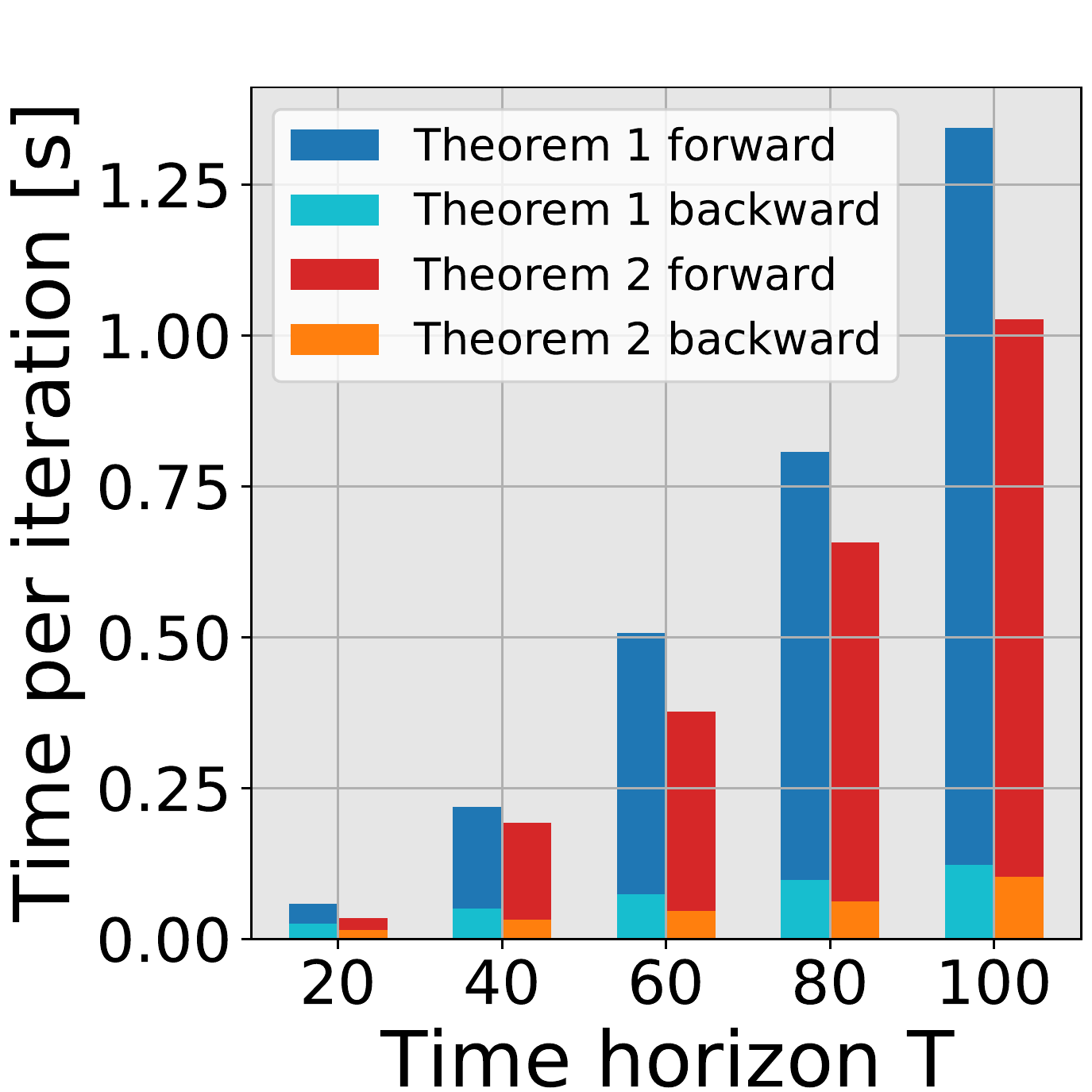}
	\caption{Timing}
	\label{figsioc.5}
\end{subfigure}
	\caption{Jointly learning dynamics,  constraints, and control cost from demonstrations.}
	\label{figsioc}
\end{figure}

\vspace{-8pt}

\section{Discussion}\label{discussion.section}

\vspace{-5pt}

\begin{wrapfigure}[11]{r}{0pt}
	\raisebox{0pt}[\dimexpr\height-1.2\baselineskip]{\includegraphics[width=3.0cm]{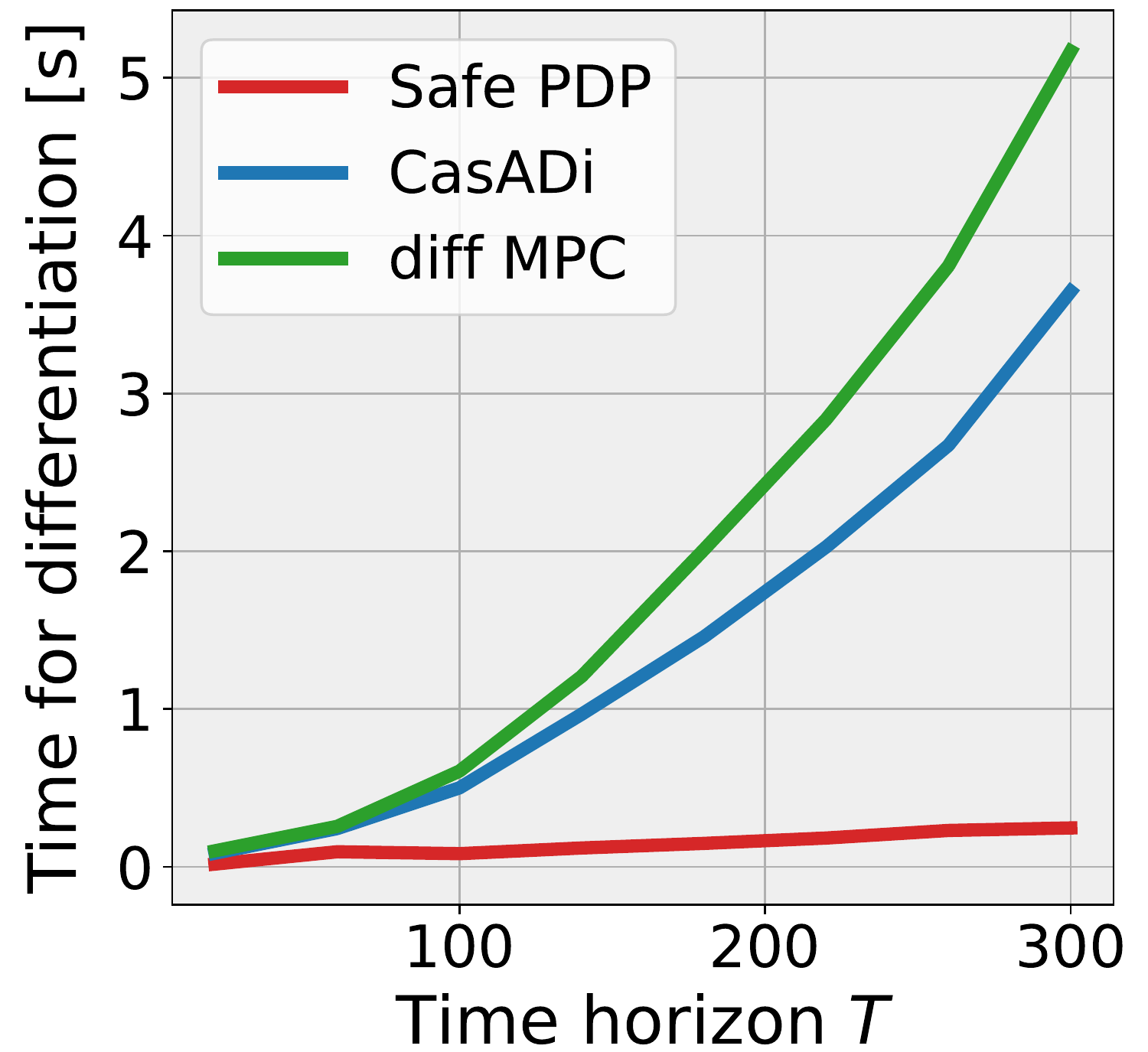}}
	\vspace{-15pt}
	\caption{\small Time for differentiating  optimal trajectory with different  $T$.}
	\label{figure.backwardtime}
\end{wrapfigure}
\textbf{Comparisons with other differentiable frameworks.} \,\, Fig. \ref{figure.backwardtime} compares Safe PDP, CasADi \cite{andersson2018sensitivity}, and Differentiable MPC \cite{amos2018differentiable} for the computational efficiency of differentiating an optimal trajectory of a constrained optimal control system with different control horizons $T$. The results  show a significantly  computational advantage of Safe PDP over CasADi and Differentiable MPC. Specifically, Safe PDP has a complexity of $\mathcal{O }(T)$, while
 CasADi and Differentiable MPC have at least $\mathcal{O }(T^2)$. This is because both CasADi and differentiable MPC are based on   the implicit function theorem \cite{rudin1976principles} and 
 need to compute the inverse of a  Hessian matrix of the size proportional to $T\times T$. In contrast, Safe PDP solves the gradient of a trajectory by constructing an Auxiliary
Control System, which can be  solved using the Riccati equation.

\textbf{Limitation of Safe PDP.} \,\,
Safe PDP requires a safe (feasible) initialization such that the log-barrier control cost or loss  is well-defined. While restrictive,  safe initialization is  common in safe learning \cite{usmanova2020safe,berkenkamp2021bayesian}. We have the following empiricism on how to provide  safe initializations for different types of problems, as adopted in our experiments in Section \ref{section.applications}. In safe policy optimization, one could first use supervised learning to learn a safe policy from some safe trajectories/demonstrations (not necessarily be optimal) and then use the learned safe policy to initialize  Safe PDP.
In safe motion planning, one could arbitrarily provide a safe trajectory (not necessarily optimal) to initialize  Safe PDP.
In learning MPCs, the goal includes learning of  constraint itself, and there is no such requirement.

\textbf{Strategies to accelerate forward pass of Safe PDP.} \,\, There are many strategies to accelerate a long-horizon  trajectory optimization (optimal control) in  the forward pass of Safe PDP. (I) One effective way is to scale the (continuous)
long-horizon problem into a smaller one (e.g., a unit) by applying a  time-warping function  to the 
dynamics and cost function \cite{jin2020sparse}. After  solving the scaled short-horizon problem,
 re-scale the trajectory back.
(II) There are also `warm-up’ tricks, e.g., one
can initialize the trajectory at the next iteration using the result of the previous iteration.
(III) One  can also use a hierarchical strategy to solve trajectory optimization from coarse to fine resolutions. We have tested and provided the comparison for the above three acceleration strategies in Appendix \ref{appendix.discussion.acceleration}.

Please refer to Appendix \ref{appendix.discussion} for more discussion, which includes \ref{appendix.discussion.pdpcompare}:  comparison between Safe-PDP and non-safe PDP;  \ref{appendix.discussion.acceleration}:  comparison of different  strategies for accelerating long-horizon trajectory optimization; \ref{appendix.discussion.tradeoff}:   trade-offs between accuracy and computational efficiency using barrier penalties; \ref{appendix.discussion.nonoptimal}: learning MPCs from non-optimal data; and \ref{appendix.discussion.limitation}:  detailed discussion on  limitation of Safe PDP.

\vspace{-2pt}
\section{Conclusions}
\vspace{-5pt}
This paper proposes a Safe Pontryagin Differentiable Programming methodology, which establishes a provable and systematic safe differentiable framework to solve a broad class of safety-critical control and learning tasks with different types of safety constraints. For a constrained system and task, Safe PDP  approximates both the solution and its gradient in  backward pass by solving their more efficient unconstrained counterparts. Safe PDP has established two results: one is the  \emph{controlled accuracy guarantee} for  approximations of  the solution and its gradient, and the other is the  \emph{safety guarantee} for constraint satisfaction throughout the control and learning process. We envision the potential of Safe PDP  for addressing various safety-critical problems in  machine learning, control, and robotics fields.

\newpage

\begin{ack}
This work is supported by the NASA University Leadership Initiative (ULI) under grant number 80NSSC20M0161. The research of Prof. George J. Pappas is supported by the AFOSR Assured Autonomy in Congested Environments under grant number FA9550-19-1-0265. This work has been done primarily in the last semester of Wanxin Jin's Ph.D. study at Purdue University. Wanxin Jin  thanks Prof. Zhaoran Wang for some discussion about this work.
\end{ack}

\bibliographystyle{unsrt}
\bibliography{diffcontrol}

\section*{Checklist}

\begin{enumerate}
	
	\item For all authors...
	\begin{enumerate}
		\item Do the main claims made in the abstract and introduction accurately reflect the paper's contributions and scope?
		\answerYes{}
		\item Did you describe the limitations of your work?
		\answerYes{Please see Sections \ref{discussion.section} and  Appendix \ref{appendix.discussion.limitation} in the paper.}
		\item Did you discuss any potential negative societal impacts of your work?
		\answerNA{}
		\item Have you read the ethics review guidelines and ensured that your paper conforms to them?
		\answerYes{}
	\end{enumerate}
	
	\item If you are including theoretical results...
	\begin{enumerate}
		\item Did you state the full set of assumptions of all theoretical results?
		\answerYes{Pleae find them in all theorems in the paper.}
		\item Did you include complete proofs of all theoretical results?
		\answerYes{Please find  complete proofs for all theoretical results in the Appendix in the supplementary file.}
	\end{enumerate}
	
	\item If you ran experiments...
	\begin{enumerate}
		\item Did you include the code, data, and instructions needed to reproduce the main experimental results (either in the supplemental material or as a URL)?
		\answerYes{Please refer to the code at \url{https://github.com/wanxinjin/Safe-PDP}.}
		\item Did you specify all the training details (e.g., data splits, hyperparameters, how they were chosen)?
		\answerYes{Please see the Appendix \ref{appendix.experimentdetails} in the supplementary file.}
		\item Did you report error bars (e.g., with respect to the random seed after running experiments multiple times)?
		\answerNA{}
		\item Did you include the total amount of compute and the type of resources used (e.g., type of GPUs, internal cluster, or cloud provider)?
		\answerYes{Please see the Appendix  \ref{appendix.experimentdetails} in the supplementary file.}
	\end{enumerate}
	
	\item If you are using existing assets (e.g., code, data, models) or curating/releasing new assets...
	\begin{enumerate}
		\item If your work uses existing assets, did you cite the creators?
		\answerYes{Please see  the citation of the experimental environments  in Table \ref{experimenttable} in the paper.}
		\item Did you mention the license of the assets?
		\answerNA{}
		\item Did you include any new assets either in the supplemental material or as a URL?
		\answerYes{Some video demo links are included in  Appendix \ref{appendix.experimentdetails} in the supplementary file.}
		\item Did you discuss whether and how consent was obtained from people whose data you're using/curating?
		\answerNA{}
		\item Did you discuss whether the data you are using/curating contains personally identifiable information or offensive content?
		\answerNA{}
	\end{enumerate}
	
	\item If you used crowdsourcing or conducted research with human subjects...
	\begin{enumerate}
		\item Did you include the full text of instructions given to participants and screenshots, if applicable?
		\answerNA{}
		\item Did you describe any potential participant risks, with links to Institutional Review Board (IRB) approvals, if applicable?
		\answerNA{}
		\item Did you include the estimated hourly wage paid to participants and the total amount spent on participant compensation?
		\answerNA{}
	\end{enumerate}
	
\end{enumerate}


\newpage
\appendix
Appendix to the Safe Pontryagin Differentiable Programming paper

\setcounter{equation}{0}
\setcounter{figure}{0}   
\setcounter{table}{0}
\renewcommand\theequation{S.\arabic{equation}}
\renewcommand{\thefigure}{S\arabic{figure}}
\renewcommand{\thetable}{S\arabic{table}}

\section{Second-order Sufficient Condition} \label{appendix.secondordercondition}

Before presenting the second-order condition for the optimal control  Problem \ref{equ_traj}, we present the second-order condition for a general constrained nonlinear programming. The interested reader can find the details in Theorem 4 in \cite{fiacco1990nonlinear}.
\begin{lemma}[Second-order sufficient condition  \cite{fiacco1990nonlinear}] \label{lemma_generalnlp}
	If all functions defining a constrained optimization 
	\begin{equation}\label{pf_generalnlp}
	\begin{aligned}
	\min_{\boldsymbol{x}} 
	\quad
	&f(\boldsymbol{x})\\
	\text{subject to} \quad & {g}_i(\boldsymbol{x})\leq 0 \quad i=1,2,\cdots,m,\\
	& h_j(\boldsymbol{x})=0 \quad j=1,2,\cdots, p,
	\end{aligned}
	\end{equation}
	are twice-continuous differentiable, the second-order sufficient condition for $\boldsymbol{x}^*$ to be a local isolated minimizing solution to (\ref{pf_generalnlp}) is that there exist vectors $\boldsymbol{v}^*$ and $\boldsymbol{w}^*$ such that $(\boldsymbol{x}^*, \boldsymbol{v}^*, \boldsymbol{w}^*)$  satisfies
	\begin{equation}
	\begin{aligned}
	g_i(\boldsymbol{x}^*)\leq 0, &\quad i=1,2,\cdots, m,\\
	h_i(\boldsymbol{x}^*)=0, & \quad  j=1,2,\cdots, p,\\
	v_ig_i(\boldsymbol{x}^*)=0, & \quad i=1,2,\cdots, m,\\
	u_i\geq 0, &\quad i=1,2,\cdots, m,\\
	\nabla L(\boldsymbol{x}^*, \boldsymbol{v}^*, \boldsymbol{w}^*)=0, &
	\end{aligned}
	\end{equation}
	with
	\begin{equation}
	L(\boldsymbol{x},\boldsymbol{v},\boldsymbol{w})=f(\boldsymbol{x})+\sum_{i=1}^{m}v_i g_i(\boldsymbol{x})+\sum_{i=1}^{p}w_i h_i(\boldsymbol{x}),
	\end{equation} and 
	$\nabla L$ being the derivative of $L$ with respect to $\boldsymbol{x}$; and  
	further  for any nonzero $\boldsymbol{y}\neq\boldsymbol{0}$ satisfying $\boldsymbol{y}^\prime \nabla g_i({\boldsymbol{x}^*})=0$ for all $i$ with ${v}_i^*>0$,  $\boldsymbol{y}^\prime \nabla g_i({\boldsymbol{x}^*})\leq 0$ for all $i$ with ${v}_i^*\geq 0$, and $\boldsymbol{y}^\prime \nabla h_j({\boldsymbol{x}^*})=0$ for all $j=1,2,\cdots, p$, it follows that 
	\begin{equation}\label{second-order-lemma_generalnlp}
	\boldsymbol{y}^\prime \nabla^2L(\boldsymbol{x}^*, \boldsymbol{v}^*, \boldsymbol{w}^*)\boldsymbol{y}>0.
	\end{equation}
\end{lemma}

The above second-order sufficient condition for nonlinear programming is well-known. The proof for  Lemma \ref{lemma_generalnlp} can be found  in Theorem 4 in \cite{fiacco1990nonlinear}. Similarly, we can establish the second-order sufficient condition for a general constrained optimal control system $\boldsymbol{\Sigma}(\boldsymbol{\theta})$ in (\ref{equ_oc}), as below.

\bigskip

\begin{lemma}[Second-order sufficient condition for  $\boldsymbol{\Sigma}(\boldsymbol{\theta})$ to have a local isolated minimizing trajectory $\boldsymbol{\xi}_{\boldsymbol{{\theta}}}$ \cite{pearson1966discrete}]\label{Lemma1}
	Given  $\boldsymbol{\theta}$, if all functions  defining the constrained optimal control system $\boldsymbol{\Sigma}(\boldsymbol{\theta})$ are twice continuously differentiable in a neighborhood (tube) of   $\boldsymbol{\xi}_{\boldsymbol{{\theta}}}=\{\boldsymbol{x}_{0:T}^{\boldsymbol{\theta}},\boldsymbol{u}_{0:T-1}^{\boldsymbol{\theta}}\}$,   $\boldsymbol{\xi}_{\boldsymbol{{\theta}}}$ is a local isolated minimizing trajectory to   Problem \ref{equ_traj} if there exist  sequences   $\boldsymbol{\lambda}_{1:T}^{\boldsymbol{\theta}}$, $\boldsymbol{v}_{0:T}^{\boldsymbol{\theta}}$, and $\boldsymbol{w}_{0:T}^{\boldsymbol{\theta}}$ such that the following  Constrained Pontryagin Maximum/Minimum Principle (C-PMP) conditions hold,
	\begin{equation}\label{CPMP}
	\begin{aligned}
	&\boldsymbol{x}_{t+1}^{\boldsymbol{\theta}}=\boldsymbol{f}(\boldsymbol{x}_{t}^{\boldsymbol{\theta}},\boldsymbol{u}_{t}^{\boldsymbol{\theta}},\boldsymbol{\theta})
	\quad\text{and}\quad
	\boldsymbol{x}_{0}^{\boldsymbol{\theta}}=\boldsymbol{x}_{0}({\boldsymbol{\theta}}),
	\\
	&\boldsymbol{\lambda}_{t}^{\boldsymbol{\theta}}=L_{t}^x(\boldsymbol{x}_{t}^{\boldsymbol{\theta}},\boldsymbol{u}_{t}^{\boldsymbol{\theta}},\boldsymbol{\lambda}_{t+1}^{\boldsymbol{\theta}},\boldsymbol{v}_{t}^{\boldsymbol{\theta}},\boldsymbol{w}_{t}^{\boldsymbol{\theta}},\boldsymbol{\theta})
	\quad\text{and}\quad
	\boldsymbol{\lambda}_{T}^{\boldsymbol{\theta}}=L_{T}^x(\boldsymbol{x}_{t}^{\boldsymbol{\theta}},\boldsymbol{v}_{t}^{\boldsymbol{\theta}},\boldsymbol{w}_{t}^{\boldsymbol{\theta}},\boldsymbol{\theta}),
	\\
	&\boldsymbol{0}=
	L_{t}^u(\boldsymbol{x}_{t}^{\boldsymbol{\theta}},\boldsymbol{u}_{t}^{\boldsymbol{\theta}},\boldsymbol{\lambda}_{t+1}^{\boldsymbol{\theta}},\boldsymbol{v}_{t}^{\boldsymbol{\theta}},\boldsymbol{w}_{t}^{\boldsymbol{\theta}},\boldsymbol{\theta}),
	\\
	&{h}_{t,j}(\boldsymbol{x}_{t}^{\boldsymbol{\theta}},\boldsymbol{u}_{t}^{\boldsymbol{\theta}},\boldsymbol{\theta})={0},\quad j=1,2,\cdots, s_t, \\
	&
	{h}_{T,j}(\boldsymbol{x}_{T}^{\boldsymbol{\theta}},\boldsymbol{\theta})={0}, \quad j=1,2,\cdots, s_T,
	\\
	&{g}_{t,i}(\boldsymbol{x}_{t}^{\boldsymbol{\theta}},\boldsymbol{u}_{t}^{\boldsymbol{\theta}},\boldsymbol{\theta})\leq{0},\quad {v}_{t,i}{g}_{t,i}(\boldsymbol{x}_{t}^{\boldsymbol{\theta}},\boldsymbol{u}_{t}^{\boldsymbol{\theta}},\boldsymbol{\theta})={0},\quad {v}_{t,i}\geq 0,
	\quad
	i=1,2,\cdots, q_t,\\
	&
	{g}_{T,i}(\boldsymbol{x}_{T}^{\boldsymbol{\theta}},\boldsymbol{\theta})\leq{0},\quad {v}_{T,i}{g}_{T,i}(\boldsymbol{x}_{T}^{\boldsymbol{\theta}},\boldsymbol{\theta})={0},\quad {v}_{T,i}\geq 0, \quad i=1,2,\cdots, q_T,
	\end{aligned}
	\end{equation}
	and further if 
	\begin{equation}\label{secordercon.1}
	\sum_{t=0}^{T-1}
	\begin{bmatrix}
	\boldsymbol{x}_t\\
	\boldsymbol{u}_t
	\end{bmatrix}^\prime
	\begin{bmatrix}
	L_t^{xx}& L_t^{xu}\\
	L_t^{ux}& L_t^{uu}
	\end{bmatrix}
	\begin{bmatrix}
	\boldsymbol{x}_t\\
	\boldsymbol{u}_t
	\end{bmatrix}+
	\boldsymbol{x}_T^\prime L_T^{xx}  \boldsymbol{x}_T \,>0
	\end{equation}
	for any non-zero trajectory $\{\boldsymbol{x}_{0:T}, \boldsymbol{u}_{0:T-1}\}\neq\boldsymbol{0}$ satisfying
	\begin{equation}\label{bindingequation}
	\begin{aligned}
	&\boldsymbol{x}_{t+1}=F_t^x\boldsymbol{x}_t+F_t^u\boldsymbol{u}_t \qquad \boldsymbol{x}_{0}=\boldsymbol{0},
	\\
	&
	H_t^x\boldsymbol{x}_t+H_t^u\boldsymbol{u}_t=\boldsymbol{0} \quad \text{and} \quad {H}_T^x\boldsymbol{x}_T=\boldsymbol{0},
	\\
	&\check{G}_t^x\boldsymbol{x}_t+\check{G}_t^u\boldsymbol{u}_t=\boldsymbol{0} 
	\quad \text{and} \quad \check{G}_T^x\boldsymbol{x}_T=\boldsymbol{0}, \\
	&
	\bar{G}_t^x\boldsymbol{x}_t+\bar{G}_t^u\boldsymbol{u}_t\leq\boldsymbol{0} 
	\quad \text{and} \quad \bar{G}_T^x\boldsymbol{x}_T\leq\boldsymbol{0}.
	\end{aligned}
	\end{equation}
	Here, $t=0,1,...,T-1$;  $L_t^x$ is the first-order derivative   of the Hamiltonian $L_t$  in (\ref{Hamiltonian}) with respect to $\boldsymbol{x}$, and  $L_t^{xx}$ is the second-derivative of $L_t$ with respect to $\boldsymbol{x}$, and similar notation convention applies to $L_T^x$, $L_t^u$, $L_t^{xu}=(L_t^{ux})^\prime$, and $L_t^{uu}$; $H_t^x$ is the first-order derivative  of  $\boldsymbol{h}_t$ with respect to $\boldsymbol{x}$ and the  similar  convention applies to $H_T^x$,  $H_t^u$, $F_t^x$ and $F_t^u$ for  $\boldsymbol{f}$, $\check{G}_t^x$ and $\check{G}_t^u$  for ${\boldsymbol{\check{g}}}_t$,  $\check{G}_T^x$ for ${\boldsymbol{\check{g}}}_T$,	 
	$\bar{G}_t^x$ and $\bar{G}_t^u$  for ${\boldsymbol{\bar{g}}}_t$, $\bar{G}_T^x$ for ${\boldsymbol{\bar{g}}}_T$, where  
	\begin{equation}\label{activeinequ}
	\begin{aligned}
	{\boldsymbol{\check{g}}}_t(\boldsymbol{x}_t,\boldsymbol{u}_t,\boldsymbol{\theta})&=\col\{g_{t,i}(\boldsymbol{x}_t,\boldsymbol{u}_t,\boldsymbol{\theta})\,| \,v_{t,i}^{\boldsymbol{\theta}}>0, i=1,...,q_t\}, 
	\\
	{\boldsymbol{\check{g}}}_T(\boldsymbol{x}_T,\boldsymbol{\theta})&=\col\{g_{T,i}(\boldsymbol{x}_T,\boldsymbol{\theta})\,| \,v_{t,i}^{\boldsymbol{\theta}}>0, i=1,...,q_T\},
	\quad\\
	{\boldsymbol{\bar{g}}}_t(\boldsymbol{x}_t,\boldsymbol{u}_t,\boldsymbol{\theta})&=\col\{g_{t,i}(\boldsymbol{x}_t,\boldsymbol{u}_t,\boldsymbol{\theta})\,|\,g_{t,i}(\boldsymbol{x}_t^{\boldsymbol{\theta}},\boldsymbol{u}_t^{\boldsymbol{\theta}},\boldsymbol{\theta})=0, i=1,...,q_t\} \in\mathbb{R}^{\bar{q}_t},
	\\
	{\boldsymbol{\bar{g}}}_T(\boldsymbol{x}_T,\boldsymbol{\theta})&=\col\{g_{T,i}(\boldsymbol{x}_T,\boldsymbol{\theta})\,|\,g_{T,i}(\boldsymbol{x}_T^{\boldsymbol{\theta}},\boldsymbol{\theta})=0, i=1,...,q_T\}\in\mathbb{R}^{\bar{q}_T},
	\end{aligned}
	\end{equation}
	i.e.,   $\bar{\boldsymbol{{g}}}_t$ and $\bar{\boldsymbol{{g}}}_T$ are the vector functions formed by stacking all active inequality constraints at $\boldsymbol{\xi}_{\boldsymbol{\theta}}$. All the above first- and second-order derivatives are evaluated at $(\boldsymbol{\xi}_{\boldsymbol{\theta}}, \boldsymbol{\lambda}_{1:T}^{\boldsymbol{\theta}}, \boldsymbol{v}_{0:T}^{\boldsymbol{\theta}},  \boldsymbol{w}_{0:T}^{\boldsymbol{\theta}} )$.
\end{lemma}

The above  second-order sufficient condition for the constrained optimal control system $\boldsymbol{\Sigma}(\boldsymbol{\theta})$  is well-known and has been well-established since \cite{pearson1966discrete}. The conditions in (\ref{CPMP}) is referred to as discrete-time Constrained Pontryagin Maximum/Minimum Principle (C-PMP) \cite{pearson1966discrete}. Note that in the case of strict complementarity, one has ${\boldsymbol{\check{g}}}_t(\boldsymbol{x}_t,\boldsymbol{u}_t,\boldsymbol{\theta})={\boldsymbol{\bar{g}}}_t(\boldsymbol{x}_t,\boldsymbol{u}_t,\boldsymbol{\theta})$ and ${\boldsymbol{\check{g}}}_T(\boldsymbol{x}_T,\boldsymbol{\theta})={\boldsymbol{\bar{g}}}_T(\boldsymbol{x}_T,\boldsymbol{\theta})$ in (\ref{activeinequ}).

\bigskip

\section{Proof of Theorem \ref{theorem1}}\label{pf_theorem1}

To prove Theorem \ref{theorem1}, in the first part, we need to  derive the Differential Constrained Pontryagin Maximum/Minimum Principle (Differential C-PMP),  which $	\frac{\partial\boldsymbol{\xi}_{{\boldsymbol{{\theta}}}} }{\partial \boldsymbol{\theta}}=\left\{
X^{{\boldsymbol{{\theta}}}}_{0:T}, U^{{\boldsymbol{{\theta}}}}_{0:T}
\right\}$ must satisfy. Then, in the second part, we formally present the proof for Theorem \ref{theorem1}.

\subsection{Differential Constrained Pontryagin Maximum/Minimum Principle}\label{pf_theorem1_diffcpmp}

From Lemma \ref{result1}, for the constrained optimal control system  $\boldsymbol{\Sigma}(\boldsymbol{\theta})$  with any $\boldsymbol{\theta}$ in a neighborhood of ${\boldsymbol{\bar\theta}}$, ($\boldsymbol{\xi}_{\boldsymbol{{\theta}}}$, $ \boldsymbol{\lambda}_{1:T}^{{\boldsymbol{{\theta}}}}$, $\boldsymbol{v}_{0:T}^{{\boldsymbol{{\theta}}}}$, $\boldsymbol{w}_{0:T}^{{\boldsymbol{{\theta}}}}$) satisfies the C-PMP conditions in (\ref{CPMP}). Since ($\boldsymbol{\xi}_{\boldsymbol{{\theta}}}$, $ \boldsymbol{\lambda}_{1:T}^{{\boldsymbol{{\theta}}}}$, $\boldsymbol{v}_{0:T}^{{\boldsymbol{{\theta}}}}$, $\boldsymbol{w}_{0:T}^{{\boldsymbol{{\theta}}}}$) is also  once-continuously differentiable with respect to $\boldsymbol{\theta}$ from Lemma \ref{result1}, one can differentiate the C-PMP conditions in (\ref{CPMP}) on both sides with respect to $\boldsymbol{\theta}$, as below.

Differentiating the first five lines in (\ref{CPMP}) is straightforward, yielding
\begin{equation}\label{diff-CPMP1}
\begin{aligned}
&\frac{\partial\boldsymbol{x}^{\boldsymbol{\theta}}_{t+1}}{\partial\boldsymbol{\theta}}=F_t^{x}\frac{\partial\boldsymbol{x}^{\boldsymbol{\theta}}_{t}}{\partial\boldsymbol{\theta}}+F_t^{u}\frac{\partial\boldsymbol{u}^{\boldsymbol{\theta}}_{t}}{\partial\boldsymbol{\theta}}+F_t^\theta \quad \text{and}\quad X_0^{\boldsymbol{\theta}}=\frac{\partial \boldsymbol{x}_0^{\boldsymbol{\theta}}}{\partial \boldsymbol{\theta}}=\frac{\partial \boldsymbol{x}_0(\boldsymbol{\theta})}{\partial \boldsymbol{\theta}},\\
&\frac{\partial \boldsymbol{\lambda}_t^{\boldsymbol{\theta}}}{\partial\boldsymbol{\theta}}=L_t^{xx}\frac{\partial\boldsymbol{x}^{\boldsymbol{\theta}}_{t}}{\partial\boldsymbol{\theta}}+L_t^{xu}\frac{\partial\boldsymbol{u}^{\boldsymbol{\theta}}_{t}}{\partial\boldsymbol{\theta}}+(F_{t}^{x})^\prime\frac{\partial \boldsymbol{\lambda}_{t+1}^{\boldsymbol{\theta}}}{\partial\boldsymbol{\theta}}+( G_{t}^{x})^\prime\frac{\partial\boldsymbol{v}^{\boldsymbol{\theta}}_{t}}{\partial\boldsymbol{\theta}}+(H_{t}^{x})^\prime \frac{\partial\boldsymbol{w}^{\boldsymbol{\theta}}_{t}}{\partial\boldsymbol{\theta}}+L_t^{x\theta}\quad \text{and}\quad
\\
&\frac{\partial \boldsymbol{\lambda}_T^{\boldsymbol{\theta}}}{\partial\boldsymbol{\theta}}=L_T^{xx}\frac{\partial\boldsymbol{x}^{\boldsymbol{\theta}}_{T}}{\partial\boldsymbol{\theta}}+({G}_{T}^{x})^\prime\frac{\partial\boldsymbol{v}^{\boldsymbol{\theta}}_{T}}{\partial\boldsymbol{\theta}}+(H_{T}^{x})^\prime \frac{\partial\boldsymbol{w}^{\boldsymbol{\theta}}_{T}}{\partial\boldsymbol{\theta}}+L_T^{x\theta},
\\
&\boldsymbol{0}=L_t^{ux}\frac{\partial\boldsymbol{x}^{\boldsymbol{\theta}}_{t}}{\partial\boldsymbol{\theta}}+L_t^{uu}\frac{\partial\boldsymbol{u}^{\boldsymbol{\theta}}_{t}}{\partial\boldsymbol{\theta}}+(F_{t}^{u})^\prime\frac{\partial \boldsymbol{\lambda}_{t+1}^{\boldsymbol{\theta}}}{\partial\boldsymbol{\theta}}+( G_{t}^{u})^\prime\frac{\partial\boldsymbol{v}^{\boldsymbol{\theta}}_{t}}{\partial\boldsymbol{\theta}}+(H_{t}^{u})^\prime \frac{\partial\boldsymbol{w}^{\boldsymbol{\theta}}_{t}}{\partial\boldsymbol{\theta}}+L_t^{u\theta},
\\
&H_t^{x}\frac{\partial\boldsymbol{x}^{\boldsymbol{\theta}}_{t}}{\partial\boldsymbol{\theta}}+H_t^{u}\frac{\partial\boldsymbol{u}^{\boldsymbol{\theta}}_{t}}{\partial\boldsymbol{\theta}}+H_t^{\theta}=\boldsymbol{0}
\quad \text{and}\quad
H_T^{x}\frac{\partial\boldsymbol{x}^{\boldsymbol{\theta}}_{t}}{\partial\boldsymbol{\theta}}+H_T^{\theta}=\boldsymbol{0}.
\end{aligned}
\end{equation}

We now consider to differentiate the two last equations (i.e., complementarity conditions) in the last two lines in (\ref{CPMP}).
We start with
\begin{equation}\label{appexeq.pathineq.pointwise}
v_{t,i}^{\boldsymbol{\theta}}\,\,g_{t,i}(\boldsymbol{x}_t^{\boldsymbol{\theta}},\boldsymbol{u}_t^{\boldsymbol{\theta}},\boldsymbol{\theta})=0, \quad\quad i=1,2,\cdots,q_t.
\end{equation}

Differentiating the above (\ref{appexeq.pathineq.pointwise}) on both sides with respect to $\boldsymbol{\theta}$ yields
\begin{equation}\label{sequ_pathinequlity_diff}
\frac{\partial v_{t,i}^{\boldsymbol{\theta}}}{\partial \boldsymbol{\theta}}
\,\,g_{t,i}(\boldsymbol{x}_t^{\boldsymbol{\theta}},\boldsymbol{u}_t^{\boldsymbol{\theta}},\boldsymbol{\theta})
+\mu_{t,i}^{\boldsymbol{\theta}}
\,\,\frac{\partial g_{t,i}(\boldsymbol{x}_t^{\boldsymbol{\theta}},\boldsymbol{u}_t^{\boldsymbol{\theta}},\boldsymbol{\theta})}{\partial \boldsymbol{\theta}}=\boldsymbol{0}, \quad  i=1,2,\cdots,q_t.
\end{equation}
In the above, we consider two following cases.  If $g_{t,i}(\boldsymbol{x}_t^{\boldsymbol{\theta}},\boldsymbol{u}_t^{\boldsymbol{\theta}},\boldsymbol{\theta})=0$, i.e., $g_{t,i}$ is an active inequality constraint,  then, $\mu_{t,i}^{\boldsymbol{\theta}}>0$ according to strict complementarity (condition (iii) in Lemma \ref{result1}). From (\ref{sequ_pathinequlity_diff}), one thus has
\begin{equation}\label{sequ_pathinequlity_diff.case1}
\frac{\partial g_{t,i}(\boldsymbol{x}_t^{\boldsymbol{\theta}},\boldsymbol{u}_t^{\boldsymbol{\theta}},\boldsymbol{\theta})}{\partial \boldsymbol{\theta}}=\boldsymbol{0}.
\end{equation}
If $g_{t,i}(\boldsymbol{x}_t^{\boldsymbol{\theta}},\boldsymbol{u}_t^{\boldsymbol{\theta}},\boldsymbol{\theta})<0$, i.e., $g_{t,i}$ is an inactive constraint,  then $v_{t,i}^{\boldsymbol{\theta}}=0$ and one has
\begin{equation}\label{sequ_pathinequlity_diff.case2}
\frac{\partial v_{t,i}^{\boldsymbol{\theta}}}{\partial \boldsymbol{\theta}}=\boldsymbol{0} \quad 
\text{for}
\quad
v_{t,i}^{\boldsymbol{\theta}}=0.
\end{equation}
Stacking (\ref{sequ_pathinequlity_diff.case1}) for all  active inequality constraints defined in (\ref{activeinequ}) will lead to 
\begin{equation}\label{diff-CPMP2}
\boldsymbol{0}=
\bar{G}_t^x\frac{\partial\boldsymbol{x}^{\boldsymbol{\theta}}_{t}}{\partial\boldsymbol{\theta}}+\bar{G}_t^u\frac{\partial\boldsymbol{u}^{\boldsymbol{\theta}}_{t}}{\partial\boldsymbol{\theta}}+\bar{G}_t^\theta.
\end{equation}
Similarly, we can show that differentiating ${v}_{T,i} {g}_{T,i}(\boldsymbol{x}_{T}^{\boldsymbol{\theta}},\boldsymbol{\theta})={0}, i=1,2,\cdots, q_T,$ will lead to 
\begin{equation}\label{diff-CPMP3}
\bar{G}_T^{x}\frac{\partial\boldsymbol{x}^{\boldsymbol{\theta}}_{T}}{\partial\boldsymbol{\theta}}+\bar{G}_T^{\theta}=\boldsymbol{0}.
\end{equation}

If we further define 
\begin{gather}\label{def_vbar}
\bar{\boldsymbol{v}}_{t}^{\boldsymbol{\theta}}=\col\{v_{t,i}^{\boldsymbol{\theta}}\,|\,v_{t,i}^{\boldsymbol{\theta}}>0, i=1,...,q_t \}\in\mathbb{R}^{\bar{q}_t},
\end{gather}
then, due to  (\ref{sequ_pathinequlity_diff.case2}),  the following terms in the second, third, and fourth lines in (\ref{diff-CPMP1}) can be  written in an equivalent way:
\begin{equation}\label{diff-CPMP4}
\left(
G_t^x
\right)^\prime \frac{\partial \boldsymbol{v}_{t}^{\boldsymbol{\theta}}}{\partial \boldsymbol{\theta}} =\left(\bar{G}_t^x
\right)^\prime\frac{\partial \boldsymbol{\bar v}_{t}^{\boldsymbol{\theta}}}{\partial \boldsymbol{\theta}}\quad\text{and}\quad\left(
G_t^u
\right)^\prime \frac{\partial \boldsymbol{v}_{t}^{\boldsymbol{\theta}}}{\partial \boldsymbol{\theta}}=\left(\bar{G}_t^u
\right)^\prime\frac{\partial \boldsymbol{\bar v}_{t}^{\boldsymbol{\theta}}}{\partial \boldsymbol{\theta}}.
\end{equation}

In sum, combining (\ref{diff-CPMP1}), (\ref{diff-CPMP2}), (\ref{diff-CPMP3}), and (\ref{diff-CPMP4}), one can finally write the Differential C-PMP:
\begin{equation}\label{diff-CPMP}
\begin{aligned}
&\frac{\partial\boldsymbol{x}^{\boldsymbol{\theta}}_{t+1}}{\partial\boldsymbol{\theta}}=F_t^{x}\frac{\partial\boldsymbol{x}^{\boldsymbol{\theta}}_{t}}{\partial\boldsymbol{\theta}}+F_t^{u}\frac{\partial\boldsymbol{u}^{\boldsymbol{\theta}}_{t}}{\partial\boldsymbol{\theta}}+F_t^\theta \quad \text{and}\quad X_0^{\boldsymbol{\theta}}=\frac{\partial \boldsymbol{x}_0^{\boldsymbol{\theta}}}{\partial \boldsymbol{\theta}},\\
&\frac{\partial \boldsymbol{\lambda}_t^{\boldsymbol{\theta}}}{\partial\boldsymbol{\theta}}=L_t^{xx}\frac{\partial\boldsymbol{x}^{\boldsymbol{\theta}}_{t}}{\partial\boldsymbol{\theta}}+L_t^{xu}\frac{\partial\boldsymbol{u}^{\boldsymbol{\theta}}_{t}}{\partial\boldsymbol{\theta}}+(F_{t}^{x})^\prime\frac{\partial \boldsymbol{\lambda}_{t+1}^{\boldsymbol{\theta}}}{\partial\boldsymbol{\theta}}+(\bar G_{t}^{x})^\prime\frac{\partial\boldsymbol{\bar v}^{\boldsymbol{\theta}}_{t}}{\partial\boldsymbol{\theta}}+(H_{t}^{x})^\prime \frac{\partial\boldsymbol{w}^{\boldsymbol{\theta}}_{t}}{\partial\boldsymbol{\theta}}+L_t^{x\theta}\quad \text{and}\quad
\\
&\frac{\partial \boldsymbol{\lambda}_T^{\boldsymbol{\theta}}}{\partial\boldsymbol{\theta}}=L_T^{xx}\frac{\partial\boldsymbol{x}^{\boldsymbol{\theta}}_{T}}{\partial\boldsymbol{\theta}}+(\bar{G}_{T}^{x})^\prime\frac{\partial\boldsymbol{\bar v}^{\boldsymbol{\theta}}_{T}}{\partial\boldsymbol{\theta}}+(H_{T}^{x})^\prime \frac{\partial\boldsymbol{w}^{\boldsymbol{\theta}}_{T}}{\partial\boldsymbol{\theta}}+L_T^{x\theta},
\\
&\boldsymbol{0}=L_t^{ux}\frac{\partial\boldsymbol{x}^{\boldsymbol{\theta}}_{t}}{\partial\boldsymbol{\theta}}+L_t^{uu}\frac{\partial\boldsymbol{u}^{\boldsymbol{\theta}}_{t}}{\partial\boldsymbol{\theta}}+(F_{t}^{u})^\prime\frac{\partial \boldsymbol{\lambda}_{t+1}^{\boldsymbol{\theta}}}{\partial\boldsymbol{\theta}}+( \bar G_{t}^{u})^\prime\frac{\partial\boldsymbol{\bar v}^{\boldsymbol{\theta}}_{t}}{\partial\boldsymbol{\theta}}+(H_{t}^{u})^\prime \frac{\partial\boldsymbol{w}^{\boldsymbol{\theta}}_{t}}{\partial\boldsymbol{\theta}}+L_t^{u\theta},
\\
&H_t^{x}\frac{\partial\boldsymbol{x}^{\boldsymbol{\theta}}_{t}}{\partial\boldsymbol{\theta}}+H_t^{u}\frac{\partial\boldsymbol{u}^{\boldsymbol{\theta}}_{t}}{\partial\boldsymbol{\theta}}+H_t^{\theta}=\boldsymbol{0}
\quad \text{and}\quad
H_T^{x}\frac{\partial\boldsymbol{x}^{\boldsymbol{\theta}}_{t}}{\partial\boldsymbol{\theta}}+H_T^{\theta}=\boldsymbol{0},\\
& 
\bar{G}_t^x\frac{\partial\boldsymbol{x}^{\boldsymbol{\theta}}_{t}}{\partial\boldsymbol{\theta}}+\bar{G}_t^u\frac{\partial\boldsymbol{u}^{\boldsymbol{\theta}}_{t}}{\partial\boldsymbol{\theta}}+\bar{G}_t^\theta=\boldsymbol{0}
\quad \text{and}\quad
\bar{G}_T^{x}\frac{\partial\boldsymbol{x}^{\boldsymbol{\theta}}_{T}}{\partial\boldsymbol{\theta}}+\bar{G}_T^{\theta}=\boldsymbol{0}.
\end{aligned}
\end{equation}

With the above Differential C-PMP, we next prove the claims in Theorem \ref{theorem1}.

\subsection{Proof of Theorem \ref{theorem1}}

We prove Theorem \ref{theorem1} by two steps.  We first  prove that the trajectory in (\ref{difftraj}), rewritten below,
\begin{equation*}\label{aux_traj_pf}
\left\{
X^{{\boldsymbol{{\theta}}}}_{0:T}, U^{{\boldsymbol{{\theta}}}}_{0:T-1}
\right\} \quad \text{with} \quad X^{{\boldsymbol{{\theta}}}}_{t}=\frac{\partial\boldsymbol{x}_t^{{\boldsymbol{{\theta}}}} }{\partial \boldsymbol{\theta}} \quad \text{and}  \quad U^{{\boldsymbol{{\theta}}}}_{t}= \frac{\partial\boldsymbol{u}_t^{{\boldsymbol{{\theta}}}} }{\partial \boldsymbol{\theta}},
\end{equation*}
is the \emph{local isolated}  minimizing trajectory to the auxiliary control system $\boldsymbol{\overline\Sigma}(\boldsymbol{\xi}_{\boldsymbol{\theta}})$ in (\ref{equ_aux}); and second, we prove that such a local minimizing trajectory is also a \emph{global} minimizing trajectory.

First, we prove that $\left\{
X^{{\boldsymbol{{\theta}}}}_{0:T}, U^{{\boldsymbol{{\theta}}}}_{0:T-1}
\right\}$ is a  \emph{local isolated} minimizing trajectory to $\boldsymbol{\overline\Sigma}(\boldsymbol{\xi}_{\boldsymbol{\theta}})$.

To show that $\left\{
X^{{\boldsymbol{{\theta}}}}_{0:T}, U^{{\boldsymbol{{\theta}}}}_{0:T-1}
\right\}$ is  a {local isolated}  minimizing trajectory to $\boldsymbol{\overline\Sigma}(\boldsymbol{\xi}_{\boldsymbol{\theta}})$, we only need to check whether it satisfies the second-order sufficient condition for the constrained optimal control system $\boldsymbol{\overline\Sigma}(\boldsymbol{\xi}_{\boldsymbol{\theta}})$, as stated in Lemma \ref{Lemma1}. To that end,  we define the following Hamiltonian for   $\boldsymbol{\overline\Sigma}(\boldsymbol{{\xi}}_{\boldsymbol{\theta}})$:
\begin{equation}
\begin{aligned}
\bar{L}_t=&\Tr\Bigg(\frac{1}{2}\small\begin{bmatrix}
{{X}}_{t}\\[5pt]
{{U}}_{t}
\end{bmatrix}^\prime\begin{bmatrix}
L_t^{xx} & L_t^{xu}\\[5pt]
L_t^{ux}& L_t^{uu}
\end{bmatrix}\begin{bmatrix}
{{X}}_{t}\\[5pt]
{{U}}_{t}
\end{bmatrix}+\begin{bmatrix}
L_t^{x\theta}\\[5pt]
L_t^{ue}
\end{bmatrix}^\prime\begin{bmatrix}
{{X}}_{t}\\[5pt]
{{U}}_{t}
\end{bmatrix}\Bigg)+\Tr\Big({\Lambda}_{t+1}^\prime(F_t^x{{X}}_{t}+F^u_t{{U}}_{t}+F^\theta_t)\Big)
\\
&+\Tr\Big({\bar{V}}_{t}^\prime(\bar{G}_t^x{{X}}_{t}+\bar{G}^u_t{{U}}_{t}+\bar{G}^\theta_t)\Big)+\Tr\Big({{W}}_{t}^\prime({H}_t^x{{X}}_{t}+{H}^u_t{{U}}_{t}+{H}^\theta_t)\Big),\quad t=0,..,T{-}1, \\[5pt]
\normalsize
\bar{L}_T=&\Tr\left(\frac{1}{2}X_T^\prime L_T^{xx}X_T+(L_T^{x\theta})^\prime X_T\right)
+\Tr\Big({\bar{M}}_{T}^\prime(\bar{G}_T^x{{X}}_{T}+\bar{G}^\theta_T)\Big)\\
&+\Tr\Big({{N}}_{T}^\prime({H}_T^x{{X}}_{T}+{H}^\theta_T)\Big)
,\quad t=T.
\end{aligned}
\end{equation}
\normalsize

\noindent
Here, ${\Lambda}_{t}\in\mathbb{R}^{n\times r}$, $t=1,2,...,T$, denotes the costate (matrix) variables for $\boldsymbol{\overline\Sigma}(\boldsymbol{\xi}_{\boldsymbol{\theta}})$; $\bar{V}_{t}\in\mathbb{R}^{\bar{q}_t\times r}$ and ${W}_{t}\in\mathbb{R}^{s_t\times r}$, $t=0,1,...,T$, are the multipliers for the  constraints in $\boldsymbol{\overline\Sigma}(\boldsymbol{{\xi}}_{\boldsymbol{\theta}})$. 
Further define 
\begin{gather}
\Lambda_t^{\boldsymbol{\theta}}=\frac{\partial \boldsymbol{\lambda}_{t}^{\boldsymbol{\theta}}}{\partial \boldsymbol{\theta}}, \quad W_t^{\boldsymbol{\theta}}=\frac{\partial \boldsymbol{w}_{t}^{\boldsymbol{\theta}}}{\partial \boldsymbol{\theta}},\quad \bar{V}_t^{\boldsymbol{\theta}}=\frac{\partial \bar{\boldsymbol{v}}_{t}^{\boldsymbol{\theta}}}{\partial \boldsymbol{\theta}},
\end{gather}
with $\bar{\boldsymbol{v}}_{t}^{\boldsymbol{\theta}}$   in (\ref{def_vbar}). Then,  the Differential C-PMP in (\ref{diff-CPMP}) is exactly the Constrained Pontryagin Minimal Principle (C-PMP) for the auxiliary control system $\boldsymbol{\overline\Sigma}(\boldsymbol{{\xi}}_{\boldsymbol{\theta}})$  because 
\begin{equation}\label{cpmp_aux}
\begin{aligned}
&{X}^{\boldsymbol{\theta}}_{t+1}=\frac{\partial \bar{L}_t}{\partial {\Lambda}_{t+1}^{\boldsymbol{\theta}}}
=F_t^x{{X}}_{t}^{\boldsymbol{\theta}}+F^u_t{{U}}_{t}^{\boldsymbol{\theta}}+F^\theta_t \quad \text{and} \quad X_0=X_0^{\boldsymbol{\theta}},
\\
&{\Lambda}_{t}^{\boldsymbol{\theta}}=\frac{\partial \bar{L}_t}{\partial {X}_{t}^{\boldsymbol{\theta}}}=L_t^{xx}{X}_t^{\boldsymbol{\theta}}+L_t^{xu}{U}_t^{\boldsymbol{\theta}}+L_t^{x\theta}+(F_t^x)^{\prime}{\Lambda}_{t+1}^{\boldsymbol{\theta}}+ (\bar{G}_t^x)^{\prime}{\bar{V}}_{t}^{\boldsymbol{\theta}}+ (H_t^x)^{\prime}{W}_{t}^{\boldsymbol{\theta}}\quad \text{and}\\
&{\Lambda}_T^{\boldsymbol{\theta}}=\frac{\partial \bar{L}_t}{\partial {X}_T^{\boldsymbol{\theta}}}=H^{xx}_T{X}_T^{\boldsymbol{\theta}}+H^{xe}_T+(\bar{G}_T^x)^{\prime}{\bar{V}}_{T}^{\boldsymbol{\theta}}+({H}_T^x)^{\prime}{{W}}_{T}^{\boldsymbol{\theta}},\\
&\boldsymbol{0}=\frac{\partial \bar{L}_t}{\partial {U}_{t}^{\boldsymbol{\theta}}}=L_t^{uu}{U}_t^{\boldsymbol{\theta}}+L_t^{ux}{X}_t^{\boldsymbol{\theta}}+L_t^{u\theta}+(F_t^u)^{\prime}{\Lambda}_{t+1}^{\boldsymbol{\theta}}+ (\bar{G}_t^u)^{\prime}{\bar{V}}_{t}^{\boldsymbol{\theta}}+ (H_t^u)^{\prime}{W}_{t}^{\boldsymbol{\theta}},\\[2pt]
&H_t^xX_t^{\boldsymbol{\theta}}
+H_t^uU_t^{\boldsymbol{\theta}}
+H_t^\theta= \boldsymbol{0} \quad\text{and}\quad 
H_T^xX_T^{\boldsymbol{\theta}}
+H_T^\theta= \boldsymbol{0},\\[8pt]
&\bar{G}_t^xX_t^{\boldsymbol{\theta}}+\bar{G}_t^uU_t^{\boldsymbol{\theta}}+\bar{G}_t^\theta = \boldsymbol{0} \quad\text{and}\quad  \bar{G}_T^xX_T^{\boldsymbol{\theta}}+\bar{G}_T^\theta = \boldsymbol{0}.
\end{aligned}
\end{equation}
Note that in (\ref{cpmp_aux}), we have used the following matrix calculus \cite{athans1967matrix} and trace properties:
\begin{gather*}
\frac{\partial \Tr(AB)}{\partial A}= B^\prime, \quad \frac{\partial f(A)}{\partial A^\prime}= \left[\frac{\partial f(A)}{\partial A}\right]^\prime, \quad \frac{\partial \Tr(X^\prime HX)}{\partial X}=HX+H^\prime X,\\
\Tr(A)=\Tr(A^\prime),\quad \Tr(ABC)=\Tr(BCA)=\Tr(CAB),\quad \Tr(A+B)=\Tr(A)+\Tr(B).\nonumber
\end{gather*}

\smallskip
Next, we need to show that the second-order condition 
\begin{equation}
\begin{aligned}\label{secordercon_aux.1}
\sum_{t=0}^{T-1}\Tr\left(
\begin{bmatrix}
\Delta{X}_{t}\\[5pt]
\Delta{U}_{t}
\end{bmatrix}^\prime
\underbrace{
	\begin{bmatrix}
	\frac{\partial \bar{L}_t^2}{\partial X_t^{\boldsymbol{\theta}}\partial X_t^{\boldsymbol{\theta}}}& \frac{\partial \bar{L}_t^2}{\partial X_t^{\boldsymbol{\theta}}\partial U_t^{\boldsymbol{\theta}}}\\[5pt]
	\frac{\partial \bar{L}_t^2}{\partial U_t^{\boldsymbol{\theta}}\partial X_t^{\boldsymbol{\theta}}}& \frac{\partial \bar{L}_t^2}{\partial U_t^{\boldsymbol{\theta}}\partial U_t^{\boldsymbol{\theta}}}
	\end{bmatrix}
}_{
	\begin{bmatrix}
	L_t^{xx} & L_t^{xu}\\[5pt]
	L_t^{ux}& L_t^{uu}
	\end{bmatrix}
}
\begin{bmatrix}
\Delta{X}_{t}\\[5pt]
\Delta{U}_{t}
\end{bmatrix}
\right)+\Tr\left(\Delta{X}_{T}^\prime \underbrace{\begin{bmatrix}
	\frac{\partial \bar{L}_t^2}{\partial X_T^{\boldsymbol{\theta}}\partial
		X_T^{\boldsymbol{\theta}}}\end{bmatrix}}_{L_T^{xx}} \Delta{X}_{T}\right)>0,
\end{aligned}
\end{equation}
hold for any trajectory $\left\{
\Delta X_{0:T}, \Delta U_{0:T-1}
\right\}\neq \boldsymbol{0}$ 
satisfying
\begin{equation}\label{secordercon_aux.2}
\begin{aligned}
&\Delta{X}_{t+1}
=F_t^x{\Delta{X}}_{t}+F^u_t{\Delta{U}}_{t} \quad \text{and}\quad
\Delta X_{0}=\boldsymbol{0},\\
& 	\bar{G}_t^x\Delta X_t+\bar{G}_t^u\Delta U_t = \boldsymbol{0}  \quad\text{and}\quad  \bar{G}_T^x\Delta X_T= \boldsymbol{0},\\
&H_t^x\Delta X_t
+H_t^u\Delta U_t=\boldsymbol{0} \quad\text{and}\quad 
H_T^x\Delta X_T= \boldsymbol{0},
\end{aligned}
\end{equation}
In fact, this is true directly due to (\ref{secordercon.1}) and (\ref{bindingequation}) in Lemma \ref{Lemma1} and the strict complementarity in condition (iii) in Lemma \ref{result1} (note that ${\boldsymbol{\check{g}}}_t(\boldsymbol{x}_t,\boldsymbol{u}_t,\boldsymbol{\theta})={\boldsymbol{\bar{g}}}_t(\boldsymbol{x}_t,\boldsymbol{u}_t,\boldsymbol{\theta})$ and ${\boldsymbol{\check{g}}}_T(\boldsymbol{x}_T,\boldsymbol{\theta})={\boldsymbol{\bar{g}}}_T(\boldsymbol{x}_T,\boldsymbol{\theta})$ because of the strict complementarity). Therefore, with the C-PMP (\ref{cpmp_aux}) and  (\ref{secordercon_aux.1})-(\ref{secordercon_aux.2}) holding for $\left\{
X^{{\boldsymbol{{\theta}}}}_{0:T}, U^{{\boldsymbol{{\theta}}}}_{0:T-1}
\right\}$, we can conclude that $\left\{
X^{{\boldsymbol{{\theta}}}}_{0:T}, U^{{\boldsymbol{{\theta}}}}_{0:T-1}
\right\}$ is a {local unique}  minimizing trajectory to the auxiliary control system $\boldsymbol{\bar\Sigma}(\boldsymbol{\xi}_{\boldsymbol{\theta}})$ according to Lemma \ref{Lemma1}.

\medskip
Second, we prove that the {local unique} minimizing trajectory $\left\{
X^{{\boldsymbol{{\theta}}}}_{0:T}, U^{{\boldsymbol{{\theta}}}}_{0:T-1}
\right\}$ is also a \emph{global} one.

We note that any feasible trajectory $\left\{
X_{0:T}, U_{0:T-1}
\right\}$ that satisfies all  constraints (dynamics, path and final constraints) in the auxiliary control system  $\boldsymbol{\overline\Sigma}(\boldsymbol{\xi}_{\boldsymbol{\theta}})$ can be written as 
\begin{equation}\label{pf_feasibletraj}
\left\{
X_{0:T}, U_{0:T-1}
\right\}=\left\{
X_{0:T}^{\boldsymbol{\theta}}, U_{0:T-1}^{\boldsymbol{\theta}}
\right\}+\left\{
\Delta X_{0:T}, \Delta U_{0:T-1}
\right\},
\end{equation}
with   $\left\{
\Delta X_{0:T}, \Delta U_{0:T-1}
\right\}$ satisfying the conditions in (\ref{secordercon_aux.2}). Let
\begin{equation}\label{pf_important}
\begin{aligned}
&\bar{J}(X_{0:T}, U_{0:T-1})-\bar{J}(X_{0:T}^{\boldsymbol{\theta}}, U_{0:T-1}^{\boldsymbol{\theta}})\\
=&\small\Tr\sum_{t=0}^{T{-}1}
\left(\frac{1}{2}\begin{bmatrix}
\Delta X_t\\
\Delta U_t
\end{bmatrix}^\prime\begin{bmatrix}
L_t^{xx}& L_t^{xu}\\
L_t^{ux}& L_t^{uu}
\end{bmatrix}\begin{bmatrix}
\Delta X_t\\
\Delta U_t
\end{bmatrix}+\begin{bmatrix}
X_t^{\boldsymbol{\theta}}\\
U_t^{\boldsymbol{\theta}}
\end{bmatrix}^\prime\begin{bmatrix}
L_t^{xx}& L_t^{xu}\\
L_t^{ux}& L_t^{uu}
\end{bmatrix}\begin{bmatrix}
\Delta X_t\\
\Delta U_t
\end{bmatrix}+
\begin{bmatrix}
L_t^{x\theta}\\
L_t^{u\theta}
\end{bmatrix}^\prime\begin{bmatrix}
\Delta X_t\\
\Delta U_t
\end{bmatrix}
\right)\\
&+\Tr\left(\frac{1}{2} \Delta X_T^\prime L_T^{xx} \Delta X_T+  (X_T^{\boldsymbol{\theta}})^\prime L_T^{xx} \Delta X_T+ (L_T^{x\theta})^\prime \Delta X_T \right).
\end{aligned}
\end{equation}

Based on (\ref{cpmp_aux}), the following term in (\ref{pf_important}) can be simplified to
\begin{equation}\label{pfeliminate.1}
\begin{aligned}
&\begin{bmatrix}
\Delta X_t\\[2pt]
\Delta U_t
\end{bmatrix}^\prime
\left(
\begin{bmatrix}
L_t^{xx}& L_t^{xu}\\[2pt]
L_t^{ux}& L_t^{uu}
\end{bmatrix}\begin{bmatrix}
X_t^{\boldsymbol{\theta}}\\[2pt]
U_t^{\boldsymbol{\theta}}
\end{bmatrix}
+\begin{bmatrix}
L_t^{x\theta}\\[2pt]
L_t^{u\theta}
\end{bmatrix}
\right)\\[2pt]
=&\begin{bmatrix}
\Delta X_t\\[2pt]
\Delta U_t
\end{bmatrix}^\prime
\begin{bmatrix}
-(F_{t}^{x})^\prime\Lambda_{t+1}^{\boldsymbol{\theta}}- (\bar{G}_t^x)^{\prime}{\bar{V}}_{t}^{\boldsymbol{\theta}}- (H_t^x)^{\prime}{W}_{t}^{\boldsymbol{\theta}}+\Lambda_{t}^{\boldsymbol{\theta}}\\[2pt]
-(F_t^u)^{\prime}{\Lambda}_{t+1}^{\boldsymbol{\theta}}- (\bar{G}_t^u)^{\prime}{\bar{V}}_{t}^{\boldsymbol{\theta}}- (H_t^u)^{\prime}{W}_{t}^{\boldsymbol{\theta}}
\end{bmatrix}\\
=&-({\Lambda}_{t+1}^{\boldsymbol{\theta}})^\prime F_t^x\Delta X_t
-\cancel{(\bar{V}_{t}^{\boldsymbol{\theta}})^\prime \bar{G}_t^x\Delta X_t}-\cancel{({W}_{t}^{\boldsymbol{\theta}})^\prime {H}_t^x\Delta X_t}+
({\Lambda}_{t}^{\boldsymbol{\theta}})^\prime\Delta X_t\\
&-({\Lambda}_{t+1}^{\boldsymbol{\theta}})^\prime F_t^u\Delta U_t
-\cancel{(\bar{V}_{t}^{\boldsymbol{\theta}})^\prime \bar{G}_t^u\Delta U_t}-\cancel{({W}_{t}^{\boldsymbol{\theta}})^\prime {H}_t^u\Delta U_t}\\
=&\underbrace{-({\Lambda}_{t+1}^{\boldsymbol{\theta}})^\prime F_t^x\Delta X_t-({\Lambda}_{t+1}^{\boldsymbol{\theta}})^\prime F_t^u\Delta U_t}_{-({\Lambda}_{t+1}^{\boldsymbol{\theta}})^\prime\Delta X_{t+1}}+({\Lambda}_{t}^{\boldsymbol{\theta}})^\prime\Delta X_{t}=-({\Lambda}_{t+1}^{\boldsymbol{\theta}})^\prime\Delta X_{t+1}+({\Lambda}_{t}^{\boldsymbol{\theta}})^\prime\Delta X_{t}   
\end{aligned}
\end{equation}
where the cancellations in the last three lines  are due to (\ref{secordercon_aux.2}). Also based on (\ref{cpmp_aux}), the following term in (\ref{pf_important}) can be simplified to
\begin{equation}\label{pfeliminate.2}
\begin{aligned}
&\left( (X_T^{\boldsymbol{\theta}})^\prime L_T^{xx} + (L_T^{x\theta})^\prime \right)\Delta X_T\\=&-\cancel{({\bar{V}}_{T}^{\boldsymbol{\theta}})^{\prime}\bar{G}_T\Delta X_T}
-\cancel{({{W}}_{T}^{\boldsymbol{\theta}})^\prime{H}_T^x\Delta X_T}+(\Lambda_T^{\boldsymbol{\theta}})^\prime\Delta X_T\\
=&(\Lambda_T^{\boldsymbol{\theta}})^\prime\Delta X_T
\end{aligned}
\end{equation}
where the cancellation here is due to  (\ref{secordercon_aux.2}).

Then, based on (\ref{pfeliminate.1}) and (\ref{pfeliminate.2}),  (\ref{pf_important}) is simplified to
\begin{equation}
\begin{aligned}
&\bar{J}(X_{0:T}, U_{0:T-1})-\bar{J}(X_{0:T}^{\boldsymbol{\theta}}, U_{0:T-1}^{\boldsymbol{\theta}})\\
=&\small\Tr\sum_{t=0}^{T{-}1}
\left(\frac{1}{2}\begin{bmatrix}
\Delta X_t\\
\Delta U_t
\end{bmatrix}^\prime\begin{bmatrix}
L_t^{xx}& L_t^{xu}\\
L_t^{ux}& L_t^{uu}
\end{bmatrix}\begin{bmatrix}
\Delta X_t\\
\Delta U_t
\end{bmatrix}+\begin{bmatrix}
X_t^{\boldsymbol{\theta}}\\
U_t^{\boldsymbol{\theta}}
\end{bmatrix}^\prime\begin{bmatrix}
L_t^{xx}& L_t^{xu}\\
L_t^{ux}& L_t^{uu}
\end{bmatrix}\begin{bmatrix}
\Delta X_t\\
\Delta U_t
\end{bmatrix}+
\begin{bmatrix}
L_t^{x\theta}\\
L_t^{u\theta}
\end{bmatrix}^\prime\begin{bmatrix}
\Delta X_t\\
\Delta U_t
\end{bmatrix}
\right)\\
&+\Tr\left(\frac{1}{2} \Delta X_T^\prime L_T^{xx} \Delta X_T+  (X_T^{\boldsymbol{\theta}})^\prime L_T^{xx} \Delta X_T+ (L_T^{x\theta})^\prime \Delta X_T \right)\\
=&\Tr\sum_{t=0}^{T{-}1}
\left(\frac{1}{2}\begin{bmatrix}
\Delta X_t\\
\Delta U_t
\end{bmatrix}^\prime\begin{bmatrix}
L_t^{xx}& L_t^{xu}\\
L_t^{ux}& L_t^{uu}
\end{bmatrix}\begin{bmatrix}
\Delta X_t\\
\Delta U_t
\end{bmatrix}-({\Lambda}_{t+1}^{\boldsymbol{\theta}})^\prime\Delta X_{t+1}+({\Lambda}_{t}^{\boldsymbol{\theta}})^\prime\Delta X_{t}
\right)\\
&+\Tr\left(\frac{1}{2} \Delta X_T^\prime L_T^{xx} \Delta X_T+(\Lambda_T^{\boldsymbol{\theta}})^\prime\Delta X_T \right)\\
=&\Tr\sum_{t=0}^{T{-}1}
\left(\frac{1}{2}\begin{bmatrix}
\Delta X_t\\
\Delta U_t
\end{bmatrix}^\prime\begin{bmatrix}
L_t^{xx}& L_t^{xu}\\
L_t^{ux}& L_t^{uu}
\end{bmatrix}\begin{bmatrix}
\Delta X_t\\
\Delta U_t
\end{bmatrix}
\right)+\Tr\left(\frac{1}{2} \Delta X_T^\prime L_T^{xx} \Delta X_T \right),
\end{aligned}
\end{equation}
where the last line is because (note $\Delta X_0=\boldsymbol{0}$ in (\ref{secordercon_aux.2}))
\begin{equation}
\Tr\sum_{t=0}^{T{-}1}
\left(-({\Lambda}_{t+1}^{\boldsymbol{\theta}})^\prime\Delta X_{t+1}+({\Lambda}_{t}^{\boldsymbol{\theta}})^\prime\Delta X_{t}
\right)+\Tr\left((\Lambda_T^{\boldsymbol{\theta}})^\prime\Delta X_T \right)=\Tr\left((\Lambda_0^{\boldsymbol{\theta}})^\prime\Delta X_0\right)=0.\nonumber
\end{equation}

Since
\begin{equation}
\Tr\sum_{t=0}^{T{-}1}
\left(\frac{1}{2}\begin{bmatrix}
\Delta X_t\\
\Delta U_t
\end{bmatrix}^\prime\begin{bmatrix}
L_t^{xx}& L_t^{xu}\\
L_t^{ux}& L_t^{uu}
\end{bmatrix}\begin{bmatrix}
\Delta X_t\\
\Delta U_t
\end{bmatrix}
\right)+\Tr\left(\frac{1}{2} \Delta X_T^\prime L_T^{xx} \Delta X_T \right)\geq 0
\end{equation}
due to  (\ref{secordercon_aux.1}) for all $\left\{
\Delta X_{0:T}, \Delta U_{0:T-1}
\right\}$ satisfying (\ref{secordercon_aux.2}), therefore
\begin{equation}
\bar{J}(X_{0:T}, U_{0:T-1})-\bar{J}(X_{0:T}^{\boldsymbol{\theta}}, U_{0:T-1}^{\boldsymbol{\theta}})\geq 0.
\end{equation}
for any feasible trajectory $\left\{
X_{0:T}, U_{0:T-1}
\right\}$ in (\ref{pf_feasibletraj}). This concludes that the {local unique} minimizing trajectory $\left\{
X^{{\boldsymbol{{\theta}}}}_{0:T}, U^{{\boldsymbol{{\theta}}}}_{0:T}
\right\}$ is also a \emph{global} one.

In sum of the two proof steps, the assertion that  the  trajectory in (\ref{difftraj}), i.e.,
\begin{equation}
\frac{\partial\boldsymbol{\xi}_{{\boldsymbol{{\theta}}}} }{\partial \boldsymbol{\theta}}=\left\{
X^{{\boldsymbol{{\theta}}}}_{0:T}, U^{{\boldsymbol{{\theta}}}}_{0:T-1}
\right\}, \nonumber
\end{equation}
is a  globally unique minimizing trajectory to the auxiliary control system $\boldsymbol{\overline\Sigma}(\boldsymbol{\xi}_{\boldsymbol{\theta}})$ in (\ref{equ_aux}) follows. This completes the proof of Theorem \ref{theorem1}.

\qed

\section{Proof of Theorem \ref{theorem2}} \label{pf_theorem2}

For the unconstrained optimal control system ${\boldsymbol{\Sigma}}(\boldsymbol{\theta},\gamma)$ in (\ref{equ_oc_penality}), we define its Hamiltonian below:
\begin{equation}\label{hamil-penalty-oc}
\begin{aligned}
\hat{L}_t&=c_t(\boldsymbol{x}_t,\boldsymbol{u}_t,\boldsymbol{\theta})+\boldsymbol{\lambda}^\prime_{t{+}1}\boldsymbol{f}(\boldsymbol{x}_t,\boldsymbol{u}_t,\boldsymbol{\theta})- \gamma\sum_{i=1}^{q_t}\ln\big({-}g_{t,i}(\boldsymbol{x}_t,\boldsymbol{u}_t,{\boldsymbol{\theta}})\big)
{+}\frac{1}{2\gamma}\sum_{i=1}^{s_t}\big(h_{t,i}(\boldsymbol{x}_t,\boldsymbol{u}_t, {\boldsymbol{\theta}})\big)^2
\\
\hat L_T&=c_T(\boldsymbol{x}_T,\boldsymbol{\theta})-\gamma\sum_{i=1}^{q_T}\ln\big({-}g_{T,i}(\boldsymbol{x}_T,{\boldsymbol{\theta}})\big)
{+}\frac{1}{2\gamma}\sum_{i=1}^{s_T}\big(h_{T,i}(\boldsymbol{x}_T, {\boldsymbol{\theta}})\big)^2 .
\end{aligned}
\end{equation}
with $t=0,1,\cdots, T-1$.

\subsection{Proof of Claim  (a)} \label{pf_theorem2_(a)}

We first modify the C-PMP condition  (\ref{CPMP}) for  the constrained optimal control system $\boldsymbol{\Sigma}(\boldsymbol{\theta})$ into the following set of equations:
\begin{equation}\label{CPMP-modify}
\begin{aligned}
&\boldsymbol{x}_{t+1}=\boldsymbol{f}(\boldsymbol{x}_{t},\boldsymbol{u}_{t},\boldsymbol{\theta})
\quad\text{and}\quad
\boldsymbol{x}_{0}=\boldsymbol{x}_{0}({\boldsymbol{\theta}}),
\\[3pt]
&\boldsymbol{\lambda}_{t}=L_{t}^x(\boldsymbol{x}_{t},\boldsymbol{u}_{t},\boldsymbol{\lambda}_{t+1},\boldsymbol{v}_{t},\boldsymbol{w}_{t},\boldsymbol{\theta})
\quad \text{and}\quad
\boldsymbol{\lambda}_{T}=L_{T}^x(\boldsymbol{x}_{t},\boldsymbol{v}_{t},\boldsymbol{w}_{t},\boldsymbol{\theta}),
\\[3pt]
&\boldsymbol{0}=
L_{t}^u(\boldsymbol{x}_{t},\boldsymbol{u}_{t},\boldsymbol{\lambda}_{t+1},\boldsymbol{v}_{t},\boldsymbol{w}_{t},\boldsymbol{\theta}),
\\[3pt]
&{h}_{t,i}(\boldsymbol{x}_{t},\boldsymbol{u}_{t},\boldsymbol{\theta})=w_{t,i}\gamma,\quad i=1,2,\cdots, s_t,\\
&
{h}_{T,i}(\boldsymbol{x}_{T},\boldsymbol{\theta})=w_{T,i}\gamma,\quad i=1,2,\cdots, s_T,
\\[3pt]
&{v}_{t,i}\,{g}_{t,i}(\boldsymbol{x}_{t},\boldsymbol{u}_{t},\boldsymbol{\theta})=-\gamma,\quad i=1,2,\cdots, q_t,
\\[3pt]
&
{v}_{T,i}\,{g}_{T,i}(\boldsymbol{x}_{T},\boldsymbol{\theta})=-\gamma, \quad i=1,2,\cdots, q_T,
\end{aligned}
\end{equation}
where the first three equations are the same with the those in  (\ref{CPMP}) and only the last two lines of equations are modified by adding some perturbation terms related to $\gamma$.

Now, one can view that 
the  parameters  $(\boldsymbol{\theta},{\gamma})$ jointly determine  $\boldsymbol{\xi}=\{\boldsymbol{x}_{0:T},\boldsymbol{u}_{0:T-1}\}$, $\boldsymbol{\lambda}_{1:T}$, $\boldsymbol{v}_{0:T}$, and $\boldsymbol{w}_{0:T}$ through the implicit equations in (\ref{CPMP-modify}). Also, one can note that by letting $\gamma=0$ and $\boldsymbol{\theta}=\boldsymbol{\bar\theta}$, the above equations in (\ref{CPMP-modify}) coincide with the C-PMP condition  (\ref{CPMP}) for $\boldsymbol{\Sigma}(\boldsymbol{\bar\theta})$. Thus, given that  the conditions (i)-(iii) in Lemma \ref{result1} hold for $\boldsymbol{\Sigma}(\boldsymbol{\bar\theta})$,   one  can  readily apply the implicit function theorem \cite{rudin1976principles} to (\ref{CPMP-modify}) in a neighborhood of $(\boldsymbol{\bar\theta},{0})$ and make the following assertion (its proof can directly follow the proof for Lemma \ref{result1} (i.e., the first-order sensitivity result) with little change):

\emph{For any $(\boldsymbol{\theta},{\gamma})$ within a  neighborhood of $(\boldsymbol{\bar\theta},{0})$, there exists a unique once-continuously differentiable function $\left(\boldsymbol{\xi}_{(\boldsymbol{\theta},\gamma)}, \boldsymbol{\lambda}_{0:T}^{(\boldsymbol{\theta},\gamma)}, \boldsymbol{v}_{0:T}^{(\boldsymbol{\theta},\gamma)},\boldsymbol{w}_{0:T}^{(\boldsymbol{\theta},\gamma)}\right)$, which satisfies (\ref{CPMP-modify}) and
	\begin{equation}
	\left(
	\boldsymbol{\xi}_{(\boldsymbol{\theta},\gamma)},
	\boldsymbol{\lambda}_{0:T}^{({\boldsymbol{\theta}},\gamma)}, \boldsymbol{v}_{0:T}^{({\boldsymbol{\theta}},\gamma)}, \boldsymbol{w}_{0:T}^{({\boldsymbol{\theta}},\gamma)}
	\big)=\big(\boldsymbol{\xi}_{{\boldsymbol{\bar{\theta}}}},\boldsymbol{\lambda}_{1:T}^{{\boldsymbol{{\bar\theta}}}},\boldsymbol{v}_{0:T}^{{\boldsymbol{{\bar\theta}}}},\boldsymbol{w}_{0:T}^{{\boldsymbol{{\bar\theta}}}}\right) \quad \text{when} \quad  (\boldsymbol{\theta},{\gamma})=(\boldsymbol{\bar\theta},{0}).
	\end{equation}
}

\medskip
With the above claim, in what follows, we will prove that  for any $(\boldsymbol{\theta},{\gamma})$ near $(\boldsymbol{\bar\theta},{0})$ additionally with $\gamma>0$,   $\boldsymbol{\xi}_{(\boldsymbol{\theta},\gamma)}$ is  a local isolated  minimizing trajectory to the unconstrained optimal control system ${\boldsymbol{\Sigma}}(\boldsymbol{\theta},\gamma)$ in (\ref{equ_oc_penality}). First, we need to show that such $\boldsymbol{\xi}_{(\boldsymbol{\theta},\gamma)}$ will make  ${\boldsymbol{\Sigma}}(\boldsymbol{\theta},\gamma)$   well-defined, which is the second part of Claim (a), rewritten below
\begin{equation}\label{welldefine}
\begin{aligned}
&{g}_{t,i}\left(\boldsymbol{x}_{t}^{(\boldsymbol{\theta},\gamma)},\boldsymbol{u}_{t}^{(\boldsymbol{\theta},\gamma)},\boldsymbol{\theta}\right)<0,\quad i=1,2,\cdots, q_t, \quad \text{and} \quad\\
& {g}_{T,i}\left(\boldsymbol{x}_{T}^{(\boldsymbol{\theta},\gamma)},\boldsymbol{\theta}\right)<0,\quad i=1,2,\cdots, q_T.
\end{aligned}
\end{equation}
In fact, such an assertion always holds because  the strict complementary for $\boldsymbol{\Sigma}(\boldsymbol{\bar\theta})$ from Lemma \ref{result1}. Specifically, for any $i=1,2,\cdots, q_t$, if ${g}_{t,i}(\boldsymbol{x}_{t}^{{\boldsymbol{\bar\theta}}},\boldsymbol{u}_{t}^{{\boldsymbol{\bar\theta}}},{{\boldsymbol{\bar\theta}}})<0$, from  continuity of $g_{t,i}$ and $\boldsymbol{\xi}_{(\boldsymbol{\theta}, \gamma)}$
\begin{equation*}
{g}_{t,i}(\boldsymbol{x}_{t}^{(\boldsymbol{\theta},\gamma)},\boldsymbol{u}_{t}^{(\boldsymbol{\theta},\gamma)},\boldsymbol{\theta})\rightarrow{g}_{t,i}(\boldsymbol{x}_{t}^{{\boldsymbol{\bar\theta}}},\boldsymbol{u}_{t}^{{\boldsymbol{\bar\theta}}},{{\boldsymbol{\bar\theta}}})<0  \quad as\quad (\boldsymbol{\theta},\gamma) \rightarrow  (\boldsymbol{\bar\theta},0),
\end{equation*}
thus ${g}_{t,i}(\boldsymbol{x}_{t}^{(\boldsymbol{\theta},\gamma)},\boldsymbol{u}_{t}^{(\boldsymbol{\theta},\gamma)},\boldsymbol{\theta})<0$ for any $(\boldsymbol{\theta},{\gamma})$ near $(\boldsymbol{\bar\theta},{0})$ with $\gamma>0$;  if ${g}_{t,i}(\boldsymbol{x}_{t}^{{\boldsymbol{\bar\theta}}},\boldsymbol{u}_{t}^{{\boldsymbol{\bar\theta}}},{{\boldsymbol{\bar\theta}}})=0$ and $v_{t,i}^{{\boldsymbol{\bar\theta}}}>0$ (due to strict complementarity), from  continuity of $\boldsymbol{v}_t^{(\boldsymbol{\theta}, \gamma)}$,
\begin{equation}
{v}_{t,i}^{(\boldsymbol{\theta},\gamma)}
\rightarrow
{v_{t,i}^{{\boldsymbol{\bar\theta}}}}>0 
\quad as\quad (\boldsymbol{\theta},\gamma) \rightarrow  (\boldsymbol{\bar\theta},0),
\end{equation}
thus ${v}_{t,i}^{(\boldsymbol{\theta},\gamma)}>0$ for $(\boldsymbol{\theta},{\gamma})$ near $(\boldsymbol{\bar\theta},{0})$ with $\gamma>0$, and also due to (\ref{CPMP-modify}),
$
{g}_{t,i}(\boldsymbol{x}_{t}^{(\boldsymbol{\theta},\gamma)},\boldsymbol{u}_{t}^{(\boldsymbol{\theta},\gamma)},\boldsymbol{\theta})=-\frac{\gamma}{{v}_{t,i}^{(\boldsymbol{\theta},\gamma)}}<0$  for $(\boldsymbol{\theta},{\gamma})$ near $(\boldsymbol{\bar\theta},{0})$ with $\gamma>0$. So, for either case, the first inequality in (\ref{welldefine}) always holds. Similar proof procedure also applies to prove the second inequality in (\ref{welldefine}). In sum, we conclude that $\boldsymbol{\xi}_{(\boldsymbol{\theta},\gamma)}$ satisfies (\ref{welldefine}) and thus makes the ${\boldsymbol{\Sigma}}(\boldsymbol{\theta},\gamma)$   well-defined  for any $(\boldsymbol{\theta},{\gamma})$ near $(\boldsymbol{\bar\theta},{0})$ with $\gamma>0$. This completes the second part of Claim (a).

\bigskip
From now on, we prove that for any $(\boldsymbol{\theta},{\gamma})$ near $(\boldsymbol{\bar\theta},{0})$ with $\gamma>0$,  $\boldsymbol{\xi}_{(\boldsymbol{\theta},\gamma)}$ is  a local isolated  minimizing trajectory to the unconstrained optimal control system ${\boldsymbol{\Sigma}}(\boldsymbol{\theta},\gamma)$ in (\ref{equ_oc_penality}).  From the last four equations in (\ref{CPMP-modify}), we  solve
\begin{equation}
\begin{aligned}
&w^{(\boldsymbol{\theta},\gamma)}_{t,i}=\frac{{h}_{t,i}(\boldsymbol{x}_{t}^{(\boldsymbol{\theta},\gamma)},\boldsymbol{u}_{t}^{(\boldsymbol{\theta},\gamma)},\boldsymbol{\theta})}{\gamma},\qquad w^{(\boldsymbol{\theta},\gamma)}_{T,i}=\frac{{h}_{T,i}(\boldsymbol{x}_{T}^{(\boldsymbol{\theta},\gamma)},\boldsymbol{\theta})}{\gamma},\\
&v^{(\boldsymbol{\theta},\gamma)}_{t,i}=-\frac{\gamma}{{g}_{t,i}(\boldsymbol{x}_{t}^{(\boldsymbol{\theta},\gamma)},\boldsymbol{u}_{t}^{(\boldsymbol{\theta},\gamma)},\boldsymbol{\theta})},\qquad v^{(\boldsymbol{\theta},\gamma)}_{T,i}=-\frac{\gamma}{{g}_{T,i}(\boldsymbol{x}_{T}^{(\boldsymbol{\theta},\gamma)},\boldsymbol{\theta})},
\end{aligned}
\end{equation}
and plug them into the first three equations in (\ref{CPMP-modify}), then one will find that the obtained equations are exactly the Pontryagin Maximum/Minimum Principle (PMP) for the unconstrained optimal control system ${\boldsymbol{\Sigma}}(\boldsymbol{\theta},\gamma)$ with its Hamiltonian already defined in (\ref{hamil-penalty-oc}), that is to say,
\begin{eqnarray}
\begin{aligned}
&\boldsymbol{x}_{t+1}^{(\boldsymbol{\theta},\gamma)}=\boldsymbol{f}(\boldsymbol{x}_{t}^{(\boldsymbol{\theta},\gamma)},\boldsymbol{u}_{t}^{(\boldsymbol{\theta},\gamma)},{\boldsymbol{\theta}})
\quad\text{and}\quad
\boldsymbol{x}_{0}^{(\boldsymbol{\theta},\gamma)}=\boldsymbol{x}_{0}({\boldsymbol{\theta}}),
\\[3pt]
&\boldsymbol{\lambda}_{t}^{(\boldsymbol{\theta},\gamma)}=\hat L_{t}^x(\boldsymbol{x}_{t}^{(\boldsymbol{\theta},\gamma)},\boldsymbol{u}_{t}^{(\boldsymbol{\theta},\gamma)},\boldsymbol{\lambda}_{t+1}^{(\boldsymbol{\theta},\gamma)},{(\boldsymbol{\theta},\gamma)}),
\\[3pt]
&
\boldsymbol{\lambda}_{T}^{(\boldsymbol{\theta},\gamma)}=\hat L_{T}^x(\boldsymbol{x}_{t}^{(\boldsymbol{\theta},\gamma)},{(\boldsymbol{\theta},\gamma)}),
\\[3pt]
&\boldsymbol{0}=
\hat L_{t}^u(\boldsymbol{x}_{t}^{(\boldsymbol{\theta},\gamma)},\boldsymbol{u}_{t}^{(\boldsymbol{\theta},\gamma)},\boldsymbol{\lambda}_{t+1}^{(\boldsymbol{\theta},\gamma)},{(\boldsymbol{\theta},\gamma)}),
\end{aligned}
\end{eqnarray}
indicating that $\boldsymbol{\xi}_{(\boldsymbol{\theta},\gamma)}=\left\{\boldsymbol{x}_{0:T}^{(\boldsymbol{\theta},\gamma)}, \boldsymbol{u}_{0:T-1}^{(\boldsymbol{\theta},\gamma)}\right\}$ already satisfies the PMP condition for unconstrained optimal control system ${\boldsymbol{\Sigma}}(\boldsymbol{\theta},\gamma)$. To show $\boldsymbol{\xi}_{(\boldsymbol{\theta},\gamma)}=\left\{\boldsymbol{x}_{0:T}^{(\boldsymbol{\theta},\gamma)}, \boldsymbol{u}_{0:T-1}^{(\boldsymbol{\theta},\gamma)}\right\}$ is a local isolated  minimizing trajectory to  ${\boldsymbol{\Sigma}}(\boldsymbol{\theta},\gamma)$ for any $(\boldsymbol{\theta},\gamma)$ near ($\boldsymbol{\bar{\theta}},0$) with $\gamma>0$, we only need to verify its second-order condition as stated in (\ref{secordercon.1})-(\ref{bindingequation}) in Lemma \ref{Lemma1}, which is presented next. In the remainder of  proof, for convenience of notation, all derivatives  are evaluated at $(\boldsymbol{\theta},\gamma)$ (or $\boldsymbol{\xi}_{(\boldsymbol{\theta},\gamma)}$) unless otherwise stated.

Before proceeding, we show two facts (easy to prove) about the   second-order derivatives of Hamiltonian $\hat L_t$ and $\hat L_T$ in (\ref{hamil-penalty-oc}). 
First, 
\begin{multline}
\begin{bmatrix}
\hat L_t^{xx} &\hat L_t^{xu}\\
\hat L_t^{ux} &\hat  L_t^{uu}
\end{bmatrix}=
\begin{bmatrix}
L_t^{xx} & L_t^{xu}\\
L_t^{ux} & L_t^{uu}
\end{bmatrix}+\\
\begin{bmatrix}
\sum\limits_{i=1}^{q_t}\frac{\gamma}{g_{t,i}^2}\frac{\partial g_{t,i}^\prime}{\partial \boldsymbol{x}_t}\frac{\partial g_{t,i}}{\partial \boldsymbol{x}_t}+\sum\limits_{i=1}^{s_t}\frac{1}{\gamma}\frac{\partial h_{t,i}^\prime}{\partial \boldsymbol{x}_t}\frac{\partial h_{t,i}}{\partial \boldsymbol{x}_t} & 
\sum\limits_{i=1}^{q_t}\frac{\gamma}{g_{t,i}^2}\frac{\partial g_{t,i}^\prime}{\partial \boldsymbol{x}_t}\frac{\partial g_{t,i}}{\partial \boldsymbol{u}_t}+\sum\limits_{i=1}^{s_t}\frac{1}{\gamma}\frac{\partial h_{t,i}^\prime}{\partial \boldsymbol{x}_t}\frac{\partial h_{t,i}}{\partial \boldsymbol{u}_t} 
\\[4pt]
\sum\limits_{i=1}^{q_t}\frac{\gamma}{g_{t,i}^2}\frac{\partial g_{t,i}^\prime}{\partial \boldsymbol{u}_t}\frac{\partial g_{t,i}}{\partial \boldsymbol{x}_t}+\sum\limits_{i=1}^{s_t}\frac{1}{\gamma}\frac{\partial h_{t,i}^\prime}{\partial \boldsymbol{u}_t}\frac{\partial h_{t,i}}{\partial \boldsymbol{x}_t}& \sum\limits_{i=1}^{q_t}\frac{\gamma}{g_{t,i}^2}\frac{\partial g_{t,i}^\prime}{\partial \boldsymbol{u}_t}\frac{\partial g_{t,i}}{\partial \boldsymbol{u}_t}+\sum\limits_{i=1}^{s_t}\frac{1}{\gamma}\frac{\partial h_{t,i}^\prime}{\partial \boldsymbol{u}_t}\frac{\partial h_{t,i}}{\partial \boldsymbol{u}_t}
\end{bmatrix},
\end{multline}
and
\begin{equation}
\hat L_T^{xx} =
L_T^{xx}
+
\sum_{i=1}^{q_T}\frac{\gamma}{g_{t,i}^2}\frac{\partial g_{T,i}^\prime}{\partial \boldsymbol{x}_T}\frac{\partial g_{T,i}}{\partial \boldsymbol{x}_T}+\small\sum_{i=1}^{s_T}\frac{1}{\gamma}\frac{\partial h_{T,i}^\prime}{\partial \boldsymbol{x}_T}\frac{\partial h_{T,i}}{\partial \boldsymbol{x}_T},
\end{equation} respectively. Second, given  any $\boldsymbol{x}$ and $\boldsymbol{u}$ with appropriate dimensions, one has
\begin{align}
&\begin{bmatrix}
\boldsymbol{x}\\
\boldsymbol{u}
\end{bmatrix}^\prime
\begin{bmatrix}
\hat L_t^{xx} &\hat L_t^{xu}\\
\hat L_t^{ux} &\hat  L_t^{uu}
\end{bmatrix}
\begin{bmatrix}
\boldsymbol{x}\\
\boldsymbol{u}
\end{bmatrix}=
\begin{bmatrix}
\boldsymbol{x}\\
\boldsymbol{u}
\end{bmatrix}^\prime
\begin{bmatrix}
L_t^{xx} & L_t^{xu}\\
L_t^{ux} & L_t^{uu}
\end{bmatrix}
\begin{bmatrix}
\boldsymbol{x}\\
\boldsymbol{u}
\end{bmatrix}+\nonumber\\
&\quad\begin{bmatrix}
\boldsymbol{x}\\
\boldsymbol{u}
\end{bmatrix}^\prime
\begin{bmatrix}
\sum\limits_{i=1}^{q_t}\frac{\gamma}{g_{t,i}^2}\frac{\partial g_{t,i}^\prime}{\partial \boldsymbol{x}_t}\frac{\partial g_{t,i}}{\partial \boldsymbol{x}_t}+\sum\limits_{i=1}^{s_t}\frac{1}{\gamma}\frac{\partial h_{t,i}^\prime}{\partial \boldsymbol{x}_t}\frac{\partial h_{t,i}}{\partial \boldsymbol{x}_t} & 
\sum\limits_{i=1}^{q_t}\frac{\gamma}{g_{t,i}^2}\frac{\partial g_{t,i}^\prime}{\partial \boldsymbol{x}_t}\frac{\partial g_{t,i}}{\partial \boldsymbol{u}_t}{+}\sum_{i=1}^{s_t}\frac{1}{\gamma}\frac{\partial h_{t,i}^\prime}{\partial \boldsymbol{x}_t}\frac{\partial h_{t,i}}{\partial \boldsymbol{u}_t} 
\\[4pt]
\sum\limits_{i=1}^{q_t}\frac{\gamma}{g_{t,i}^2}\frac{\partial g_{t,i}^\prime}{\partial \boldsymbol{u}_t}\frac{\partial g_{t,i}}{\partial \boldsymbol{x}_t}{+}\sum\limits_{i=1}^{s_t}\frac{1}{\gamma}\frac{\partial h_{t,i}^\prime}{\partial \boldsymbol{u}_t}\frac{\partial h_{t,i}}{\partial \boldsymbol{x}_t}& \sum\limits_{i=1}^{q_t}\frac{\gamma}{g_{t,i}^2}\frac{\partial g_{t,i}^\prime}{\partial \boldsymbol{u}_t}\frac{\partial g_{t,i}}{\partial \boldsymbol{u}_t}{+}\sum\limits_{i=1}^{s_t}\frac{1}{\gamma}\frac{\partial h_{t,i}^\prime}{\partial \boldsymbol{u}_t}\frac{\partial h_{t,i}}{\partial \boldsymbol{u}_t}
\end{bmatrix}
\begin{bmatrix}
\boldsymbol{x}\\
\boldsymbol{u}
\end{bmatrix}\nonumber\\
&=\begin{bmatrix}
\boldsymbol{x}\\
\boldsymbol{u}
\end{bmatrix}^\prime
\begin{bmatrix}
L_t^{xx} & L_t^{xu}\\
L_t^{ux} & L_t^{uu}
\end{bmatrix}
\begin{bmatrix}
\boldsymbol{x}\\
\boldsymbol{u}
\end{bmatrix}+ 
\small\sum\limits_{i=1}^{q_t}\frac{\gamma}{g_{t,i}^2}\Big(
\frac{\partial g_{t,i}}{\partial \boldsymbol{x}_t}\boldsymbol{x}+\frac{\partial g_{t,i}}{\partial \boldsymbol{u}}\boldsymbol{u}
\Big)^2
+\sum\limits_{i=1}^{s_t}\frac{1}{\gamma}
\Big(
\frac{\partial h_{t,i}}{\partial \boldsymbol{x}_t}\boldsymbol{x}+\frac{\partial h_{t,i}}{\partial \boldsymbol{u}}\boldsymbol{u}
\Big)^2, \label{pf_prodt}
\end{align}
and 
\begin{align}
\boldsymbol{x}^\prime \hat L_T^{xx}\boldsymbol{x}& =
\boldsymbol{x}^\prime L_T^{xx} \boldsymbol{x}
+\small
\boldsymbol{x}^\prime
\Big(\sum_{i=1}^{q_T}\frac{\gamma}{g_{t,i}^2}\frac{\partial g_{T,i}^\prime}{\partial \boldsymbol{x}_T}\frac{\partial g_{T,i}}{\partial \boldsymbol{x}_T}+\sum_{i=1}^{s_T}\frac{1}{\gamma}\frac{\partial h_{T,i}^\prime}{\partial \boldsymbol{x}_T}\frac{\partial h_{T,i}}{\partial \boldsymbol{x}_T}\Big)
\boldsymbol{x}\nonumber\\
&=
\boldsymbol{x}^\prime L_T^{xx} \boldsymbol{x}+\sum_{i=1}^{q_T}\frac{\gamma}{g_{T,i}^2}\Big(
\frac{\partial g_{T,i}}{\partial \boldsymbol{x}_T}\boldsymbol{x}
\Big)^2
+\small\sum_{i=1}^{s_T}\frac{1}{\gamma}
\Big(
\frac{\partial h_{T,i}}{\partial \boldsymbol{x}_T}\boldsymbol{x}
\Big)^2. \label{pf_prodT}
\end{align}

For the second-order condition of the unconstrained optimal control system ${\boldsymbol{\Sigma}}(\boldsymbol{\theta},\gamma)$ with any $(\boldsymbol{\theta},\gamma)$ near ($\boldsymbol{\bar{\theta}},0$) with $\gamma>0$,  we need to prove  that
\begin{equation}\label{pfhessian}
\sum_{t=0}^{T-1}
\begin{bmatrix}
\boldsymbol{x}_t\\
\boldsymbol{u}_t
\end{bmatrix}^\prime
\begin{bmatrix}
\hat L_t^{xx}{(\boldsymbol{\theta}, \gamma)}& \hat L_t^{xu}{(\boldsymbol{\theta}, \gamma)}\\
\hat L_t^{ux}{(\boldsymbol{\theta}, \gamma)}& \hat L_t^{uu}{(\boldsymbol{\theta}, \gamma)}
\end{bmatrix}
\begin{bmatrix}
\boldsymbol{x}_t\\
\boldsymbol{u}_t
\end{bmatrix}+
\boldsymbol{x}_T^\prime \hat L_T^{xx}{(\boldsymbol{\theta}, \gamma)}  \boldsymbol{x}_T>0,
\end{equation} for any $\{\boldsymbol{x}_{0:T}, \boldsymbol{u}_{0:T-1}\}\neq\boldsymbol{0}$ satisfying
\begin{equation}\label{pflineardyn}
\boldsymbol{x}_{t+1}=F_t^x{(\boldsymbol{\theta}, \gamma)}\boldsymbol{x}_t+F_t^u{(\boldsymbol{\theta}, \gamma)}\boldsymbol{u}_t 
\quad \text{and}\quad
\boldsymbol{x}_{0}=\boldsymbol{0}.
\end{equation}
Here, for  convenience, the dependence in  $F_t^x({\boldsymbol{\theta},\gamma})$, $F_t^u({\boldsymbol{\theta},\gamma})$, $H_t^{xx}({\boldsymbol{\theta},\gamma})$, $H_t^{xu}({\boldsymbol{\theta},\gamma})$, $H_t^{uu}({\boldsymbol{\theta},\gamma})$, and $H_T^{xx}({\boldsymbol{\theta},\gamma})$ means that these first- and second-order derivatives are evaluated at trajectory $\boldsymbol{\xi}_{({\boldsymbol{\theta},\gamma})}$ (the same notation convention applies below).

\medskip
Proof by contradiction: suppose that the above second-order condition in (\ref{pfhessian})-(\ref{pflineardyn}) is false. Then,  there must exist a sequence of parameters  $(\boldsymbol{\theta}^k,\gamma^k)$ with $\gamma^k>0$ and a sequence of trajectories $\{\boldsymbol{x}_{0:T}^k, \boldsymbol{u}_{0:T-1}^k\}\neq\boldsymbol{0}$  such that $(\boldsymbol{\theta}^k,\gamma^k)\rightarrow (\boldsymbol{\bar{\theta}},0)$, $\boldsymbol{x}_{t+1}^k=F_t^x({\boldsymbol{\theta}^k,\gamma^k})\boldsymbol{x}_t^k+F_t^u({\boldsymbol{\theta}^k,\gamma^k})\boldsymbol{u}_t^k$ with $
\boldsymbol{x}_{0}^k=\boldsymbol{0}$, and 
\begin{equation}\label{pfhessian_k}
\sum_{t=0}^{T-1}
\begin{bmatrix}
\boldsymbol{x}_t^k\\
\boldsymbol{u}_t^k
\end{bmatrix}^\prime
\begin{bmatrix}
\hat L_t^{xx}({\boldsymbol{\theta}^k,\gamma^k})& \hat L_t^{xu}({\boldsymbol{\theta}^k,\gamma^k})\\
\hat L_t^{ux}({\boldsymbol{\theta}^k,\gamma^k})& \hat L_t^{uu}({\boldsymbol{\theta}^k,\gamma^k})
\end{bmatrix}
\begin{bmatrix}
\boldsymbol{x}_t^k\\
\boldsymbol{u}_t^k
\end{bmatrix}+
{\boldsymbol{x}_T^k}^\prime \hat L_T^{xx}({\boldsymbol{\theta}^k,\gamma^k})  \boldsymbol{x}_T^k\leq 0,
\end{equation}
for $k=1,2,3,\cdots$. Here, the dependence $({\boldsymbol{\theta}^k,\gamma^k})$ means that  these first- and second-order derivatives  are  evaluated at trajectory $\boldsymbol{\xi}_{({\boldsymbol{\theta}^k,\gamma^k})}$ for notation convenience. Without loss of generality, assume $\norm{\col\{\boldsymbol{x}^k_{0:T}, \boldsymbol{u}^k_{0:T-1}\}}=1$ for all $k$. Select a convergent sub-sequence  $\{\boldsymbol{x}^k_{0:T}, \boldsymbol{u}^k_{0:T-1}\}$, relabel the sequence $\{\boldsymbol{x}^k_{0:T}, \boldsymbol{u}^k_{0:T-1}\}$ for  convenience, and call its limit $\{\boldsymbol{x}^*_{0:T}, \boldsymbol{u}^*_{0:T-1}\}$, that is,  $\{\boldsymbol{x}^k_{0:T}, \boldsymbol{u}^k_{0:T-1}\}\rightarrow\{\boldsymbol{x}^*_{0:T}, \boldsymbol{u}^*_{0:T-1}\}$ and $(\boldsymbol{\theta}^k,\gamma^k)\rightarrow (\boldsymbol{\bar{\theta}},0)$ as $k\rightarrow +\infty$ and 
$
\boldsymbol{x}_{t+1}^*=F_t^x(\boldsymbol{\bar{\theta}},0)\boldsymbol{x}_{t}^*+F_t^u(\boldsymbol{\bar{\theta}},0)\boldsymbol{u}_{t}^*$ with
$ \boldsymbol{x}_0^*=\boldsymbol{0}.$ Then, the limit $\{\boldsymbol{x}^*_{0:T}, \boldsymbol{u}^*_{0:T-1}\}
$ must fall into either of two cases discussed below.

\smallskip

\emph{Case 1:} $\norm{\col{\{\boldsymbol{x}^*_{0:T}, \boldsymbol{u}^*_{0:T-1}\}}}=1$ and at least one of the following holds:
\begin{equation}\label{pf_case1conditions}
\begin{aligned}
&\bar{{G}}_t^{{x}}(\boldsymbol{\bar{\theta}},0)\boldsymbol{x}^*_t+\bar{{G}}_t^{{u}}(\boldsymbol{\bar{\theta}},0)\boldsymbol{u}^*_t\neq\boldsymbol{0} \quad \exists t\quad \text{or}\quad {{H}}_t^{{x}}(\boldsymbol{\bar{\theta}},0)\boldsymbol{x}^*_t+{{H}}_t^{{u}}(\boldsymbol{\bar{\theta}},0)\boldsymbol{u}^*_t\neq\boldsymbol{0}\quad \exists t\\ &\qquad\qquad\qquad\text{or}\quad\bar{{G}}_T^{{x}}(\boldsymbol{\bar{\theta}},0)\boldsymbol{x}^*_T\neq\boldsymbol{0}
\quad\text{or}\quad
H_T^{{x}}(\boldsymbol{\bar{\theta}},0)\boldsymbol{x}^*_T\neq\boldsymbol{0}.
\end{aligned} 
\end{equation}
In this case,  as $k\rightarrow 0$,  $\{\boldsymbol{x}^k_{0:T}, \boldsymbol{u}^k_{0:T-1}\}\rightarrow\{\boldsymbol{x}^*_{0:T}, \boldsymbol{u}^*_{0:T-1}\}$, $(\boldsymbol{\theta}^k,\gamma^k)\rightarrow (\boldsymbol{\bar{\theta}},0)$, we will have
\begin{multline}\label{pf_limits}
\sum_{t=0}^{T-1}\Bigg(
\sum\limits_{i=1}^{q_t}\frac{\gamma^k}{(g_{t,i}^k)^2}\Big(
\frac{\partial g^k_{t,i}}{\partial \boldsymbol{x}_t}\boldsymbol{x}_t^k+\frac{\partial g^k_{t,i}}{\partial \boldsymbol{u}_t^k}\boldsymbol{u}_t^k
\Big)^2
+\sum\limits_{i=1}^{s_t}\frac{1}{\gamma^k}
\Big(
\frac{\partial h_{t,i}^k}{\partial \boldsymbol{x}_t^k}\boldsymbol{x}_t^k+\frac{\partial h_{t,i}^k}{\partial \boldsymbol{u}_t^k}\boldsymbol{u}_t^k
\Big)^2
\Bigg)\\+
\sum_{i=1}^{q_T}\frac{\gamma^k}{(g_{T,i}^k)^2}\Big(
\frac{\partial g_{T,i}^k}{\partial \boldsymbol{x}_T}\boldsymbol{x}_T^k
\Big)^2
+\sum_{i=1}^{s_T}\frac{1}{\gamma^k}
\Big(
\frac{\partial h_{T,i}^k}{\partial \boldsymbol{x}_T}\boldsymbol{x}_T^k
\Big)^2\rightarrow +\infty,
\end{multline}
where  $\frac{\partial g_{t,i}^k}{\partial \boldsymbol{x}_t}$,  $\frac{\partial g_{T,i}^k}{\partial \boldsymbol{x}_T}$, $g_{t,i}^k$, $g_{T,i}^k$, $\frac{\partial h_{t,i}^k}{\partial \boldsymbol{x}_t}$,  $\frac{\partial h_{T,i}^k}{\partial \boldsymbol{x}_T}$ are  with superscript $k$ to denote their values are evaluated at $\boldsymbol{\xi}_{(\boldsymbol{\theta}^k,\gamma^k)}$ for notation  convenience.  (\ref{pf_limits})  is because at least one of the terms in the summation is $ +\infty$. Here, we have used the following facts from the last two equations in (\ref{CPMP-modify}):
\begin{equation*}
\frac{\gamma}{\big({g}_{t,i}(\boldsymbol{x}_{t}^{(\boldsymbol{\theta},\gamma)},\boldsymbol{u}_{t}^{(\boldsymbol{\theta},\gamma)},\boldsymbol{\theta})\big)^2}=-\frac{v^{(\boldsymbol{\theta},\gamma)}_{t,i}}{{{g}_{t,i}(\boldsymbol{x}_{t}^{(\boldsymbol{\theta},\gamma)},\boldsymbol{u}_{t}^{(\boldsymbol{\theta},\gamma)},\boldsymbol{\theta})}}\rightarrow 0 \quad \text{or} \quad \rightarrow+\infty \quad \text{as} \quad (\boldsymbol{\theta},\gamma)\rightarrow (\boldsymbol{\bar{\theta}},0),
\end{equation*}
where $\rightarrow0$ corresponds to the inactive inequalities   ${{g}_{t,i}(\boldsymbol{x}_{t}^{\boldsymbol{\bar\theta}},\boldsymbol{u}_{t}^{\boldsymbol{\bar\theta}},\boldsymbol{\bar\theta})}<0$ and $\rightarrow +\infty$ corresponds to the active inequalities   ${{g}_{t,i}(\boldsymbol{x}_{t}^{\boldsymbol{\bar\theta}},\boldsymbol{u}_{t}^{\boldsymbol{\bar\theta}},\boldsymbol{\bar\theta})}=0$  (${v}_{t,i}^{\boldsymbol{\theta}}>0$ due to strict complementarity); and also
\begin{equation*}
\frac{\gamma}{\big({g}_{T,i}(\boldsymbol{x}_{T}^{(\boldsymbol{\theta},\gamma)},\boldsymbol{\theta})\big)^2}=-\frac{v^{(\boldsymbol{\theta},\gamma)}_{T,i}}{{{g}_{T,i}(\boldsymbol{x}_{T}^{(\boldsymbol{\theta},\gamma)},\boldsymbol{\theta})}}\rightarrow 0 \quad \text{or} \quad \rightarrow+\infty \quad \text{as} \quad (\boldsymbol{\theta}, \gamma)\rightarrow (\boldsymbol{\bar\theta}, 0).
\end{equation*}
where $\rightarrow0$ corresponds to the inactive inequalities  ${{g}_{T,i}(\boldsymbol{x}_{T}^{\boldsymbol{\bar\theta}},\boldsymbol{\bar\theta})}<0$ and $\rightarrow +\infty$ corresponds to the active inequalities ${{g}_{T,i}(\boldsymbol{x}_{T}^{\boldsymbol{\bar\theta}},\boldsymbol{\bar\theta})}=0$ (${v}_{T,i}^{\boldsymbol{\theta}}>0$ due to strict complementarity).

By extending the left side of (\ref{pfhessian_k}) based on the facts  (\ref{pf_prodt}) and (\ref{pf_prodT}), (\ref{pf_limits}) immediately leads to
\begin{equation}\label{pfcase1}
\lim\limits_{k\rightarrow +\infty}\Bigg(\sum_{t=0}^{T-1}
\begin{bmatrix}
\boldsymbol{x}_t^k\\
\boldsymbol{u}_t^k
\end{bmatrix}^\prime
\begin{bmatrix}
\hat L_t^{xx}({\boldsymbol{\theta}^k,\gamma^k})& \hat L_t^{xu}({\boldsymbol{\theta}^k,\gamma^k})\\
\hat L_t^{ux}({\boldsymbol{\theta}^k,\gamma^k})& \hat L_t^{uu}({\boldsymbol{\theta}^k,\gamma^k})
\end{bmatrix}
\begin{bmatrix}
\boldsymbol{x}_t^k\\
\boldsymbol{u}_t^k
\end{bmatrix}+
{\boldsymbol{x}_T^k}^\prime \hat L_T^{xx}({\boldsymbol{\theta}^k,\gamma^k})  \boldsymbol{x}_T^k\Bigg) \rightarrow +\infty,
\end{equation}
which obviously contradicts  (\ref{pfhessian_k}).

\smallskip

\emph{ Case 2:} $\norm{\col{\{\boldsymbol{x}^*_{0:T}, \boldsymbol{u}^*_{0:T-1}\}}}=1$ and all of the following holds: 
\begin{equation}
\begin{aligned}
&\bar{{G}}_t^{{x}}(\boldsymbol{\bar{\theta}},0)\boldsymbol{x}^*_t+\bar{{G}}_t^{{u}}(\boldsymbol{\bar{\theta}},0)\boldsymbol{u}^*_t=\boldsymbol{0} \quad \forall t\quad \text{and}\quad {{H}}_t^{{x}}(\boldsymbol{\bar{\theta}},0)\boldsymbol{x}^*_t+{{H}}_t^{{u}}(\boldsymbol{\bar{\theta}},0)\boldsymbol{u}^*_t=\boldsymbol{0}\quad \forall t\\ &\qquad\qquad\qquad\text{and}\quad\bar{{G}}_T^{{x}}(\boldsymbol{\bar{\theta}},0)\boldsymbol{x}^*_T=\boldsymbol{0}
\quad\text{and}\quad
H_T^{{x}}(\boldsymbol{\bar{\theta}},0)\boldsymbol{x}^*_T=\boldsymbol{0}.
\end{aligned} 
\end{equation}

In this case, we have
\begin{multline}\label{pfcase2}
\lim\limits_{k\rightarrow +\infty}\Bigg(\sum_{t=0}^{T-1}
\begin{bmatrix}
\boldsymbol{x}_t^k\\
\boldsymbol{u}_t^k
\end{bmatrix}^\prime
\begin{bmatrix}
\hat L_t^{xx}({\boldsymbol{\theta}^k,\gamma^k})& \hat L_t^{xu}({\boldsymbol{\theta}^k,\gamma^k})\\
\hat L_t^{ux}({\boldsymbol{\theta}^k,\gamma^k})& \hat L_t^{uu}({\boldsymbol{\theta}^k,\gamma^k})
\end{bmatrix}
\begin{bmatrix}
\boldsymbol{x}_t^k\\
\boldsymbol{u}_t^k
\end{bmatrix}+
{\boldsymbol{x}_T^k}^\prime \hat L_T^{xx}({\boldsymbol{\theta}^k,\gamma^k})  \boldsymbol{x}_T^k\Bigg)
\\
\geq 
\lim\limits_{k\rightarrow +\infty}\Bigg(\sum_{t=0}^{T-1}
\begin{bmatrix}
\boldsymbol{x}_t^k\\
\boldsymbol{u}_t^k
\end{bmatrix}^\prime
\begin{bmatrix}
L_t^{xx}({\boldsymbol{\theta}^k,\gamma^k})& L_t^{xu}({\boldsymbol{\theta}^k,\gamma^k})\\
L_t^{ux}({\boldsymbol{\theta}^k,\gamma^k})& \hat L_t^{uu}({\boldsymbol{\theta}^k,\gamma^k})
\end{bmatrix}
\begin{bmatrix}
\boldsymbol{x}_t^k\\
\boldsymbol{u}_t^k
\end{bmatrix}+
{\boldsymbol{x}_T^k}^\prime  L_T^{xx}({\boldsymbol{\theta}^k,\gamma^k})  \boldsymbol{x}_T^k\Bigg)
\\
=\sum_{t=0}^{T-1}
\begin{bmatrix}
\boldsymbol{x}_t^*\\
\boldsymbol{u}_t^*
\end{bmatrix}^\prime
\begin{bmatrix}
L_t^{xx}({\boldsymbol{\bar\theta},0})& L_t^{xu}({\boldsymbol{\bar\theta},0})\\
L_t^{ux}({\boldsymbol{\bar\theta},0})& \hat L_t^{uu}({\boldsymbol{\bar\theta},0})
\end{bmatrix}
\begin{bmatrix}
\boldsymbol{x}_t^*\\
\boldsymbol{u}_t^*
\end{bmatrix}+
{\boldsymbol{x}_T^*}^\prime  L_T^{xx}({\boldsymbol{\bar\theta},0})  \boldsymbol{x}_T^* \,>0.
\end{multline}
Here, the first inequality is based on the fact that the residual term is always non-negative, i.e., 
\begin{multline}
\sum_{t=0}^{T-1}\Bigg(
\sum\limits_{i=1}^{q_t}\frac{\gamma^k}{(g_{t,i}^k)^2}\Big(
\frac{\partial g^k_{t,i}}{\partial \boldsymbol{x}_t}\boldsymbol{x}_t^k+\frac{\partial g^k_{t,i}}{\partial \boldsymbol{u}_t^k}\boldsymbol{u}_t^k
\Big)^2
+\sum\limits_{i=1}^{s_t}\frac{1}{\gamma^k}
\Big(
\frac{\partial h_{t,i}^k}{\partial \boldsymbol{x}_t^k}\boldsymbol{x}_t^k+\frac{\partial h_{t,i}^k}{\partial \boldsymbol{u}_t^k}\boldsymbol{u}_t^k
\Big)^2
\Bigg)\\+
\sum_{i=1}^{q_T}\frac{\gamma^k}{(g_{T,i}^k)^2}\Big(
\frac{\partial g_{T,i}^k}{\partial \boldsymbol{x}_T}\boldsymbol{x}_T^k
\Big)^2
+\sum_{i=1}^{s_T}\frac{1}{\gamma^k}
\Big(
\frac{\partial h_{T,i}^k}{\partial \boldsymbol{x}_T}\boldsymbol{x}_T^k
\Big)^2\geq 0,
\end{multline}
the last inequality is directly from the second-order condition in (\ref{secordercon.1})-(\ref{bindingequation})  in Lemma \ref{Lemma1}. Obviously, (\ref{pfcase2}) also contracts    (\ref{pfhessian_k}).

\medskip
Combining the above two cases, we can conclude  that  
for any $(\boldsymbol{\theta},\gamma)$ near ($\boldsymbol{\bar{\theta}},0$) with $\gamma>0$, the trajectory $\boldsymbol{\xi}_{(\boldsymbol{\theta},\gamma)}$ to  the unconstrained optimal control system ${\boldsymbol{\Sigma}}(\boldsymbol{\theta},\gamma)$ satisfies both its PMP condition in (\ref{hamil-penalty-oc}) and the second-order condition  in (\ref{pfhessian})-(\ref{pflineardyn}). Thus, one can assert that  $\boldsymbol{\xi}_{(\boldsymbol{\theta},\gamma)}$ is a local isolated  minimizing trajectory to ${\boldsymbol{\Sigma}}(\boldsymbol{\theta},\gamma)$.
This completes the proof of Claim (a) in Theorem \ref{theorem2}.

\subsection{Proof of Claim  (b)}
Given that  the conditions (i)-(iii) in Lemma \ref{result1} hold for $\boldsymbol{\Sigma}(\boldsymbol{\bar\theta})$, we have the following conclusions:

(1) From Claim (a) and its proof, we know that for any $(\boldsymbol{\theta},{\gamma})$ in the neighborhood of $(\boldsymbol{\bar\theta},{0})$, there exists a unique once-continuously differentiable function $\left(\boldsymbol{\xi}_{(\boldsymbol{\theta},\gamma)}, \boldsymbol{\lambda}_{0:T}^{(\boldsymbol{\theta},\gamma)}, \boldsymbol{v}_{0:T}^{(\boldsymbol{\theta},\gamma)},\boldsymbol{w}_{0:T}^{(\boldsymbol{\theta},\gamma)}\right)$, which satisfies (\ref{CPMP-modify}). Additionally provided $\gamma>0$, such   $\boldsymbol{\xi}_{(\boldsymbol{\theta},\gamma)}$ is also a local isolated minimizing trajectory for the well-defined unconstrained optimal control system ${\boldsymbol{\Sigma}}(\boldsymbol{\theta},\gamma)$.

(2) Additionally let  $\gamma=0$ in (\ref{CPMP-modify}), and   (\ref{CPMP-modify}) becomes the C-PMP condition for the constrained optimal control system $\boldsymbol{\Sigma}(\boldsymbol{\theta})$. From Lemma \ref{result1},  for any  $\boldsymbol{\theta}$ near $\boldsymbol{\bar\theta}$,
$\boldsymbol{\xi}_{\boldsymbol{\theta}}=\boldsymbol{\xi}_{(\boldsymbol{\theta},\gamma=0)}$ is a differentiable  local isolated minimizing trajectory for  $\boldsymbol{\Sigma}(\boldsymbol{\theta})$, associated with the unique once-continuously differentiable  function $\left( 
\boldsymbol{\lambda}_{1:T}^{(\boldsymbol{\theta},\gamma=0)}, \boldsymbol{v}_{0:T}^{(\boldsymbol{\theta},\gamma=0)},
\boldsymbol{w}_{0:T}^{(\boldsymbol{\theta},\gamma=0)}
\right)$.

Therefore,  due to the uniqueness  and once-continuous differentiability of $\boldsymbol{\xi}_{(\boldsymbol{\theta},\gamma)}$ with respect to  $(\boldsymbol{\theta},{\gamma})$ near $(\boldsymbol{\bar\theta},{0})$, one can obtain
\begin{equation}\label{pfresults1}
\boldsymbol{\xi}_{(\boldsymbol{\theta}, \gamma)}\rightarrow \boldsymbol{\xi}_{(\boldsymbol{\theta}, 0)}=\boldsymbol{\xi}_{\boldsymbol{\theta}}\quad \text{as}\quad \gamma\rightarrow 0,
\end{equation}
and
\begin{equation}
\frac{\partial \boldsymbol{\xi}_{(\boldsymbol{\theta}, \gamma)}}{\partial \boldsymbol{\theta}}\rightarrow \frac{\partial \boldsymbol{\xi}_{(\boldsymbol{\theta}, 0)}}{\partial \boldsymbol{\theta} }=\frac{\boldsymbol{\xi}_{\boldsymbol{\theta}}}{\partial \boldsymbol{\theta}}\quad \text{as}\quad \gamma\rightarrow 0.
\end{equation}

Here (\ref{pfresults1}) is due to that $\boldsymbol{\xi}_{(\boldsymbol{\theta}, \gamma)}$ is unique and  continuous at ($\boldsymbol{\theta},\gamma=0$), and (\ref{pfresults1}) is because $\boldsymbol{\xi}_{(\boldsymbol{\theta},\gamma)}$ is unique and once-continuously differentiable at ($\boldsymbol{\theta},\gamma=0$). This completes the proof of Claim (b) in Theorem \ref{theorem2}.

\subsection{Proof of Claim  (c)}

For the unconstrained optimal control system  ${\boldsymbol{\Sigma}}(\boldsymbol{\theta},\gamma)$ with any $(\boldsymbol{\theta},\gamma)$ near ($\boldsymbol{\bar{\theta}},0$), $\gamma>0$, in order to show that its trajectory derivative $\frac{\partial \boldsymbol{\xi}_{(\boldsymbol{\theta}, \gamma)}}{\partial \boldsymbol{\theta}}$ is  a globally unique minimizing trajectory to  its corresponding auxiliary control system ${\boldsymbol{\overline\Sigma}}(\boldsymbol{\xi}_{(\boldsymbol{\theta},\gamma)})$, similarly to the claim of Theorem \ref{theorem1}, we need to verify if the following three conditions hold for   ${\boldsymbol{\Sigma}}(\boldsymbol{\theta},\gamma)$ at $(\boldsymbol{\theta}, \gamma)$.

\begin{itemize}
	\item[(i)] The second-order condition holds for  $\boldsymbol{\xi}_{(\boldsymbol{\theta}, \gamma)}$ to be a local isolated minimizing trajectory for ${\boldsymbol{\Sigma}}(\boldsymbol{\theta},\gamma)$. In fact, this has been proved in the proof of Claim (a).
	\item [(ii)]  The gradients of all binding constraints (i.e., all equality and active inequality constraints) are linearly independent at $\boldsymbol{\xi}_{(\boldsymbol{\theta}, \gamma)}$.  Since we do not have inequality constraints in ${\boldsymbol{\Sigma}}(\boldsymbol{\theta},\gamma)$, we only need to show the  gradients of the dynamics constraint  
	are linearly independent at $\boldsymbol{\xi}_{(\boldsymbol{\theta}, \gamma)}$. Specifically, we need to show that the following linear equations are independent
	\begin{equation}\label{pf_linearequation}
	\boldsymbol{x}_{t+1}=F_t^x{(\boldsymbol{\theta}, \gamma)}\boldsymbol{x}_t+F_t^u{(\boldsymbol{\theta}, \gamma)}\boldsymbol{u}_t,  \quad \text{and}\quad
	\boldsymbol{x}_{0}=\boldsymbol{0}, \quad t=0,1,\cdots, T.
	\end{equation}
	where the dependence $(\boldsymbol{\theta}, \gamma)$   means that the derivative  matrices are  evaluated at trajectory $\boldsymbol{\xi}_{({\boldsymbol{\theta},\gamma})}$,  $\boldsymbol{x}_{0:T}$ and $\boldsymbol{u}_{0:T-1}$ here are  variables. In fact, the above linear equations in (\ref{pf_linearequation}) can be equivalently written as
	\begin{equation}
	\boldsymbol{F}_x\boldsymbol{x}_{1:T} +\boldsymbol{F}_u\boldsymbol{u}_{0:T-1}=\boldsymbol{0},
	\end{equation}
	with
	\begin{equation}
	\boldsymbol{F}_u=\begin{bmatrix}
	-F_0^u{(\boldsymbol{\theta}, \gamma)} & 0 &   \cdots  &   0\\
	0 & -F_1^u{(\boldsymbol{\theta}, \gamma)} & \cdots & 0\\
	\vdots& \vdots& \ddots & \vdots\\
	0 &0 & \cdots & -F_{T-1}^x{(\boldsymbol{\theta}, \gamma)} \\
	\end{bmatrix},
	\end{equation}
	and
	\begin{equation}
	\boldsymbol{F}_x=\begin{bmatrix}
	I & 0 &   \cdots & 0  &   0\\
	-F_1^x{(\boldsymbol{\theta}, \gamma)} & I & \cdots& 0 & 0\\
	\vdots& \vdots& \ddots & \vdots& \vdots\\
	0 &0&\cdots & I  & 0\\
	0 &0 & \cdots &-F_{T-1}^x{(\boldsymbol{\theta}, \gamma)} & I\\
	\end{bmatrix}.
	\end{equation}
	Obviously, all rows in the concatenation matrix $[\boldsymbol{F}_u, \boldsymbol{F}_x]$ are linear-independent because $[\boldsymbol{F}_u, \boldsymbol{F}_x]$ is already in its the reduced echelon form and has full row rank. Thus, one can conclude that the linear equations in (\ref{pf_linearequation}) are linearly independent.
	
	\item[(iii)]  Strict complementarity does not apply because there are no inequality constraints in   ${\boldsymbol{\Sigma}}(\boldsymbol{\theta},\gamma)$ at $(\boldsymbol{\theta}, \gamma)$.

\end{itemize}

With the above three conditions satisfied, by  applying Theorem \ref{theorem1}, we can conclude that $\frac{\partial \boldsymbol{\xi}_{(\boldsymbol{\theta},\gamma)}}{\partial \boldsymbol{\theta}}$ is a globally unique minimizing trajectory to  the auxiliary control system ${\boldsymbol{\overline\Sigma}}(\boldsymbol{\xi}_{(\boldsymbol{\theta},\gamma)})$. This completes the Claim (c) in Theorem \ref{theorem2}.

\bigskip

With the Claims (a),  (b), and (c) proved, we have completed the proof of  Theorem \ref{theorem2}.\qed

\bigskip

\section{Proof of Theorem \ref{theorem3}}\label{pf_theorem3}
We know  from the proof of Claim (a) of Theorem \ref{theorem2} in Appendix \ref{pf_theorem2_(a)} that given  the conditions in Theorem \ref{theorem3}, 
\begin{itemize}
	\item for any $(\boldsymbol{\theta},{\gamma})$ in the neighborhood of $(\boldsymbol{\theta}^*,{0})$, there exists a unique once-continuously differentiable function $\left(\boldsymbol{\xi}_{(\boldsymbol{\theta},\gamma)}, \boldsymbol{\lambda}_{0:T}^{(\boldsymbol{\theta},\gamma)}, \boldsymbol{v}_{0:T}^{(\boldsymbol{\theta},\gamma)},\boldsymbol{w}_{0:T}^{(\boldsymbol{\theta},\gamma)}\right)$, which satisfies (\ref{CPMP-modify}), 
	and
	\begin{equation}
	\left(
	\boldsymbol{\xi}_{(\boldsymbol{\theta},\gamma)},
	\boldsymbol{\lambda}_{0:T}^{({\boldsymbol{\theta}},\gamma)}, \boldsymbol{v}_{0:T}^{({\boldsymbol{\theta}},\gamma)}, \boldsymbol{w}_{0:T}^{({\boldsymbol{\theta}},\gamma)}
	\right)=\left(\boldsymbol{\xi}_{{\boldsymbol{{\theta}^*}}},\boldsymbol{\lambda}_{1:T}^{{\boldsymbol{{\theta}^*}}},\boldsymbol{v}_{0:T}^{{\boldsymbol{{\theta}^*}}},\boldsymbol{w}_{0:T}^{{\boldsymbol{{\theta^*}}}}\right) \,\, \text{when} \,\,  (\boldsymbol{\theta},{\gamma})=(\boldsymbol{\theta}^*,{0});
	\nonumber
	\end{equation}
	
	\item additionally, if all functions defining $\boldsymbol{\Sigma}(\boldsymbol{\theta})$ are \emph{three-times} continuously differentiable, it immediately follows that  $\boldsymbol{\xi}_{(\boldsymbol{\theta},\gamma)}$ is then \emph{twice} continuously differentiable near $(\boldsymbol{\theta}^*,0)$. 
	This is a direct result by applying the $C^k$ implicit function theorem \cite{dieudonne2011foundations}, to the C-PMP condition  (\ref{CPMP-modify}) in the neighborhood of $(\boldsymbol{\theta}^*,0)$.
	
	\item  additionally provided $\gamma>0$, such   $\boldsymbol{\xi}_{(\boldsymbol{\theta},\gamma)}$ is also a local isolated minimizing trajectory for the well-defined unconstrained optimal control system ${\boldsymbol{\Sigma}}(\boldsymbol{\theta},\gamma)$ in Problem \ref{pdpapprx_oc}.
\end{itemize}

 Thus, in the following, we will ignore the computation process for obtaining $\boldsymbol{\xi}_{(\boldsymbol{\theta},\gamma)}$ and simply   view that  $\boldsymbol{\xi}_{(\boldsymbol{\theta},\gamma)}$is the twice continuously differentiable function of $(\boldsymbol{\theta},\gamma)$ near $(\boldsymbol{\theta}^*,0)$ and $\boldsymbol{\xi}_{\boldsymbol{\theta}^*}=\boldsymbol{\xi}_{(\boldsymbol{\theta}=\boldsymbol{\theta}^*,\gamma=0)}$. 
The following proof of  Theorem \ref{theorem3} follows the procedure of  the general interior-point minimization methods, which are systematically studied in \cite{fiacco1990nonlinear} (see Theorem 14, p. 80).

Recall the optimization in Problem \ref{equ_problem_penalty_approx}, re-write it below for easy reference,
\begin{equation}\label{appendix-unconstrainedopt}
\boldsymbol{\theta}^*(\epsilon, \gamma)= \arg\min_{\boldsymbol{{\theta}}}\quad W\big(\boldsymbol{\theta},\epsilon,\gamma\big)
\end{equation} 
with
\begin{equation}
W\big(\boldsymbol{\theta},\epsilon,\gamma\big)=\ell\big(\boldsymbol{\xi}_{(\boldsymbol{\theta}, \gamma)},\boldsymbol{\theta}\big)-\epsilon\sum_{i=1}^{l}\ln\Big({-}R_i\big(\boldsymbol{\xi}_{(\boldsymbol{\theta}, \gamma)},\boldsymbol{\theta}\big)\Big)
\end{equation}

Given in Theorem \ref{theorem3} that $\boldsymbol{\theta}^*$ satisfies the second-order sufficient condition for a local isolated minimizer to Problem \ref{equ_problem} (recall the general second-order sufficient condition in Lemma \ref{lemma_generalnlp}), one can say that there  exists a multiplier $\boldsymbol{u}^*\in\mathbb{R}^{l}$ such that
\begin{equation}\label{appendix-kkt}
\begin{aligned}
&\nabla L(\boldsymbol{\theta}^*,\boldsymbol{u}^*)=\nabla\ell(\boldsymbol{\xi}_{\boldsymbol{\theta}^*},\boldsymbol{{\theta}}^*)+\sum\nolimits_{i=1}^{l}u_i^*\nabla R_i{(\boldsymbol{\xi}_{\boldsymbol{\theta}^*},\boldsymbol{{\theta}}^*)}=\boldsymbol{0},\\
&u_i^* R_i{(\boldsymbol{\xi}_{\boldsymbol{\theta}^*},\boldsymbol{{\theta}}^*)}=0, \quad i=1,2,...,l,\\
&R_i{(\boldsymbol{\xi}_{\boldsymbol{\theta}^*},\boldsymbol{{\theta}}^*)}\leq 0,\quad u_i^*\geq 0,\quad i=1,2,...,l,
\end{aligned}
\end{equation} 
with the Lagrangian defined as
\begin{equation}\label{Lagproblemp}
L(\boldsymbol{\theta},\boldsymbol{u})=\ell(\boldsymbol{\xi}_{\boldsymbol{\theta}},\boldsymbol{{\theta}})+\sum\nolimits_{i=1}^{l}u_i R_i{(\boldsymbol{\xi}_{\boldsymbol{\theta}},\boldsymbol{{\theta}})},
\end{equation}
and further for any $\boldsymbol{\theta}\neq\boldsymbol{0}$ satisfying $\boldsymbol{\theta}^\prime \nabla R_i{(\boldsymbol{\xi}_{\boldsymbol{\theta}^*},\boldsymbol{{\theta}}^*)}=0$  with ${u}_i^*>0$ and $\boldsymbol{\theta}^\prime \nabla R_i{(\boldsymbol{\xi}_{\boldsymbol{\theta}^*},\boldsymbol{{\theta}}^*)}\leq 0$  with ${u}_i^*\geq 0$, it follows  
\begin{equation}\label{appendix-second}
\boldsymbol{\theta}^\prime \nabla^2L(\boldsymbol{\theta}^*, \boldsymbol{u}^*)\boldsymbol{\theta}>0.
\end{equation} 
Here,  $\nabla L$ and $\nabla^2 L$ denote the first- and second-order derivatives of $L$ with respect to $\boldsymbol{\theta}$, respectively; and $\boldsymbol{\xi}_{\boldsymbol{\theta}^*}=\boldsymbol{\xi}_{(\boldsymbol{\theta}=\boldsymbol{\theta}^*,\gamma=0)}$.

\subsection{Proof of Claim (a)} \label{pf_theorem3.a}

We modify the first two equations in (\ref{appendix-kkt}) into 
\begin{equation}\label{appendix-modified-kkt}
\begin{aligned}
&\nabla\ell\Big(\boldsymbol{\xi}_{(\boldsymbol{\theta}^*{(\epsilon,\gamma),\gamma})},\boldsymbol{{\theta}}^*(\epsilon,\gamma)\Big)+\sum\nolimits_{i=1}^{l}u_i^*(\epsilon,\gamma)\nabla R_i{\Big(\boldsymbol{\xi}_{(\boldsymbol{\theta}^*{(\epsilon,\gamma),\gamma})},\boldsymbol{{\theta}}^*(\epsilon,\gamma)\Big)}=\boldsymbol{0},\\
&u_i^*(\epsilon,\gamma) R_i{\Big(\boldsymbol{\xi}_{(\boldsymbol{\theta}^*{(\epsilon,\gamma),\gamma})},\boldsymbol{{\theta}}^*(\epsilon,\gamma)\Big)}=-\epsilon, \quad i=1,2,...,l,
\end{aligned}
\end{equation}
respectively, and consider both $\boldsymbol{\theta}^*(\epsilon,\gamma)$ and $ \boldsymbol{u}^*(\epsilon,\gamma)$ are   implicitly determined by $\epsilon$ and $\gamma$ through the above equations.

Look at (\ref{appendix-modified-kkt}) and note that 
when $\epsilon=0$ and $\gamma=0$, (\ref{appendix-modified-kkt}) is identical to the first two equations in (\ref{appendix-kkt}). Given in Theorem \ref{theorem3} that  all binding constraint gradients $\nabla R_i(\boldsymbol{\xi}_{\boldsymbol{\theta}^*},\boldsymbol{\theta}^*)$ are linearly independent at  $\boldsymbol{\theta}^*$ and  the strict complementary holds  at  $\boldsymbol{\theta}^*$, similar to the proof of Theorem~\ref{theorem2},
one can apply the well-known  implicit function theorem \cite{rudin1976principles} to (\ref{appendix-modified-kkt}) in a neighborhood of $(\epsilon,\gamma)={(0,0)}$, leading to  the following claim (i.e., the first-order sensitivity result in Theorem 14 in \cite{fiacco1990nonlinear}): 

\emph{\noindent
	In a neighborhood of $(\epsilon, \gamma)=(0,0)$,
	there exists a unique once continuously differentiable function $\big(\boldsymbol{\theta}^*(\epsilon,\gamma), \boldsymbol{u}^*(\epsilon,\gamma)\big)$, which   satisfies (\ref{appendix-modified-kkt}) and $\big(\boldsymbol{\theta}^*(\epsilon,\gamma), \boldsymbol{u}^*(\epsilon,\gamma)\big)=(\boldsymbol{\theta}^*, \boldsymbol{u}^*)$ when $(\epsilon,\gamma)={(0,0)}$.
} 

\bigskip

Next, we show that the above $\boldsymbol{\theta}^*(\epsilon,\gamma)$ always respects the constraints $R_i{\Big(\boldsymbol{\xi}_{(\boldsymbol{\theta}^*{(\epsilon,\gamma),\gamma})},\boldsymbol{{\theta}}^*(\epsilon,\gamma)\Big)}<0$, $i=1,2,...,l$, for any small $\epsilon>0$ and any small $\gamma>0$, which is the second-part of Claim (a).

In fact, for any inactive constraint,  $R_i{(\boldsymbol{\xi}_{\boldsymbol{\theta}^*},\boldsymbol{{\theta}}^*)}< 0$, due to the continuity of  ${\Big(\boldsymbol{\xi}_{(\boldsymbol{\theta}^*{(\epsilon,\gamma),\gamma})},\boldsymbol{{\theta}}^*(\epsilon,\gamma)\Big)}$, one has
\begin{equation}
R_i{\Big(\boldsymbol{\xi}_{(\boldsymbol{\theta}^*{(\epsilon,\gamma),\gamma})},\boldsymbol{{\theta}}^*(\epsilon,\gamma)\Big)} \rightarrow
R_i{(\boldsymbol{\xi}_{\boldsymbol{\theta}^*},\boldsymbol{{\theta}}^*)}< 0  \quad \text{as} \quad (\epsilon,\gamma) \rightarrow (0,0),
\end{equation}
and thus  $R_i{\Big(\boldsymbol{\xi}_{(\boldsymbol{\theta}^*{(\epsilon,\gamma),\gamma})},\boldsymbol{{\theta}}^*(\epsilon,\gamma)\Big)}<0$ for any small $\epsilon>0$ and $\gamma>0$. For any active constraint, $R_i{(\boldsymbol{\xi}_{\boldsymbol{\theta}^*},\boldsymbol{{\theta}}^*)}= 0$, and since the corresponding $u_i^*>0$ (due to the strict complementarity given in Theorem \ref{theorem3}) and the continuity of $\boldsymbol{u}^*(\epsilon,\gamma)$, one has
\begin{equation}
{u}_i^*(\epsilon,\gamma)   \rightarrow
u_i^* > 0  \quad \text{as} \quad (\epsilon,\gamma) \rightarrow (0,0),
\end{equation} 
and thus 
$u_i^*{(\epsilon,\gamma)} >0$ for small $\epsilon>0$ and consequently
\begin{equation}
R_i{\Big(\boldsymbol{\xi}_{(\boldsymbol{\theta}^*{(\epsilon,\gamma),\gamma})},\boldsymbol{{\theta}}^*(\epsilon,\gamma)\Big)}=    - \frac{\epsilon}{u_i^*{(\epsilon,\gamma)} } <0
\end{equation}
because of (\ref{appendix-modified-kkt}). Therefore, we have proved that for any small $\epsilon>0$ and $\gamma>0$,  $\boldsymbol{\theta}^*(\epsilon,\gamma)$ always respect the constraints $R_i{\Big(\boldsymbol{\xi}_{(\boldsymbol{\theta}^*{(\epsilon,\gamma),\gamma})},\boldsymbol{{\theta}}^*(\epsilon,\gamma)\Big)}<0$, $i=1,2,...,l$. This prove the second part of Claim (a).

\bigskip
From now on, we show that the above $\boldsymbol{\theta}^*(\epsilon,\gamma)$ with any small $\epsilon>0$ and $\gamma>0$ also is a local isolated minimizer to the unconstrained optimization (\ref{appendix-unconstrainedopt}). From the last equation in (\ref{appendix-modified-kkt}), we solve
\begin{equation}
u_i^*(\epsilon,\gamma)=-
\frac{\epsilon}{R_i{\Big(\boldsymbol{\xi}_{(\boldsymbol{\theta}^*{(\epsilon,\gamma),\gamma})},\boldsymbol{{\theta}}^*(\epsilon,\gamma)\Big)}}, \quad i=1,2,...,l,
\end{equation}
and substitute it to the first equation, yielding
\begin{equation}\label{appendix_kkt_unconstrained}
\begin{aligned}
\nabla\ell{\Big(\boldsymbol{\xi}_{(\boldsymbol{\theta}^*{(\epsilon,\gamma),\gamma})},\boldsymbol{{\theta}}^*(\epsilon,\gamma)\Big)}-\sum_{i=1}^{l}
\frac{\epsilon}{R_i{{\Big(\boldsymbol{\xi}_{(\boldsymbol{\theta}^*{(\epsilon,\gamma),\gamma})},\boldsymbol{{\theta}}^*(\epsilon,\gamma)\Big)}}}\nabla R_i{{\Big(\boldsymbol{\xi}_{(\boldsymbol{\theta}^*{(\epsilon,\gamma),\gamma})},\boldsymbol{{\theta}}^*(\epsilon,\gamma)\Big)}}
=\boldsymbol{0}.
\end{aligned}
\end{equation}
One can find that the obtained equation in (\ref{appendix_kkt_unconstrained}) is exactly the first-order optimality condition (KKT condition) for the unconstrained optimization in Problem \ref{equ_problem_penalty_approx} in (\ref{appendix-unconstrainedopt}), and this indicates that $\boldsymbol{\theta}^*(\epsilon,\gamma)$ satisfies the KKT condition for Problem \ref{equ_problem_penalty_approx}. To further show that  $\boldsymbol{\theta}^*(\epsilon,\gamma)$ is a local isolated minimizing solution to Problem \ref{equ_problem_penalty_approx} in (\ref{appendix-unconstrainedopt}), we only need to verify the second-order condition, that is, for any nonzero $\boldsymbol{\theta}\neq\boldsymbol{0}$, 
\begin{equation}\label{appendix-seconrodercondition}
\boldsymbol{\theta}^\prime\Big(\nabla^2  W\big({\boldsymbol{\theta}^*(\epsilon,\gamma)},\epsilon,\gamma \big)
\Big)\boldsymbol{\theta}>0,
\end{equation} for any small $\epsilon>0$ and $\gamma>0$, 
 which will be proved next. 

Proof by contradiction: suppose that the second-order condition (\ref{appendix-seconrodercondition}) is false. Then, there must exist a sequence of $(\epsilon_k, \gamma_k)>0$ and a sequence of $\boldsymbol{\theta}_k$ for $k=1,2,...$ such that $(\epsilon_k,\gamma_k)\rightarrow (0,0)$  and
\begin{equation}\label{appendix-assumption}
\boldsymbol{\theta}_k^\prime\big(\nabla^2  W\big({\boldsymbol{\theta}^*(\epsilon_k,\gamma_k)},\epsilon_k,\gamma_k \big)
\Big)\boldsymbol{\theta}_k\leq 0.
\end{equation}
as  $k\rightarrow +\infty$. 
Without loss of generality, assume $\norm{\boldsymbol{\theta}_k}=1$ for all $k$. Select a   convergent sub-sequence of $\boldsymbol{\theta}_k$, relabel the sequence $\boldsymbol{\theta}_k$ for convenience, and call the limit $\boldsymbol{\bar\theta}$, that is, $\boldsymbol{\theta}_k\rightarrow \boldsymbol{\bar\theta}$ and $(\epsilon_k,\gamma_k)\rightarrow 0$ as $k\rightarrow +\infty$. Then, 
\begin{equation}\label{appendix-limit}
\begin{aligned}
&\lim\limits_{k\rightarrow +\infty}\boldsymbol{\theta}_k^\prime\big(\nabla^2  W\big({\boldsymbol{\theta}^*(\epsilon_k,\gamma_k)},\epsilon_k,\gamma_k \big)
\Big)\boldsymbol{\theta}_k\\
=&\lim\limits_{k\rightarrow +\infty}\Bigg(\boldsymbol{\theta}_k^\prime\Big(
\nabla^2  L({\boldsymbol{\theta}^*(\epsilon_k,\gamma_k)},\boldsymbol{u}^*(\epsilon_k,\gamma_k))+\sum_{i=1}^{l}
\frac{\epsilon_k}{(R_i{(\epsilon_k,\gamma_k)})^2}
(\nabla R_i{(\epsilon_k,\gamma_k)}\nabla R_i{(\epsilon_k,\gamma_k)}^\prime)
\Big)\boldsymbol{\theta}_k
\Bigg)\\
=&\lim\limits_{k\rightarrow +\infty}\Bigg(\boldsymbol{\theta}_k^\prime\Big(
\nabla^2  L({\boldsymbol{\theta}^*(\epsilon_k,\gamma_k)},\boldsymbol{u}^*(\epsilon_k,\gamma_k))
\Big)\boldsymbol{\theta}_k
\Bigg)+\lim\limits_{k\rightarrow +\infty}
\Bigg(
\sum_{i=1}^{l}
\frac{\epsilon_k}{(R_i{(\epsilon_k,\gamma_k)})^2}
(\nabla R_i{(\epsilon_k,\gamma_k)}^\prime\boldsymbol{\theta}_k)^2
\Bigg)\\
= &\quad\boldsymbol{\bar\theta}^\prime\Big(
\nabla^2  L({\boldsymbol{\theta}^*},\boldsymbol{u}^*)
\Big)\boldsymbol{\bar\theta} +\lim\limits_{k\rightarrow +\infty}
\Bigg(
\sum_{i=1}^{l}
\frac{\epsilon_k}{(R_i{(\epsilon_k,\gamma_k)})^2}
(\nabla R_i{(\epsilon_k,\gamma_k)}^\prime\boldsymbol{\theta}_k)^2
\Bigg),
\end{aligned}
\end{equation}
where we write $R_i{(\epsilon_k,\gamma_k)}=R_i(\boldsymbol{\xi}_{(\boldsymbol{\theta}^*(\epsilon_k,\gamma_k),\gamma_k)}, \boldsymbol{\theta}^*(\epsilon_k,\gamma_k))$  and $\nabla R_i{(\epsilon_k)}=\nabla R_i(\boldsymbol{\xi}_{(\boldsymbol{\theta}^*(\epsilon_k,\gamma_k),\gamma_k)}, \boldsymbol{\theta}^*(\epsilon_k,\gamma_k))$ for notation convenience, and the last line is because $L(\boldsymbol{\theta}, \boldsymbol{u})$ in (\ref{Lagproblemp}) is twice-continuously differentiable with respect to $(\boldsymbol{\theta}, \boldsymbol{u})$ near $(\boldsymbol{\theta}^*, \boldsymbol{u}^*)$, and $\big({\boldsymbol{\theta}^*(\epsilon,\gamma)}, \boldsymbol{u}^*(\epsilon,\gamma)\big)$ is once-continuously differentiable with respect to $(\epsilon,\gamma)$  near $(0,0)$. In (\ref{appendix-limit}), we consider two cases for $\boldsymbol{\bar{\theta}}$:

\emph{Case 1:} $\norm{\boldsymbol{\bar{\theta}}}=1$ and there exists at least an active inequality constraint $R_i(\boldsymbol{\xi}_{\boldsymbol{\theta}^*},\boldsymbol{\theta}^*)=0$, such that $\boldsymbol{\bar{\theta}}^\prime\nabla R_i(\boldsymbol{\xi}_{\boldsymbol{\theta}^*},\boldsymbol{\theta}^*)\neq0$. Then,
\begin{equation}\label{appendix-limit2}
\begin{aligned}
&\lim\limits_{k\rightarrow +\infty}
\Bigg(
\sum_{i=1}^{l}
\frac{\epsilon_k}{(R_i{(\epsilon_k,\gamma_k)})^2}
(\nabla R_i{(\epsilon_k,\gamma_k)}^\prime\boldsymbol{\theta}_k)^2
\Bigg)\\
=&\lim\limits_{k\rightarrow +\infty}
\Bigg(
\sum_{i=1}^{l}
\frac{-u_i(\epsilon_k,\gamma_k)}{R_i{(\epsilon_k,\gamma_k)}}
(\nabla R_i{(\epsilon_k,\gamma_k)}^\prime\boldsymbol{\theta}_k)^2
\Bigg)=+\infty.
\end{aligned}
\end{equation}
This is because the following term corresponding to such active constraint has
\begin{equation}
\lim\limits_{k\rightarrow +\infty}\frac{-u_i(\epsilon_k,\gamma_k)}{R_i(\boldsymbol{\xi}_{(\boldsymbol{\theta}^*(\epsilon_k,\gamma_k),\gamma_k)}, \boldsymbol{\theta}^*(\epsilon_k,\gamma_k))}=+\infty.
\end{equation}
due to the strict complementarity given in Theorem \ref{theorem3}. 
Therefore, (\ref{appendix-limit}) will have
\begin{equation}
\lim\limits_{k\rightarrow +\infty}\boldsymbol{\theta}_k^\prime\Big(\nabla^2  W\big({\boldsymbol{\theta}^*(\epsilon_k,\gamma_k)},\epsilon_k,\gamma_k \big)
\Big)\boldsymbol{\theta}_k=+\infty,
\end{equation}
which contradicts the assumption in (\ref{appendix-assumption}) in that 
\begin{equation}
\lim\limits_{k\rightarrow +\infty}\boldsymbol{\theta}_k^\prime\Big(\nabla^2  W\big({\boldsymbol{\theta}^*(\epsilon_k,\gamma_k)},\epsilon_k,\gamma_k \big)
\Big)\boldsymbol{\theta}_k\leq 0.
\end{equation}

\emph{Case 2:} $\norm{\boldsymbol{\bar{\theta}}}=1$ and for any active constraint $R_i(\boldsymbol{\xi}_{\boldsymbol{\theta}^*},\boldsymbol{\theta}^*)=0$ (and $u_i^*>0$ due to strict complementarity given in Theorem \ref{theorem3}), $\boldsymbol{\bar{\theta}}^\prime\nabla R_i(\boldsymbol{\xi}_{\boldsymbol{\theta}^*},\boldsymbol{\theta}^*)=0$. Then, from (\ref{appendix-limit}),
\begin{equation}\label{theorem3_case2}
\begin{aligned}
&\lim\limits_{k\rightarrow +\infty}\boldsymbol{\theta}_k^\prime\big(\nabla^2  W\big({\boldsymbol{\theta}^*(\epsilon_k,\gamma_k)},\epsilon_k,\gamma_k \big)
\Big)\boldsymbol{\theta}_k\geq  \boldsymbol{\bar\theta}^\prime\Big(
\nabla^2  L({\boldsymbol{\theta}^*},\boldsymbol{u}^*)
\Big)\boldsymbol{\bar\theta}>0,
\end{aligned}
\end{equation}
where the last inequality is because of the second-order condition in (\ref{appendix-second}) satisfied for  $\boldsymbol{\theta}^*$  given in Theorem \ref{theorem3}. The obtained (\ref{theorem3_case2}) also contradicts the assumption in (\ref{appendix-assumption}).

Combining the above two cases, we can conclude  that  
given any small $\epsilon>0$ and $\gamma>0$,  $\boldsymbol{\theta}^*(\epsilon,\gamma)$  satisfies both the KKT condition  (\ref{appendix_kkt_unconstrained}) and   the second-order condition  (\ref{appendix-seconrodercondition}) for $W(\boldsymbol{\theta},\epsilon,\gamma)$. Thus, one can assert that 
$\boldsymbol{\theta}^*(\epsilon,\gamma)$
is a local isolated  minimizer to the unconstrained optimization $W(\boldsymbol{\theta},\epsilon,\gamma)$ in (\ref{appendix-unconstrainedopt}), i.e., Problem \ref{equ_problem_penalty_approx}.
This completes the proof of the Claim (a) in Theorem \ref{theorem3}.

\bigskip
\subsection{Proof of Claim (b)}

From the previous proof for Claim (a), we have the following conclusions: first, for any  $(\epsilon,\gamma)$ in a neighborhood of $(0,0)$, there exists a \emph{unique} once-continuously differentiable function  $\big(\boldsymbol{\theta}^*(\epsilon,\gamma),\boldsymbol{u}^*(\epsilon,\gamma)\big)$,  satisfying (\ref{appendix-modified-kkt}); second, additionally provided small $\epsilon>0$ and $\gamma>0$, such   $\boldsymbol{\theta}^*(\epsilon,\gamma)$ is also a local isolated minimizer to the well-defined unconstrained minimization $W(\boldsymbol{\theta},\epsilon,\gamma)$ in (\ref{appendix-unconstrainedopt}); and third, when  $(\epsilon,\gamma)=(0,0)$,    (\ref{appendix-modified-kkt}) becomes the KKT condition for Problem \ref{equ_problem}, whose solution    $(\boldsymbol{\theta}^*, \boldsymbol{u}^*)$ must satisfy. Therefore,  due to the uniqueness  and continuity of the function $(\boldsymbol{\theta}^*(\epsilon,\gamma),\boldsymbol{u}^*(\epsilon,\gamma))$  near $(\epsilon,\gamma)=(0,0)$, one can obtain
\begin{equation}
\boldsymbol{\theta}^*(\epsilon,\gamma)\rightarrow \boldsymbol{\theta}^*(0,0)=\boldsymbol{\theta}^*,\quad \text{as}\quad (\epsilon,\gamma)\rightarrow (0,0).
\end{equation}
This completes the proof of Claim (b) in Theorem \ref{theorem3}.

\bigskip
\subsection{Proof of Claim (c)}

To prove Claim (c) in Theorem \ref{theorem3}, we use the following facts: first, as proved in Claim (a), for any small $\epsilon>0$ and $\gamma>0$,  $\boldsymbol{\theta}^*(\epsilon,\gamma)$   always respects the constraints $R_i{(\boldsymbol{\xi}_{(\boldsymbol{\theta}^*(\epsilon,\gamma),\gamma)},\boldsymbol{{\theta}}^*(\epsilon,\gamma))}<0$, $i=1,2,...,l$; second, as also proved in Claim (a), $\boldsymbol{\theta}^*(\epsilon,\gamma)$ is  differentiable with respect to $(\epsilon,\gamma)$ near $(0,0)$ and  $\boldsymbol{\theta}^*(\epsilon,\gamma)\rightarrow\boldsymbol{\theta}^*$ as $(\epsilon,\gamma)\rightarrow(0,0)$; and third, as proved in Theorem \ref{theorem2}, $\boldsymbol{\xi}_{(\boldsymbol{\theta},\gamma)}$ is a differentiable  function of $(\boldsymbol{\theta},\gamma)$ near $(\boldsymbol{\theta}^*,0)$. All these  facts lead to that for small $\gamma>0$, $R_i{(\boldsymbol{\xi}_{(\boldsymbol{\theta},\gamma)},\boldsymbol{{\theta}})}$, $i=1,2,...,l$, is also a continuous function of $\boldsymbol{\theta}$ near  ${\boldsymbol{\theta}^*(\epsilon,\gamma)}$, and 
\begin{equation}
R_i{(\boldsymbol{\xi}_{(\boldsymbol{\theta},\gamma)},\boldsymbol{{\theta}})} \rightarrow R_i{\big(\boldsymbol{\xi}_{(\boldsymbol{\theta}^*(\epsilon,\gamma),\gamma)},\boldsymbol{{\theta}}^*(\epsilon,\gamma)\big)}<0, \quad \text{as} \quad \boldsymbol{\theta}\rightarrow {\boldsymbol{\theta}^*(\epsilon,\gamma)}, \quad \forall i=1,2,...,l.
\end{equation}
Thus $R_i{(\boldsymbol{\xi}_{(\boldsymbol{\theta},\gamma)},\boldsymbol{{\theta}})}<0$, $i=1,2,...,l$, holds for any ${\boldsymbol{\theta}}$ in a small neighborhood of ${\boldsymbol{\theta}^*(\epsilon,\gamma)}$ with small $\epsilon>0$ and small $\gamma>0$. This completes the proof of Claim (c) in Theorem \ref{theorem3}.

\bigskip
With the above proofs for Claims (a),  (b), and (c), we have completed the proof of  Theorem \ref{theorem3}.\qed

\newpage
\section{Algorithms for Safe PDP}\label{appendix.algorithms}

 We have implemented  Safe PDP in Python and made it as a stand-alone package with friendly interfaces. Please download at \url{https://github.com/wanxinjin/Safe-PDP}.

\subsection{Algorithm for  Theorem \ref{theorem1}}\label{appendix.algorithm1}

\vspace{-8pt}

\begin{algorithm2e}[H]
		\label{alg_theorem1}
		
	\SetAlgoNoLine
	\SetKwComment{Comment}{$\triangleright$\ }{}
	
	\smallskip
	\SetKwInput{given}{Input}
	\given{
		 $\boldsymbol{{\xi}}_{\boldsymbol{\theta}}$,  with the costates  	$\boldsymbol{\lambda}_{1:T}^{{\boldsymbol{{\theta}}}}$ and multiplies $\boldsymbol{v}_{0:T}^{{\boldsymbol{{\theta}}}}$ and $\boldsymbol{w}_{0:T}^{{\boldsymbol{{\theta}}}}$  from solving    Problem \ref{equ_traj} 
	} 
	
	\Indp\Indp 
	
	\bigskip
	
	\textbf{def} Identify\_Active\_Inequality\_Constraints (a small threshold $\delta>0$):
	
	\Indp
	
	\smallskip
	$\boldsymbol{\bar{g}}_t(\boldsymbol{x}_t,\boldsymbol{u}_t,\boldsymbol{\theta})=\col\{{g}_{t,i}\,|\, {g}_{t,i}(\boldsymbol{x}_t^{\boldsymbol{\theta}},\boldsymbol{u}_t^{\boldsymbol{\theta}},\boldsymbol{\theta}){\geq} {-}\delta, \,\,\, i=1,2,...,q_t\}$, \,\,$t=0,1,...,T{-}1$\;
	
	\smallskip
		$\boldsymbol{\bar{g}}_T(\boldsymbol{x}_T,\boldsymbol{\theta})=\col\{{g}_{T,i}\,|\, {g}_{T,i}(\boldsymbol{x}_T^{\boldsymbol{\theta}},\boldsymbol{\theta}){\geq} {-}\delta, \,\,\, i=1,2,...,q_T\}$\;
		\medskip
		
		\textbf{Return:}    $\boldsymbol{\bar{g}}_t(\boldsymbol{x}_t,\boldsymbol{u}_t,\boldsymbol{\theta})$ and $\boldsymbol{\bar{g}}_T(\boldsymbol{x}_T,\boldsymbol{\theta})$
	 
	\medskip
	
	\Indm
		
	\smallskip

	Compute the derivative matrices $L_t^{xx}$, $L_t^{xu}$,  $L_t^{uu}$, $L_t^{x\theta}$, $L_t^{u\theta}$, $L_T^{xx}$,  $L_T^{x\theta}$, $F_t^{x}$, $F_t^{u}$, $F_t^{\theta}$, $H_t^{x}$, $H_t^{u}$, $H_t^{\theta}$, $H_T^{x}$, $H_T^{\theta}$, $\bar{G}_t^{x}$, $\bar{G}_t^{u}$, $\bar{G}_t^{\theta}$, $\bar{G}_T^{x}$, $\bar{G}_T^{\theta}$ to establish  $\boldsymbol{\overline\Sigma}(\boldsymbol{{\xi}}_{\boldsymbol{\theta}})$ in (\ref{equ_aux})\;
	
	\bigskip

	\textbf{def} Equality\_Constrained\_LQR\_ Solver ( $\boldsymbol{\overline\Sigma}(\boldsymbol{{\xi}}_{\boldsymbol{\theta}})$ ):

	\Indp

	\smallskip
	Implementation of the equality constrained LQR algorithm in \cite{laine2019efficient}\;
	
	\medskip

	\textbf{Return:} $\{{{X}}_{0:T}^{\boldsymbol{\theta}},{{U}}_{0:T-1}^{\boldsymbol{\theta}}\}$
	
	\bigskip

	\Indm\Indm\Indm
	\SetKwInput{return}{Return}
	\return{$\frac{\partial  \boldsymbol{{\xi}}_{\boldsymbol{{\theta}}}}{\partial \boldsymbol{{\theta}}}=\{{{X}}_{0:T}^{\boldsymbol{\theta}},{{U}}_{0:T-1}^{\boldsymbol{\theta}}\}$}
	
	\caption{Solving $\frac{\partial  \boldsymbol{{\xi}}_{\boldsymbol{{\theta}}}}{\partial \boldsymbol{{\theta}}}$ by establishing auxiliary control system $\boldsymbol{\overline\Sigma}(\boldsymbol{{\xi}}_{\boldsymbol{\theta}})$}

\end{algorithm2e}

Note that  	$\boldsymbol{\lambda}_{1:T}^{{\boldsymbol{{\theta}}}}$, $\boldsymbol{v}_{0:T}^{{\boldsymbol{{\theta}}}}$, and $\boldsymbol{w}_{0:T}^{{\boldsymbol{{\theta}}}}$ are normally the by-product outputs of an optimal control solver  \cite{andersson2019casadi}, and can also be obtained by solving a linear equation of C-PMP (\ref{CPMP}) given $\boldsymbol{{\xi}}_{\boldsymbol{\theta}}$, as done in \cite{jin2019pontryagin}. Also note that the threshold $\delta$ to determine the active inequality constraints can be set according to the accuracy of the solver; in our experiments, we use $\delta=10^{-3}$.

\subsection{Algorithm for  Theorem \ref{theorem2}} \label{appendix.algorithm2}
\vspace{-8pt}

\begin{algorithm2e}[H]
	\label{alg_theorem2}
	
	\SetAlgoNoLine
	\SetKwComment{Comment}{$\triangleright$\ }{}
	
	\smallskip
	
	\SetKwInput{given}{Input}
	\given{
		The constrained optimal control system $\boldsymbol{\Sigma}({\boldsymbol{\theta}})$ and a choice of small  $\gamma>0$  		 
	} 
	
	\Indp\Indp 
	
	\medskip
	
	Convert  $\boldsymbol{\Sigma}({\boldsymbol{\theta}})$ to an unconstrained optimal control  system $\boldsymbol{\Sigma}(\boldsymbol{\theta},\gamma)$ in (\ref{equ_oc_penality}) by adding all  constraints in $\boldsymbol{\Sigma}({\boldsymbol{\theta}})$ to its control cost through  barrier functions with barrier parameter $\gamma$\;

    \bigskip
	
	\tcc{Below is an implmentation of uncsontrained PDP \cite{jin2019pontryagin}}
	
	\smallskip
	
	\textbf{def}  Optimal\_Control\_Solver ( $\boldsymbol{\Sigma}(\boldsymbol{\theta},\gamma)$ ): 

	\Indp

	\smallskip
	Implementation of any trajectory optimization algorithms, such  as iLQR \cite{li2004iterative} and DDP \cite{jacobson1970differential}, or use any optimal control solver \cite{andersson2019casadi}\;
	
	\medskip
	
	\textbf{Return:}  $\boldsymbol{\xi}_{(\boldsymbol{\theta},\gamma)}$ 
	
	\medskip

	\Indm

	Use $\boldsymbol{\xi}_{(\boldsymbol{\theta},\gamma)}$ to compute the derivative matrices $\hat{L}_t^{xu}$,  $\hat L_t^{uu}$, $\hat L_t^{x\theta}$, $\hat L_t^{u\theta}$, $\hat L_T^{xx}$,  $\hat L_T^{x\theta}$, $F_t^{x}$, $F_t^{u}$, $F_t^{\theta}$ to establish the  auxiliary control system  ${\boldsymbol{\overline\Sigma}}(\boldsymbol{\xi}_{(\boldsymbol{\theta},\gamma)})$ in (\ref{equ_aux2}) \;
	
	\medskip

	\textbf{def}\,\, LQR\_ Solver ( ${\boldsymbol{\overline\Sigma}}(\boldsymbol{\xi}_{(\boldsymbol{\theta},\gamma)})$ ):

	\Indp

	\smallskip
	Implementation of any LQR algorithm such as  Lemma 2 in \cite{jin2019pontryagin}\;
	
	\medskip

	\textbf{Return:} $\left\{{{X}}_{0:T}^{(\boldsymbol{\theta},\gamma)},{{U}}_{0:T-1}^{(\boldsymbol{\theta},\gamma)}\right\}=\frac{\partial  \boldsymbol{{\xi}}_{(\boldsymbol{\theta},\gamma)}}{\partial \boldsymbol{{\theta}}}$
	
	\bigskip

	\Indm\Indm\Indm
	\SetKwInput{return}{Return}
	\return{$\boldsymbol{\xi}_{(\boldsymbol{\theta},\gamma)}$ and 
		$\frac{\partial  \boldsymbol{{\xi}}_{(\boldsymbol{\theta},\gamma)}}{\partial \boldsymbol{{\theta}}}$}
	
	\caption{Safe unconstrained  approximations for $\boldsymbol{\xi}_{\boldsymbol{\theta}}$ and  $\frac{\partial\boldsymbol{\xi}_{\boldsymbol{\theta}}}{\partial\boldsymbol{\theta}}$}

\end{algorithm2e}

		Note that the auxiliary control system ${\boldsymbol{\overline\Sigma}}(\boldsymbol{\xi}_{(\boldsymbol{\theta},\gamma)})$ corresponding to  $\boldsymbol{\Sigma}(\boldsymbol{\theta},\gamma)$ is

\begin{longfbox}[padding-top=1pt,margin-top=0pt, padding-bottom=4pt, margin-bottom=0pt]
	\mathleft
	\begin{equation}\label{equ_aux2}
	{\boldsymbol{\overline\Sigma}}(\boldsymbol{\xi}_{(\boldsymbol{\theta},\gamma)}):\,\,\,
	\begin{aligned}
	\emph{control cost} &\quad \bar{W}=\small\Tr\sum_{t=0}^{T{-}1}
	\left(\frac{1}{2}\begin{bmatrix}
	X_t\\
	U_t
	\end{bmatrix}^\prime\begin{bmatrix}
	\hat{L}_t^{xx}& \hat{L}_t^{xu}\\
	\hat{L}_t^{ux}& \hat{L}_t^{uu}
	\end{bmatrix}\begin{bmatrix}
	X_t\\
	U_t
	\end{bmatrix}+
	\begin{bmatrix}
	\hat{L}_t^{x\theta}\\
	\hat{L}_t^{u\theta}
	\end{bmatrix}^\prime\begin{bmatrix}
	X_t\\
	U_t
	\end{bmatrix}
	\right)
	\\
	&\qquad\quad+\small\Tr\left(\frac{1}{2}X_T^\prime \hat{L}_T^{xx}X_T+(\hat{L}_T^{x\theta})^\prime X_T\right)\\
	\text{subject to}&\\
	\emph{dynamics} &\quad {X}_{t+1}=F_t^xX_t
	+F_t^uU_{t}+F_t^\theta\quad \text{with}\quad {X}_{0}=X_{0}^{\boldsymbol{\theta}}.
	\end{aligned}
	\end{equation}	\mathcenter
\end{longfbox}

Here, $\hat{L}_t$, $t=0,1,..., T{-}1,$  and $\hat{L}_T$ are the {Hamiltonian}, defined in (\ref{hamil-penalty-oc}), for the unconstrained optimal control  system
${\boldsymbol{\Sigma}}(\boldsymbol{\theta},\gamma)$. The derivative (coefficient) matrices $\hat{L}_t^{xu}$,  $\hat L_t^{uu}$, $\hat L_t^{x\theta}$, $\hat L_t^{u\theta}$, $\hat L_T^{xx}$,  $\hat L_T^{x\theta}$, $F_t^{x}$, $F_t^{u}$, $F_t^{\theta}$ in (\ref{equ_aux2}) are defined in the similar notation convention as in (\ref{equ_aux}).

\vspace{10pt}

\subsection{Algorithm for  Theorem \ref{theorem3}}
\label{appendix.algorithm3}

\begin{algorithm2e}[H]
	\label{alg_theorem3}
	
	\SetAlgoNoLine
	\SetKwComment{Comment}{$\triangleright$\ }{}
	
	\smallskip
	
	\SetKwInput{given}{Input}
	\given{
		  Small  barrier parameter $\epsilon>0$ for outer-level  and  $\gamma>0$ for inner-level, initialization $\boldsymbol{\theta}_0$
	} 

	\medskip
	
	\Indp\Indp 
	
		\tcc{Convert   inner-level  into its safe unconstrained approximation}
		\medskip
	
\nl	Convert the inner-level constrained control system $\boldsymbol{\Sigma}({\boldsymbol{\theta}})$ in (\ref{equ_oc}) into an unconstrained   system $\boldsymbol{\Sigma}(\boldsymbol{\theta},\gamma)$ in (\ref{equ_oc_penality}) by adding all  constraints  to its control cost through  barrier functions with the inner-level barrier parameter $\gamma>0$\;

	\medskip
	
			\tcc{Convert   outer-level  into its safe unconstrained approximation}
	\medskip
	
\nl		Convert the constrained Problem \ref{equ_problem} to an unconstrained Problem \ref{equ_problem_penalty_approx}  by adding all task constraints $R_i$ to the task loss  through  barrier functions with the outer-level barrier parameter  $\epsilon>0$\;
		
	\bigskip
	
			\tcc{Gradient-based update for $\boldsymbol{\theta}$}
	\medskip
	
	\For{$k=0,1,2,\cdots$}{

		\medskip
	
\nl	Apply Algorithm \ref{alg_theorem2} to the inner-level safe unconstrained approximation  system  $\boldsymbol{\Sigma}({\boldsymbol{\theta}_k},\gamma)$ to compute   $\boldsymbol{\xi}_{(\boldsymbol{\theta}_k,\gamma)}$ and 
	$\frac{\partial  \boldsymbol{{\xi}}_{(\boldsymbol{\theta},\gamma)}}{\partial \boldsymbol{{\theta}}}\rvert_{\boldsymbol{\theta}_k}$\;

	\medskip
\nl	For the outer-level unconstrained  Problem \ref{equ_problem_penalty_approx} with objective function $\small W\big(\boldsymbol{\theta},\epsilon,\gamma\big)=\ell\big(\boldsymbol{\xi}_{(\boldsymbol{\theta}, \gamma)},\boldsymbol{\theta}\big)-\epsilon\sum_{i=1}^{l}\ln\Big({-}R_i\big(\boldsymbol{\xi}_{(\boldsymbol{\theta}, \gamma)},\boldsymbol{\theta}\big)\Big)$, compute the partial gradients 	$\frac{\partial  W}{\partial \boldsymbol{{\theta}}}\rvert_{\boldsymbol{\theta}_k}$ and $\frac{\partial  W}{\partial   \boldsymbol{{\xi}}_{(\boldsymbol{\theta},\gamma)}
	}\rvert_{\boldsymbol{\xi}_{(\boldsymbol{\theta}_k,\gamma)}}$\;

	\medskip
\nl	Apply the chain rule to obtain the gradient of the outer-level unconstrained objective $W\big(\boldsymbol{\theta},\epsilon,\gamma\big)$ with respect to $\boldsymbol{\theta}$, i.e.,
	$\frac{d  W}{d \boldsymbol{{\theta}}}\rvert_{\boldsymbol{\theta}_k}=\frac{\partial  W}{\partial \boldsymbol{{\theta}}}\rvert_{\boldsymbol{\theta}_k}+\frac{\partial  W}{\partial   \boldsymbol{{\xi}}_{(\boldsymbol{\theta},\gamma)}
	}\rvert_{\boldsymbol{\xi}_{(\boldsymbol{\theta}_k,\gamma)}}\frac{\partial  \boldsymbol{{\xi}}_{(\boldsymbol{\theta},\gamma)}}{\partial \boldsymbol{{\theta}}}\rvert_{\boldsymbol{\theta}_k}$\;

		\medskip
\nl Gradient-based update: $\boldsymbol{\theta}_{k+1}=\boldsymbol{\theta}_{k}-\eta\frac{d  W}{d \boldsymbol{{\theta}}}\rvert_{\boldsymbol{\theta}_k}$ with $\boldsymbol{\eta}$ being the learning rate\;

\medskip

}

\bigskip
	\Indm\Indm
	\SetKwInput{return}{Return}
	\return{$\boldsymbol{\theta}^*{(\epsilon, \gamma)}$ for the given barrier parameters $\epsilon>0$ and $\gamma>0$}

	\caption{Safe PDP  to solve Problem \ref{equ_problem}}

\end{algorithm2e}

Note that after obtaining $\boldsymbol{\theta}^*(\epsilon, \gamma)$ from Algorithm \ref{alg_theorem3}, one can  sequentially refine  $\boldsymbol{\theta}^*(\epsilon, \gamma)$ by choosing a sequence of $\{(\epsilon, \gamma)\}$ such that  $(\epsilon, \gamma)\rightarrow (0,0)$. 

Also note that in the case where the original inner-level control system $\boldsymbol{\Sigma}({\boldsymbol{\theta}})$ is already unconstrained, such as in the applications of safe policy optimization and safe motion planning in Section \ref{section.applications}, please modify  lines 1 and 3 in  Algorithm \ref{alg_theorem3} and just compute   exact   $\boldsymbol{\xi}_{\boldsymbol{\theta}_k}$ and 
$\frac{\partial  \boldsymbol{{\xi}}_{\boldsymbol{\theta}}}{\partial \boldsymbol{{\theta}}}\rvert_{\boldsymbol{\theta}_k}$ by following  PDP \cite{jin2019pontryagin}.

\newpage

\section{Experiment Details}\label{appendix.experimentdetails}

The proposed Safe PDP has been evaluated in different simulated systems in Table \ref{experimenttable}, where each system has the immediate constraints $\boldsymbol{g}(\boldsymbol{\theta}_{\text{cstr}})$  on  both its state and input during the entire time horizon (around $T=50$). For the detailed description and physical models of each system in Table \ref{experimenttable}, we refer the reader to \cite{jin2019pontryagin} and its accompanying codes. We have developed the Python code of  Safe PDP as a stand-alone package, which can be accessed at    \url{https://github.com/wanxinjin/Safe-PDP}.

\subsection{Safe Policy Optimization}\label{appendix.application.spo}

In this experiment, we apply Safe PDP to perform  safe policy optimization for the systems  in Table \ref{experimenttable}. In  Problem \ref{equ_problem}, we  set the details of   $\boldsymbol{\Sigma}(\boldsymbol{\theta})$ as   (\ref{sys.opt}), where the dynamics $\boldsymbol{f}$ is learned from demonstrations in Section \ref{appendix.application.mpc}, and the policy $\boldsymbol{u}_t=\boldsymbol{\pi}(\boldsymbol{x}_t,\boldsymbol{\theta})$ is represented using a neural network (NN) with $\boldsymbol{\theta}$   the NN parameter.  In our experiment, we have  used a fully-connected feedforward NN  to represent the policy;  the number of nodes in the NN is  ${n{-}n{-}m}$ (meaning that the input layer has ${n}$ nodes, hidden layer ${n}$ nodes, and output layer   ${m}$ nodes, with ${n}$ and $m$ the dimensions of the system state and input, respectively); and the activation function of the NN is $\tanh$.  
In Problem \ref{equ_problem},   set the task loss  $\ell(\boldsymbol{\xi}_{\boldsymbol{\theta}},\boldsymbol{\theta})$ as the  control cost $J(\boldsymbol{\theta}_{\text{obj}})$, and set the task constraints $R_i(\boldsymbol{\xi}_{\boldsymbol{\theta}},\boldsymbol{\theta})$ as the system constraints  $\boldsymbol{g}(\boldsymbol{\theta}_{\text{cstr}})$, both   in Table \ref{experimenttable}, with both $\boldsymbol{\theta}_{\text{obj}}$ and $\boldsymbol{\theta}_{\text{cstr}}$ known.

 Note that since the parameterized   $\boldsymbol{\Sigma}(\boldsymbol{\theta})$ in  (\ref{sys.opt})   does not include the control cost  $J$ anymore,  solving Problem \ref{equ_traj} for $\boldsymbol{\xi}_{\boldsymbol{\theta}}$ becomes a simple integration of (\ref{sys.opt}) from $t=0$ to $T{-}1$, and the auxiliary control system $\boldsymbol{\overline{\Sigma}}(\boldsymbol{\xi}_{\boldsymbol{\theta}})$ in (\ref{equ_aux}) to compute  $ \frac{\partial\boldsymbol{\xi}_{\boldsymbol{\theta}}}{ \partial\boldsymbol{\theta}}$ is simplified to a feedback control system \cite{jin2019pontryagin}  below:
 
\begin{longfbox}[padding-top=-3pt, margin-top=-4pt, padding-bottom=-1pt]
	\mathleft
	\begin{equation}\label{mode3controlback}
	\boldsymbol{\overline\Sigma}(\boldsymbol{\xi}_{\boldsymbol{\theta}}):\qquad\qquad
	\begin{aligned}
	\text{dynamics:}&\quad {{X}}_{t+1}^{\boldsymbol{{\theta}}}=
	F_t^{x}{{X}}^{\boldsymbol{{\theta}}}_{t}+F_t^{u}{{U}}^{\boldsymbol{\theta}}_{t} \quad \text{with} \quad {{X}}_{0}=\boldsymbol{0},\\
	\text{control policy:} &\quad \quad{{U}}^{\boldsymbol{\theta}}_{t}=U_t^{{x}}X_{t}^{\boldsymbol{\theta}}+U_t^{{e}}.
	\end{aligned}
	\end{equation} 
	\mathcenter
\end{longfbox} 

Here,
$
{{U}}^x_t{=}\frac{\partial \boldsymbol{\pi}_t}{\partial \boldsymbol{x}_t^{\boldsymbol{\theta}}}$ and ${{U}}_t^e{=}\frac{\partial \boldsymbol{\pi}_t}{\partial \boldsymbol{\theta}}$.
Integrating  (\ref{mode3controlback}) from $t=0$ to $T{-}1$ leads to
$
	\{{{X}}_{0:T}^{\boldsymbol{\theta}},{{U}}_{0:T-1}^{\boldsymbol{\theta}}\}{=} \frac{\partial \boldsymbol{{\xi}}_{\boldsymbol{\theta}}}{\partial \boldsymbol{\theta}}.
$

\begin{figure}[h]
	\centering
	\includegraphics[width=0.7\linewidth]{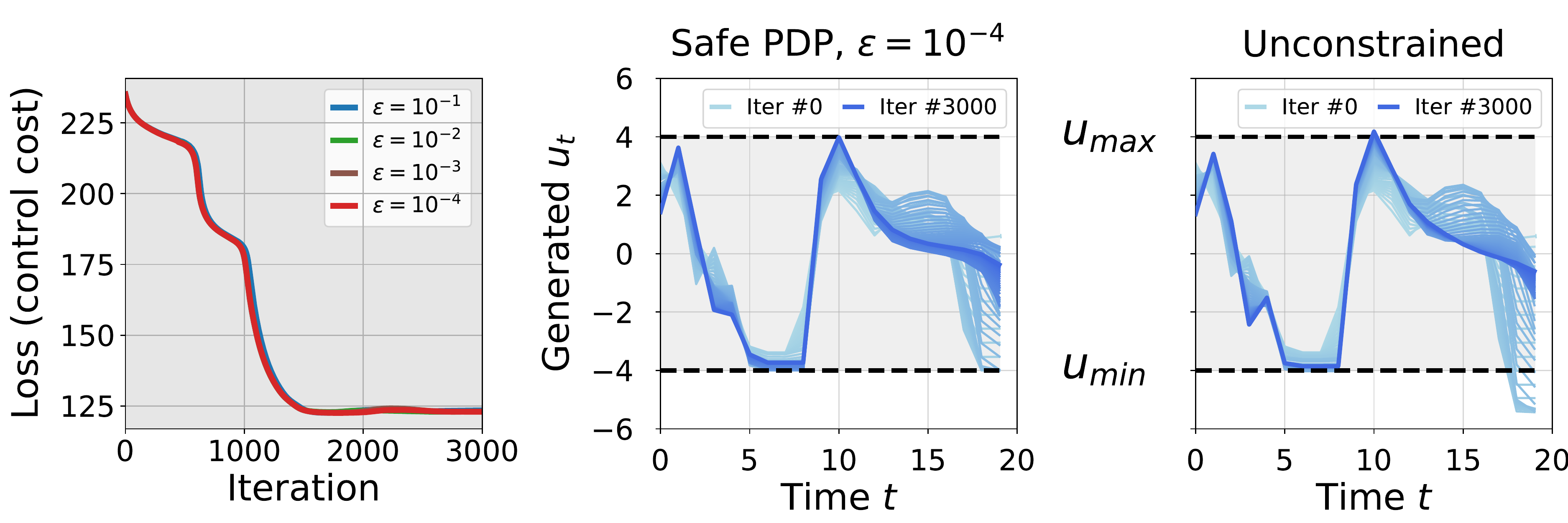}
	\caption{Safe neural policy optimization for cartpole. The first panel plots the loss (control cost) versus gradient-descent iteration under different outer-level barrier  parameter $\epsilon$; the second panel plots all intermediate control trajectories generated by the NN policy during the entire gradient-descent  iteration ($\epsilon=10^{-4}$); and the third panel plots all intermediate control trajectories  generated by the NN policy for the {unconstrained  policy optimization} under the same experimental conditions. The system constraints  are also marked using black dashed lines in the second and third panels.}
	\label{appendix-spo-cartpole}
\end{figure}

In our experiments, in order to make sure the initial NN policy is feasible (safe), we initialize the NN policy using supervised learning from a random demonstration trajectory (note that this demonstration trajectory does not have to be optimal but only to be feasible/safe).  For each system in Table \ref{experimenttable}, we apply  Safe PDP Algorithm \ref{alg_theorem3} to optimize the NN policy, and the complete experiment results are shown in  Fig. \ref{appendix-spo-cartpole}-\ref{appendix-spo-quadrotor}. More discussions about how to give a safe initialization are presented in Appendix \ref{appendix.discussion.limitation}

For each system, at a fixed barrier parameter ${\epsilon}$, we have applied the vanilla gradient descent to solve Problem \ref{equ_problem_penalty} with the step size (learning rate $\eta$ in Algorithm \ref{alg_theorem3})   set around $10^{-3}$. We plot the task loss (i.e., control cost) $\ell(\boldsymbol{\xi}_{\boldsymbol{\theta}},\boldsymbol{\theta})$ versus  iteration of the gradient descent in the first panel in Fig. \ref{appendix-spo-cartpole}-\ref{appendix-spo-quadrotor}, where we  only show the results for outer-level barrier parameter  $\epsilon$ taking from $10^{0}$ to $10^{-4}$ because  the NN policy has already achieved a good  convergence when $\epsilon\leq 10^{-2}$. As shown in the first panel in Fig. \ref{appendix-spo-cartpole}-\ref{appendix-spo-quadrotor}, for each system,   the policy  achieves a good convergence after a small number of iterations for each  $\epsilon$,  and obtains a good convergence after $\epsilon\leq 10^{-2}$.

\begin{figure}[h]
	\centering
	\includegraphics[width=0.7\linewidth]{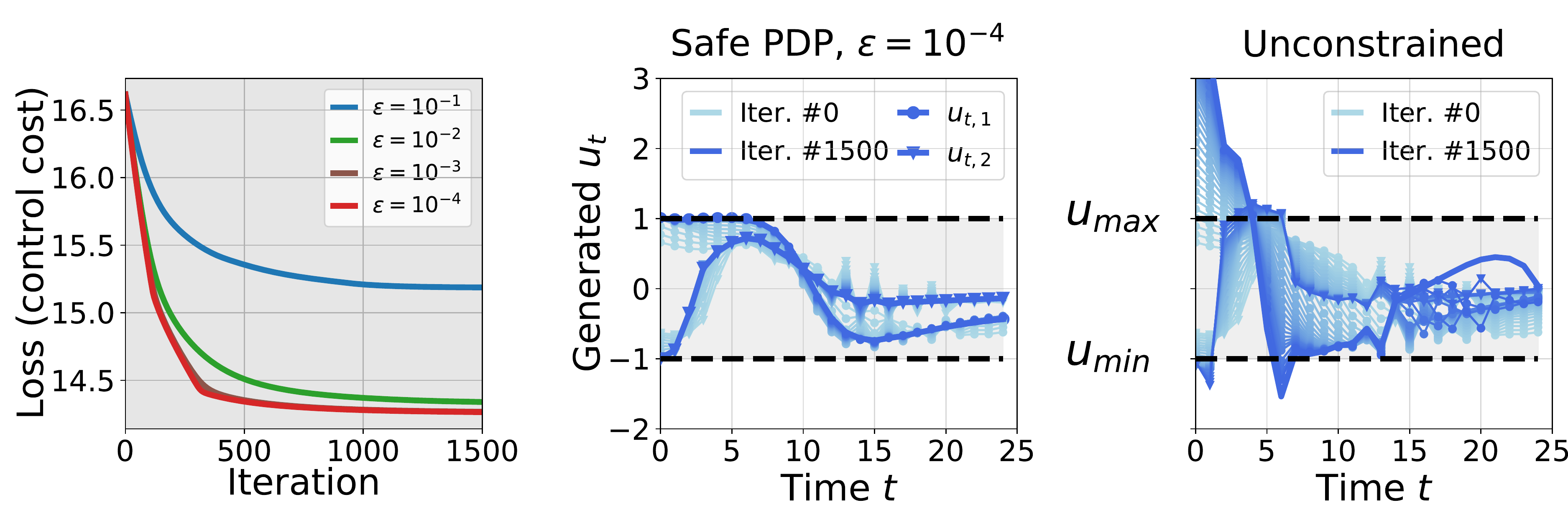}
	\caption{Safe neural  policy optimization for robot arm. The first panel plots the loss (control cost) versus gradient-descent iteration under different outer-level barrier  parameter $\epsilon$; the second panel plots all intermediate control trajectories generated by the NN policy during the entire gradient-descent  iteration ($\epsilon=10^{-4}$); and the third panel plots all intermediate control trajectories  generated by the NN policy for the {unconstrained  policy optimization} under the same experimental conditions. The system constraints  are also marked using black dashed lines in the second and third panels.}
	\label{appendix-spo-robotarm}
\end{figure}

\begin{figure}[h]
	\centering
	\includegraphics[width=0.7\linewidth]{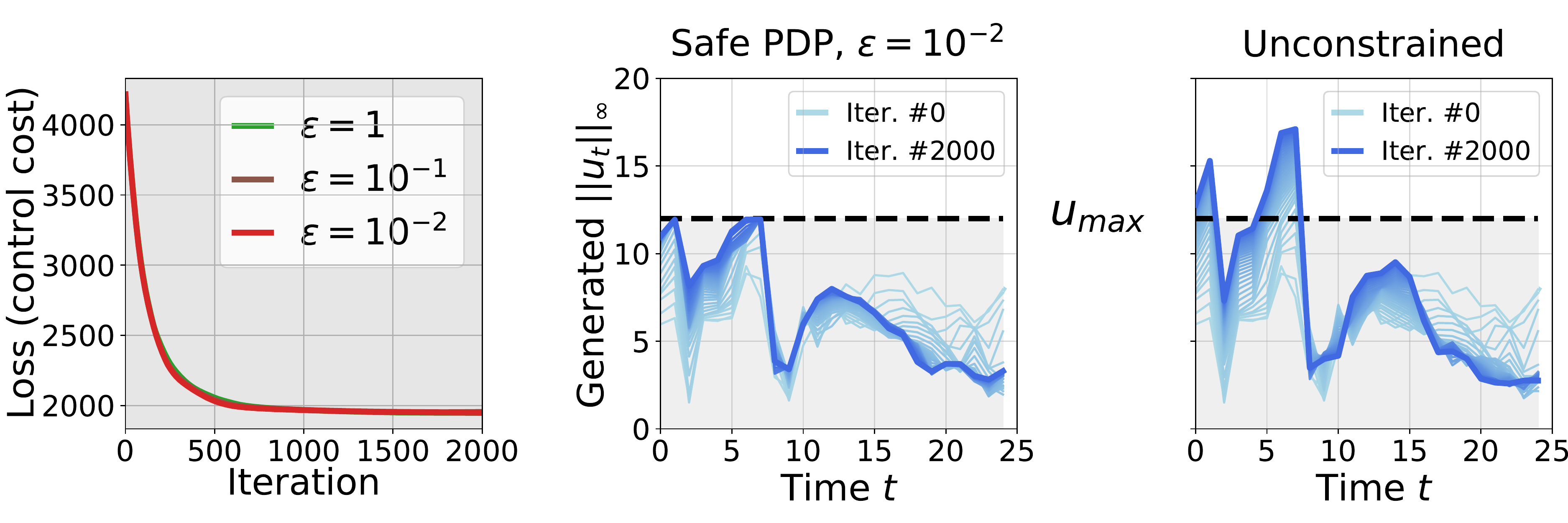}
	\caption{Safe neural policy optimization for 6-DoF maneuvering quadrotor.  The first panel plots the loss (control cost) versus gradient-descent iteration under different outer-level barrier  parameter $\epsilon$; the second panel plots all intermediate control trajectories generated by the NN policy during the entire gradient-descent  iteration ($\epsilon=10^{-2}$); and the third panel plots all intermediate control trajectories  generated by the NN policy for the {unconstrained  policy optimization} under the same experimental conditions. The system constraints  are also marked using black dashed lines in the second and third panels.}
	\label{appendix-spo-quadrotor}
\end{figure}

In order to show the constraint satisfaction  of  Safe PDP throughout the entire policy optimization process, in the second panel in Fig. \ref{appendix-spo-cartpole}-\ref{appendix-spo-quadrotor}, respectively, we  plot all  intermediate control trajectories generated from the NN policy throughout the  entire gradient-descent iteration of Safe PDP, as shown from  the light to  dark blue. From the second panel in Fig. \ref{appendix-spo-cartpole}-\ref{appendix-spo-quadrotor}, we note that throughout the optimization process, the NN policy is guaranteed safe, meaning that the generated trajectory will never violate the constraints. Under the same experimental conditions (NN configuration, policy initialization, learning rate), we also compare with the unconstrained policy optimization and plot its results in the third panel in Fig. \ref{appendix-spo-cartpole}-\ref{appendix-spo-quadrotor}, respectively. By comparing the  results between Safe PDP and unconstrained policy optimization,  we can confirm that  Safe PDP enables to achieve an optimal  policy while guaranteeing that any intermediate policy throughout optimization is safe, as asserted in Theorem \ref{theorem3}.

We have provided the  video demonstrations for the above safe policy optimization using Safe PDP; please visit   \url{https://youtu.be/sC81qc2ip8U}. The codes for all experiments here can be downloaded at \url{https://github.com/wanxinjin/Safe-PDP}.

\subsection{Safe Motion Planning}\label{appendix.application.splan}

In this experiment, we apply Safe PDP to solve the  safe motion planning problem for the systems  in Table \ref{experimenttable}. In  Problem \ref{equ_problem}, we  set the details of   $\boldsymbol{\Sigma}(\boldsymbol{\theta})$ as  follows,

\begin{longfbox}[padding-top=-4pt,margin-top=-6pt, padding-bottom=-3pt, margin-bottom=-10pt]
	\mathleft
	\begin{equation}\label{sys.opt2}
	\boldsymbol{\Sigma}(\boldsymbol{\theta}):\quad\qquad\qquad\qquad
	\begin{aligned}
	\emph{dynamics:} &\quad \boldsymbol{x}_{t+1}=\boldsymbol{f}(\boldsymbol{x}_{t},\boldsymbol{u}_{t}) \quad \text{with}\quad \boldsymbol{x}_{0}, \\
	\emph{control input:}& \quad
	\boldsymbol{u}_t=\boldsymbol{u}(t, \boldsymbol{\theta}),
	\end{aligned}
	\end{equation}	\mathcenter
\end{longfbox}
where the dynamics $\boldsymbol{f}$ is learned from demonstrations in Section \ref{appendix.application.mpc}, and we parameterize the control input function $\boldsymbol{u}_t=\boldsymbol{u}(t, \boldsymbol{\theta})$ using the Lagrangian polynomial \cite{abramowitz1964handbook} as follows,
\begin{equation}\label{polypolicy1}
\boldsymbol{u}(t, \boldsymbol{\theta})=\sum_{i=0}^{N}\boldsymbol{u}_{i}b_{i}(t) \quad \quad \text{with} \quad\quad b_{i}(t)=\prod_{\substack{0\leq j \leq N,  j \neq i}}\frac{t-t_j}{t_i-t_j}.
\end{equation}
Here, $b_{i}(t)$ is called  Lagrange basis, and the policy parameter $ \boldsymbol{\theta} $ is defined as
\begin{equation}
\boldsymbol{\theta}=[\boldsymbol{u}_0,\cdots,\boldsymbol{u}_N]^\prime\in\mathbb{R}^{m(N+1)},
\end{equation}
which is the vector of the pivots of the Lagrange polynomial.  The benefit of the above parameterization is that  the trajectory of system states, which results from integrating (\ref{sys.opt2}) given the input polynomial trajectory $\boldsymbol{u}_t=\boldsymbol{u}(t, \boldsymbol{\theta})$, is inherently smooth and dynamics-feasible. In our experiments, the  degree $N$ of the Lagrange polynomial is set as $N=10$. 
Also in Problem \ref{equ_problem}, we  set the task/planning loss  $\ell(\boldsymbol{\xi}_{\boldsymbol{\theta}},\boldsymbol{\theta})$ as the  control cost $J(\boldsymbol{\theta}_{\text{obj}})$, and set the task constraints $R_i(\boldsymbol{\xi}_{\boldsymbol{\theta}},\boldsymbol{\theta})$ as the system constraints  $\boldsymbol{g}(\boldsymbol{\theta}_{\text{cstr}})$, both given  in Table \ref{experimenttable} with $\boldsymbol{\theta}_{\text{obj}}$ and $\boldsymbol{\theta}_{\text{cstr}}$ known.

Since the system  $\boldsymbol{\Sigma}(\boldsymbol{\theta})$ in  (\ref{sys.opt2})  now does not include the control cost  $J$ anymore,  solving Problem \ref{equ_traj} for $\boldsymbol{\xi}_{\boldsymbol{\theta}}$ becomes a simple integration of (\ref{sys.opt2}) from $t=0$ to $T{-}1$, and the auxiliary control system $\boldsymbol{\overline{\Sigma}}(\boldsymbol{\xi}_{\boldsymbol{\theta}})$ in (\ref{equ_aux}) to compute  $ \frac{\partial\boldsymbol{\xi}_{\boldsymbol{\theta}}}{ \partial\boldsymbol{\theta}}$ is simplified to a feedback control system \cite{jin2019pontryagin} below:

\begin{longfbox}[padding-top=-3pt, margin-top=-5pt, padding-bottom=-0pt, margin-bottom=6pt]
	\mathleft
	\begin{equation}\label{planningcontrolback}
	\boldsymbol{\overline\Sigma}(\boldsymbol{\xi}_{\boldsymbol{\theta}}):\qquad\qquad
	\begin{aligned}
	\text{dynamics:}&\quad {{X}}_{t+1}^{\boldsymbol{{\theta}}}=
	F_t^{x}{{X}}^{\boldsymbol{{\theta}}}_{t}+F_t^{u}{{U}}^{\boldsymbol{\theta}}_{t} \quad \text{with} \quad {{X}}_{0}=\boldsymbol{0},\\
	\text{control input:} &\quad \quad{{U}}^{\boldsymbol{\theta}}_{t}=U_t^{{e}},
	\end{aligned}
	\end{equation} 
	\mathcenter
\end{longfbox} 
where
${{U}}_t^e=\frac{\partial \boldsymbol{\pi}_t}{\partial \boldsymbol{\theta}}$.
Integrating  (\ref{planningcontrolback}) from $t=0$ to $T-1$ leads to
$
\{{{X}}_{0:T}^{\boldsymbol{\theta}},{{U}}_{0:T-1}^{\boldsymbol{\theta}}\}= \frac{\partial \boldsymbol{{\xi}}_{\boldsymbol{\theta}}}{\partial \boldsymbol{\theta}}.
$

\begin{figure} [h]
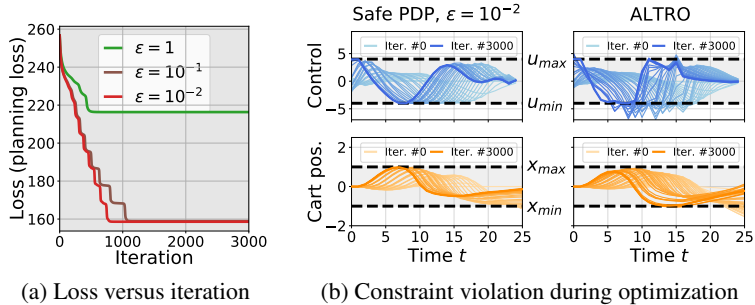

	\centering
\begin{subfigure}{.25\textwidth}
	\centering
	\includegraphics[width=\linewidth]{splan_cartpole_loss.pdf}
	\caption{Loss versus iteration}
	\label{appendix-splan-cartpole.1}
\end{subfigure}
\hspace{10pt}
\begin{subfigure}{.425\textwidth}
	\centering
	\includegraphics[width=\linewidth]{splan_cartpole_compare.pdf}
	\caption{Constraint violation during optimization}
	\label{appendix-splan-cartpole.2}
\end{subfigure} 
\caption{Safe motion planning for cartpole. (a)  plots the loss (i.e., planning loss) versus gradient-descent iteration under different outer-level barrier  parameter $\epsilon$. The left figure in (b) shows all  intermediate trajectories during the entire iteration of Safe PDP ($\epsilon=10^{-2}$), and the right figure in (b) shows all  intermediate trajectories during the entire iteration of the ALTRO algorithm \cite{howell2019altro}. The state and control constraints are also marked in (b).}
\label{appendix-splan-cartpole}
\end{figure}

\begin{figure} [h]
	\centering
\begin{subfigure}{.25\textwidth}
	\centering
	\includegraphics[width=\linewidth]{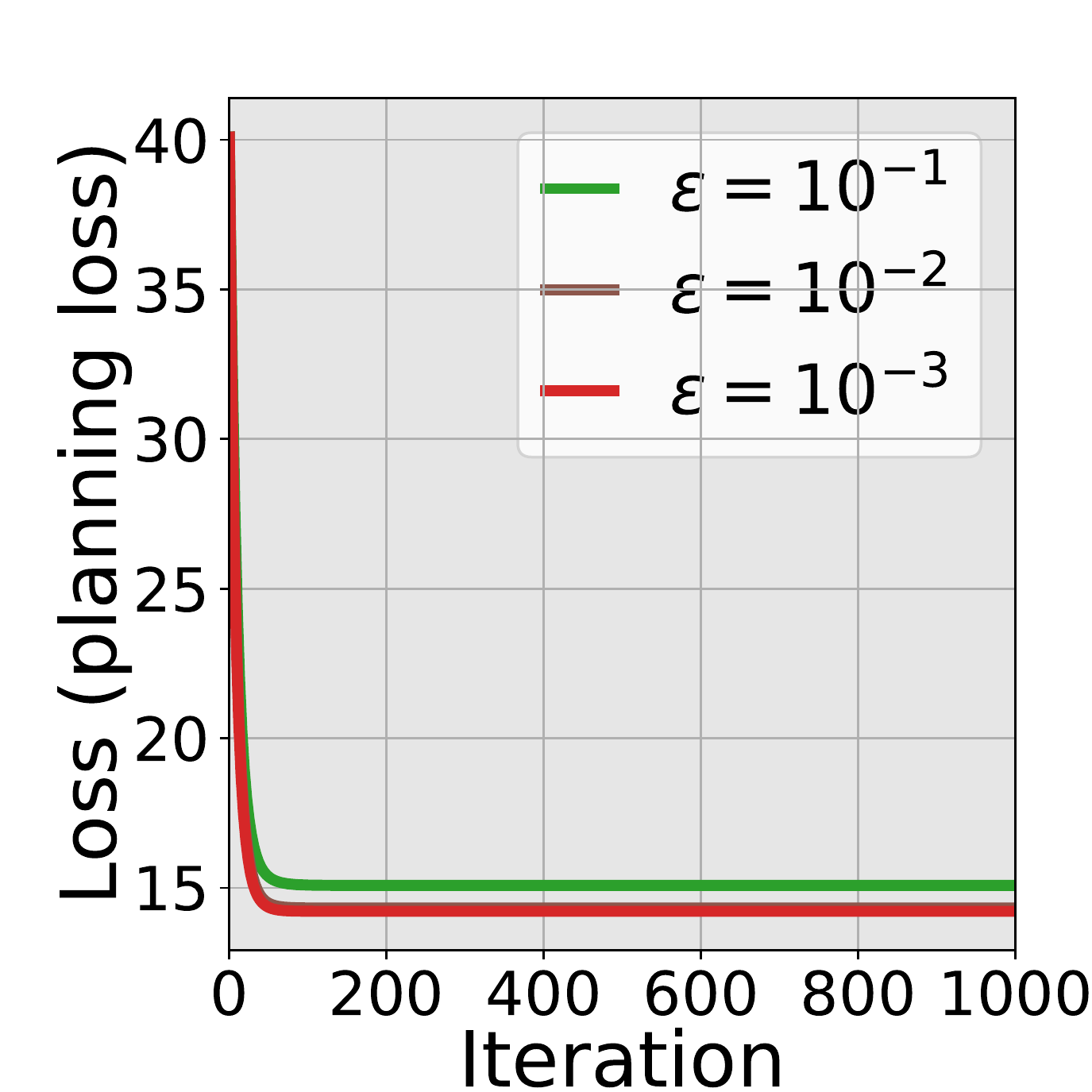}
	\caption{Loss versus iteration}
	\label{appendix-splan-robotarm.1}
\end{subfigure}
\hspace{10pt}
\begin{subfigure}{.425\textwidth}
	\centering
	\includegraphics[width=\linewidth]{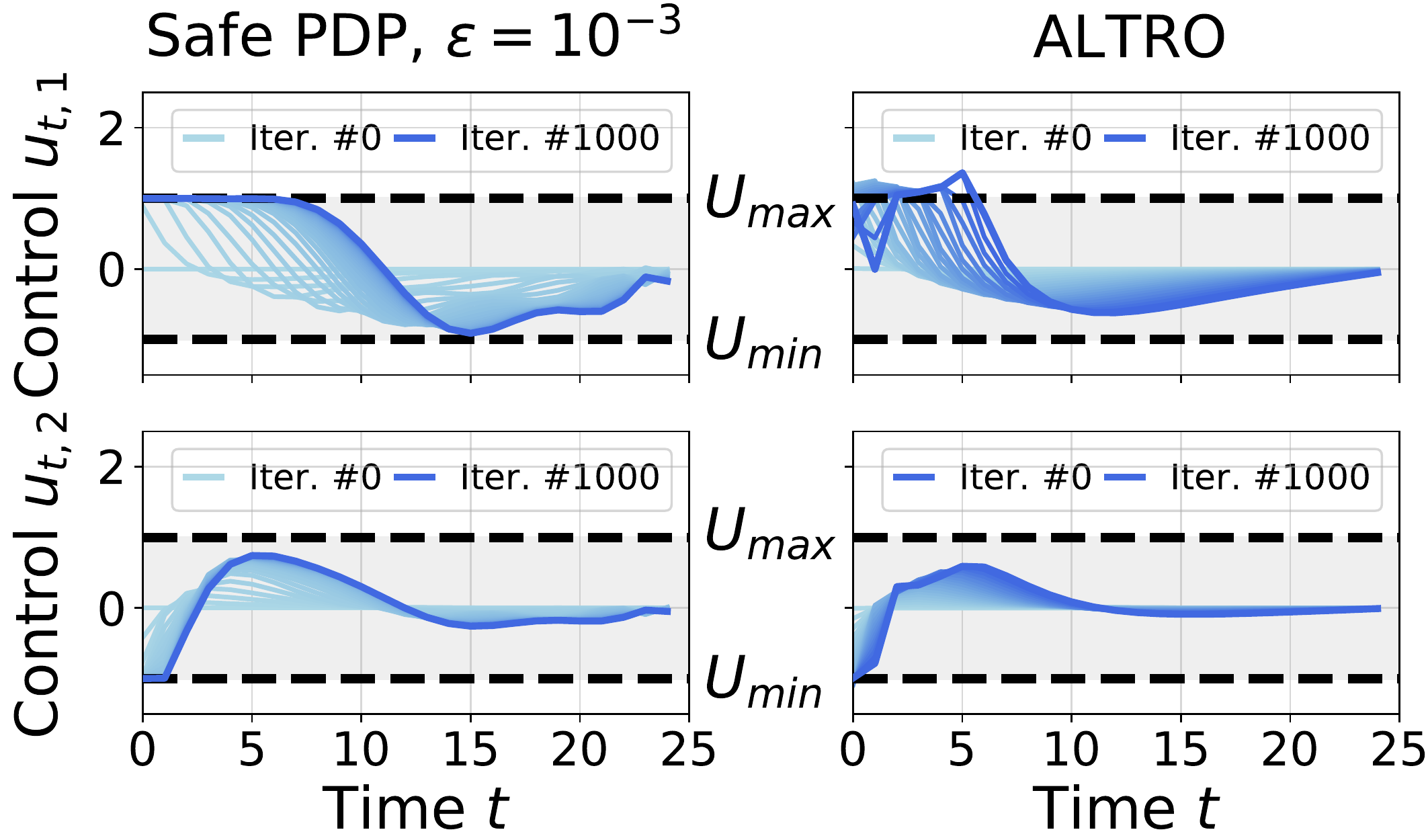}
	\caption{Constraint violation during optimization}
	\label{appendix-splan-robotarm.2}
\end{subfigure} 
\caption{Safe motion planning for robot arm. (a)  plots the loss (i.e., planning loss) versus gradient-descent iteration under different outer-level barrier  parameter $\epsilon$. The left figure in (b) shows all  intermediate trajectories during the entire iteration of Safe PDP ($\epsilon=10^{-3}$), and the right figure in (b) shows all  intermediate trajectories during the entire iteration of the ALTRO algorithm \cite{howell2019altro}. The  control constraints are also marked in (b).}
\label{appendix-splan-robotarm}
\end{figure}

\begin{figure} [h]
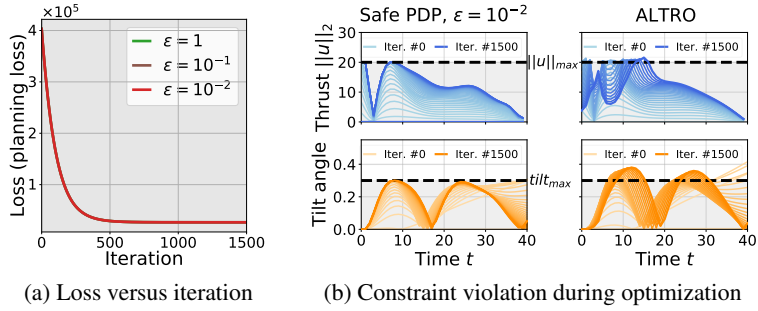

	\centering
	\begin{subfigure}{.25\textwidth}
		\centering
		\includegraphics[width=\linewidth]{splan_rocket_loss.pdf}
		\caption{Loss versus iteration}
		\label{appendix-splan-rocket.1}
	\end{subfigure}
	\hspace{10pt}
	\begin{subfigure}{.425\textwidth}
		\centering
		\includegraphics[width=\linewidth]{splan_rocket_compare.pdf}
		\caption{Constraint violation during optimization}
		\label{appendix-splan-rocket.2}
	\end{subfigure} 
\caption{Safe motion planning for 6-DoF rocket powered landing. (a)  plots the loss (i.e., planning loss) versus gradient-descent iteration under different outer-level barrier  parameter $\epsilon$.  The left figure in (b) shows all  intermediate trajectories during the entire iteration of Safe PDP ($\epsilon=10^{-2}$), and the right figure in (b) shows all  intermediate trajectories during the entire iteration of the ALTRO algorithm \cite{howell2019altro}. The state and control constraints are also marked in (b).}
	\label{appendix-splan-rocket}
\end{figure}

For each system in Table \ref{experimenttable}, we apply  Safe PDP Algorithm \ref{alg_theorem3} to perform safe motion planning, and the complete experiment results are shown in  Fig. \ref{appendix-splan-cartpole}-\ref{appendix-splan-rocket}.  For each system, at a fixed outer-level barrier parameter ${\epsilon}$, we have applied the vanilla gradient descent to solve Problem \ref{equ_problem_penalty} with the step size  (learning rate $\eta$ in Algorithm \ref{alg_theorem3})  set to  $10^{-2}$ or $10^{-1}$. We plot  the planning loss $\ell(\boldsymbol{\xi}_{\boldsymbol{\theta}},\boldsymbol{\theta})$ versus  gradient descent iteration  in Fig.  \ref{appendix-splan-cartpole.1}-\ref{appendix-splan-rocket.1}, respectively; here  we only show the results for  $\epsilon$ taking from $10^{0}$ to $10^{-2}$ because the trajectory has already achieved a good convergence when $\epsilon\leq 10^{-2}$. As shown in Fig. \ref{appendix-splan-cartpole.1}-\ref{appendix-splan-rocket.1}, for each system,  the trajectory  achieves a good convergence after a small number of iterations given a fixed  $\epsilon$,  and obtains a good convergence after $\epsilon\leq 10^{-2}$.

To demonstrate that Safe PDP can guarantee  safety throughout the optimization process, we plot all  intermediate trajectories during the entire iteration of Safe PDP in \ref{appendix-splan-cartpole.2}-\ref{appendix-splan-rocket.2}. At the same time, we also show the results of the  ALTRO method \cite{howell2019altro}, which is a state-of-the-art method for  constrained trajectory optimization. 
By comparing the results in Fig. \ref{appendix-splan-cartpole.2}-\ref{appendix-splan-rocket.2},  we can observe that Safe PDP enables to find the optimal trajectory while guaranteeing  strict constraint satisfaction throughout the entire optimization process; while for ALTRO, although the trajectory satisfies  the constraints at convergence, the intermediate trajectories during  optimization may violate the constraints, making it not suitable to handle  safety-critical motion planning tasks.

We have provided the videos for the above safe motion planning using Safe PDP. Please visit  the link  \url{https://youtu.be/vZVxgo30mDs}. The codes for all   experiments here can be downloaded at \url{https://github.com/wanxinjin/Safe-PDP}.

\subsection{Learning MPCs from Demonstrations}\label{appendix.application.mpc}

In this experiment, we apply  Safe PDP to learn dynamics $\boldsymbol{f}$, constraints $\boldsymbol{g}$, or/and control cost $J$ for the systems in Table \ref{experimenttable} from demonstration data. This type of problems has been extensively studied in system identification \cite{johansson1993system} (neural ODEs \cite{chen2018neural}), inverse optimal control  (inverse reinforcement learning) \cite{jin2021inverse,jin2021distributed,jin2019inverse}, and learning from demonstrations \cite{jin2020correction,jin2020sparse}. However, existing methods have the following two technical gaps; first, existing methods are typically developed without considering  constraints; second, there are  rarely the methods that are capable to \emph{jointly} learn  dynamics, state-input constraints, and control cost for continuous control systems.
In this part, we will show that the above technical gaps can be addressed by  Safe PDP. Throughout this part, we define the task loss  in Problem \ref{equ_problem} as the \emph{reproducing loss} as below
\begin{equation}\label{loss_lfd}
\ell(\boldsymbol{\xi}_{\boldsymbol{\theta}}, \boldsymbol{\theta})= \norm{\boldsymbol{\xi}^\text{demo}- \boldsymbol{\xi}_{\boldsymbol{\theta}}}_2^2,
\end{equation} 
which is to penalize the distance between the reproduced trajectory $\boldsymbol{\xi}_{\boldsymbol{\theta}}$ from the learnable model $\boldsymbol{\Sigma}(\boldsymbol{\theta})$  and the given demonstrations $\boldsymbol{\xi}^{\text{demo}}$, and there is no task constraint. For  $\boldsymbol{\Sigma}(\boldsymbol{\theta})$ in Problem \ref{equ_problem}, only the unknown parts (dynamics, control cost, or/and constraints) are parameterized by $\boldsymbol{\theta}$. Thus, by solving Problem \ref{equ_problem}, we are able to learn $\boldsymbol{\Sigma}(\boldsymbol{\theta})$ such that its  trajectory $\boldsymbol{\xi}_{\boldsymbol{\theta}}$ has closest distance to the given demonstrations $\boldsymbol{\xi}^\text{demo}
$.

In our experiment, when dealing with $\boldsymbol{\xi}_{\boldsymbol{\theta}}$  and $\frac{\partial \boldsymbol{\xi}_{\boldsymbol{\theta}}}{\partial \boldsymbol{\theta}}$ for  $\boldsymbol{\Sigma}(\boldsymbol{\theta})$, we use the following three strategies. 

\begin{itemize}
	\item \emph{Strategy (A)}: use an optimal control solver \cite{andersson2019casadi} to solve the constrained optimal control $\boldsymbol{\Sigma}(\boldsymbol{\theta})$ in Problem \ref{equ_traj} to obtain  $\boldsymbol{\xi}_{\boldsymbol{\theta}}$, and use Theorem \ref{theorem1} (i.e., Algorithm \ref{alg_theorem1}) to obtain the trajectory derivative $\frac{\partial \boldsymbol{\xi}_{\boldsymbol{\theta}}}{\partial \boldsymbol{\theta}}$  by solving $\boldsymbol{\overline\Sigma}(\boldsymbol{\xi}_{\boldsymbol{\theta}})$ in (\ref{equ_aux}).

	\item\emph{Strategy (B)}: by applying  Theorem \ref{alg_theorem2} (i.e., Algorithm \ref{alg_theorem2}),  approximate $\boldsymbol{\xi}_{\boldsymbol{\theta}}$ and $\frac{\partial \boldsymbol{\xi}_{\theta}}{\partial \boldsymbol{\theta}}$  using  $\boldsymbol{\xi}_{(\boldsymbol{\theta},\gamma)}$ and $\frac{\partial \boldsymbol{\xi}_{(\boldsymbol{\theta},\gamma)}}{\partial \boldsymbol{\theta}}$, respectively, with a choice of  small barrier parameter $\gamma>0$.
	
	\item \emph{Strategy (C)}:  obtain $\boldsymbol{\xi}_{\boldsymbol{\theta}}$  by solving $\boldsymbol{\Sigma}(\boldsymbol{\theta})$ in Problem \ref{equ_traj} via a  solver \cite{andersson2019casadi}, and apply Theorem \ref{theorem2} (i.e., Algorithm \ref{alg_theorem2}) only to approximate   $\frac{\partial\boldsymbol{\xi}_{\boldsymbol{\theta}}}{\partial \boldsymbol{\theta}}$ using $\frac{\partial \boldsymbol{\xi}_{(\boldsymbol{\theta},\gamma)}}{\partial \boldsymbol{\theta}}$. 
\end{itemize}

In the following experiments, when using Algorithm \ref{alg_theorem2}, we choose  $\gamma=10^{-2}$ because the corresponding inner-level approximations $\boldsymbol{\xi}_{(\boldsymbol{\theta},\gamma)}$ and $\frac{\partial \boldsymbol{\xi}_{(\boldsymbol{\theta},\gamma)}}{\partial \boldsymbol{\theta}}$ already achieve a good accuracy,  as shown in previous experiments. In practice, the choice of  $\gamma>0$ is very flexible  depending on the desired accuracy (a smaller $\gamma$ never hurts but would decrease the computational efficiency, as discussed in Appendix \ref{appendix.discussion.tradeoff}).

\subsubsection{Learning Constrained ODEs from Demonstrations}

In the first experiment,  consider that in $\boldsymbol{\Sigma}(\boldsymbol{\theta})$ the control cost  $J$ is known  while the dynamics (Ordinary Difference Equation) $\boldsymbol{f}(\boldsymbol{\theta}_{\text{dyn}})$ and constraints $\boldsymbol{g}_t(\boldsymbol{\theta}_{\text{cstr}})$ are unknown and parameterized, as in Table \ref{experimenttable}, $\boldsymbol{\theta}{=}\{\boldsymbol{\theta}_{\text{dyn}}, \boldsymbol{\theta}_{\text{cstr}}\}$. We aim to learn $\boldsymbol{\theta}$  from  given demonstrations $\boldsymbol{\xi}^\text{demo}$ by solving Problem \ref{equ_problem}. Here, the demonstrations are generated by simulating the true system (i.e., expert) with $\boldsymbol{\theta}$ known; the demonstrations  for each system contain  two episode  trajectories with time horizon around $T=50$.

\begin{figure} [h]
	\begin{subfigure}{.245\textwidth}
		\centering
		\includegraphics[width=\linewidth]{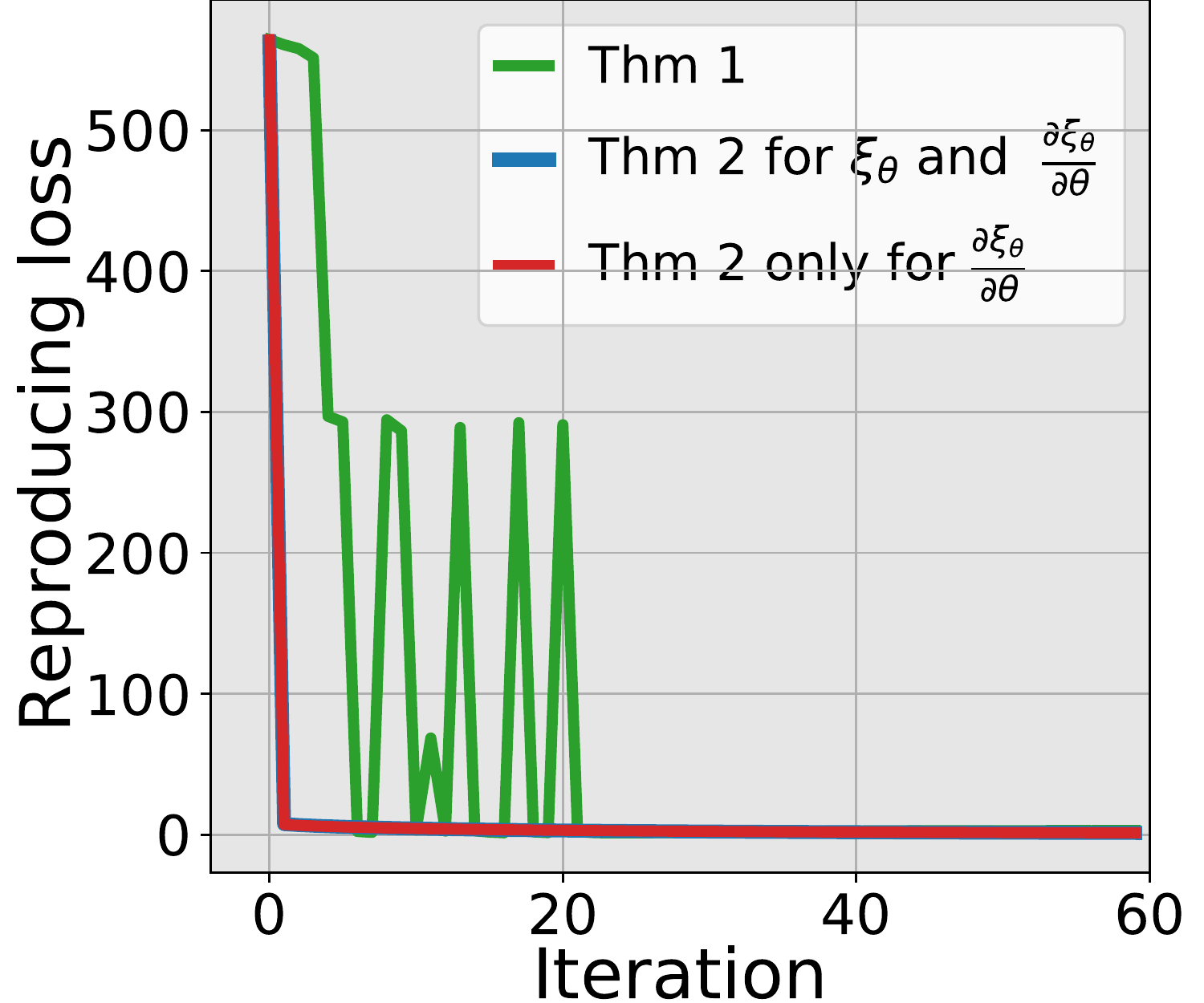}
		\caption{Cartpole}
		\label{figscode.1}
	\end{subfigure}
	\begin{subfigure}{.245\textwidth}
		\centering
		\includegraphics[width=\linewidth]{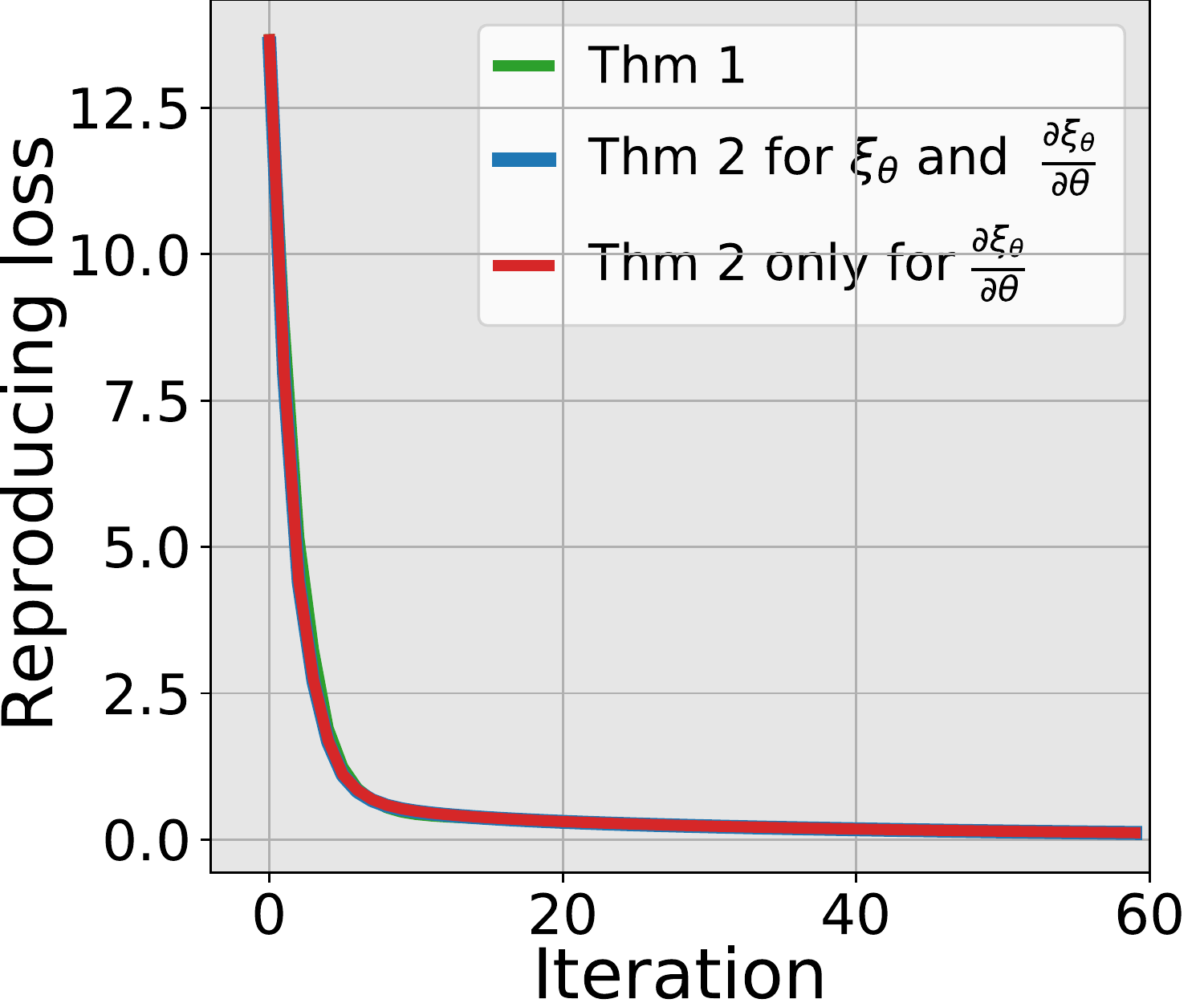}
		\caption{Robot arm }
		\label{figscode.2}
	\end{subfigure}
	\begin{subfigure}{.245\textwidth}
		\centering
		\includegraphics[width=\linewidth]{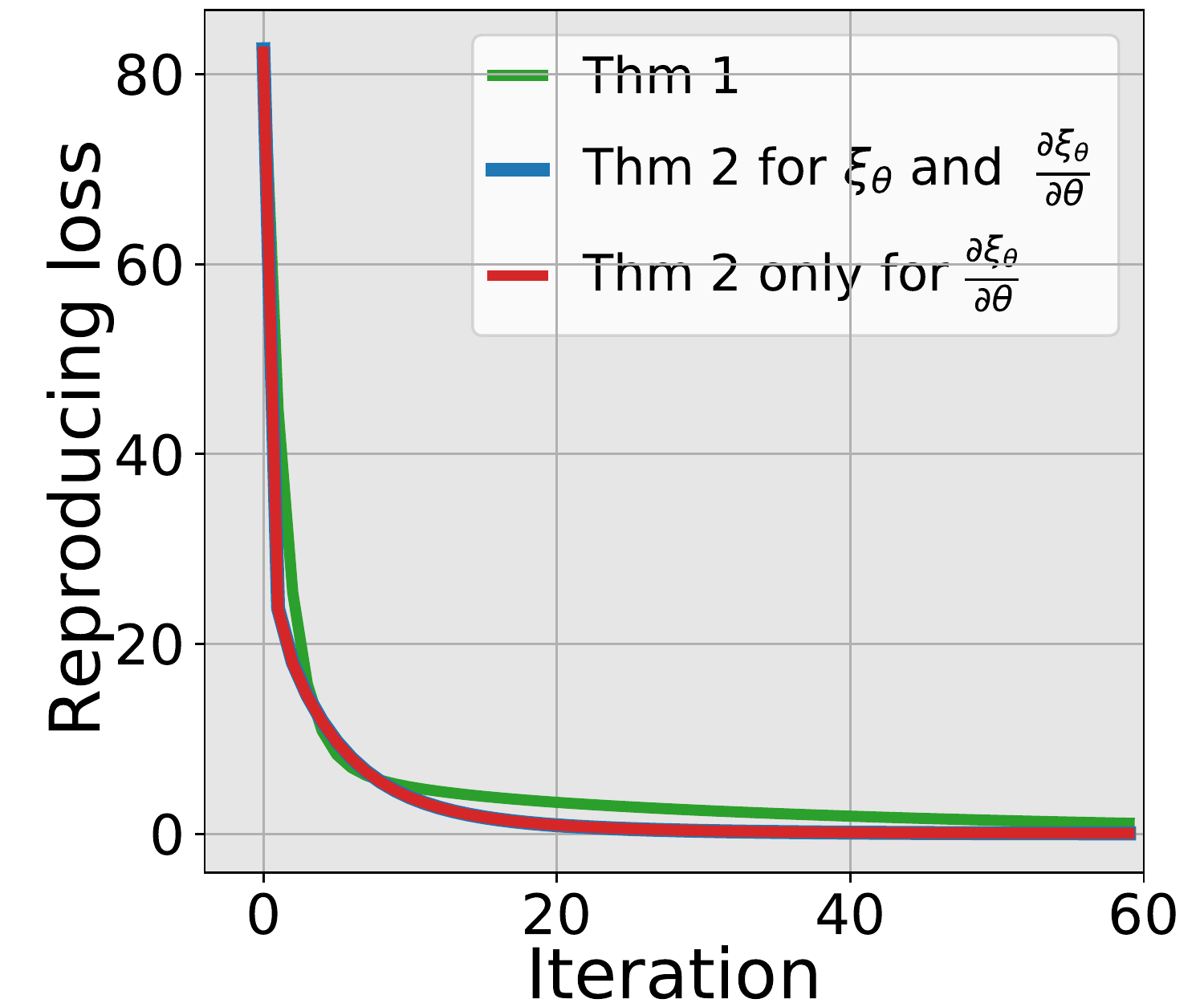}
		\caption{Quadrotor}
		\label{figscode.3}
	\end{subfigure}
	\begin{subfigure}{.245\textwidth}
		\centering
		\includegraphics[width=\linewidth]{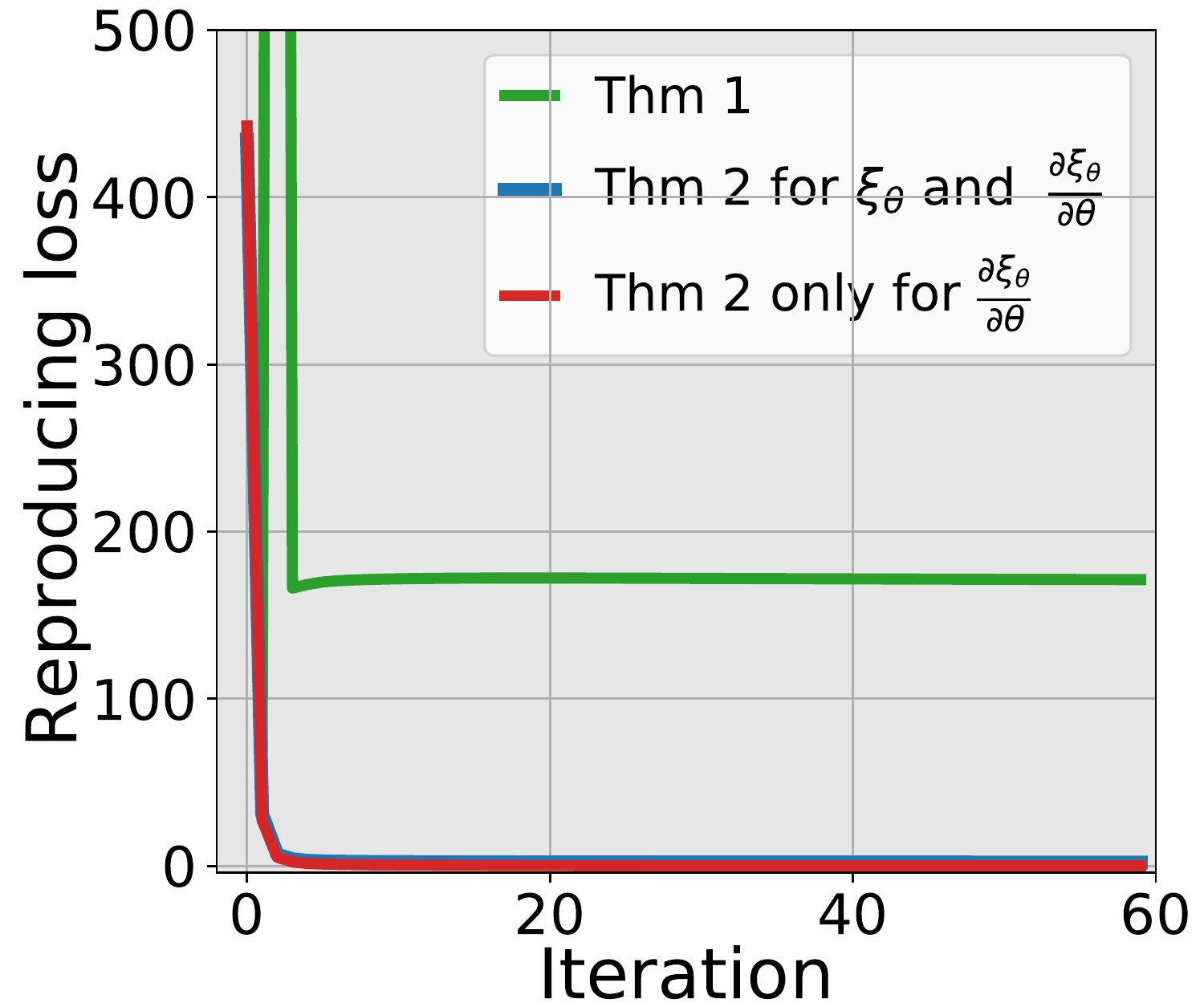}
		\caption{Rocket}
		\label{figscode.4}
	\end{subfigure}
	\caption{Learning both dynamics and constraints from demonstrations.}
	\label{figscode}
\end{figure}

To solve Problem \ref{equ_problem}, since there are no task constraints, we use the vanilla gradient descent to minimize the reproducing loss  (\ref{loss_lfd}) while using  the three strategies as mentioned above to handle the lower-level Problem \ref{equ_traj}. The initial condition for the gradient descent is given randomly, and the learning rate  for the gradient descent is set as $10^{-5}$.
The complete results for all systems in Table \ref{experimenttable} are given in Fig. \ref{figscode}.

Fig. \ref{figscode.1}-\ref{figscode.4} plot the reproducing loss (\ref{loss_lfd}) versus gradient-descent iteration. The results show that for  Strategies (B) and (C) (in blue and red, respectively), the reproducing loss (\ref{loss_lfd}) is quickly covering to zeros, indicating that the  dynamics and constraints are successfully learned to reproduce the demonstrations. However, we also note that Strategy (A) (in green) suffers from some numerical instability, and this will be discussed later.

\subsubsection{Jointly Learning Dynamics, Constraints, and Control Cost  from Demonstrations}

In the second experiment, suppose  in all systems in Table \ref{experimenttable},   the control cost $J(\boldsymbol{\theta}_{\text{cost}})$,   dynamics $\boldsymbol{f}(\boldsymbol{\theta}_{\text{dyn}})$, and state and input constraints $\boldsymbol{g}_t(\boldsymbol{\theta}_{\text{cstr}})$ are all  unknown and parameterized as in Table \ref{experimenttable}. We aim to jointly learn $\boldsymbol{\theta}=\{\boldsymbol{\theta}_{\text{cost}}, \boldsymbol{\theta}_{\text{dyn}}, \boldsymbol{\theta}_{\text{cstr}}\}$  from  given demonstrations $\boldsymbol{\xi}^\text{demo}$ by solving Problem \ref{equ_problem}.  Here, the demonstrations are generated by simulating the system (i.e., expert) with $\boldsymbol{\theta}$ known, the demonstrations for each system contain two episode  trajectories for each system with time horizon around $T=50$.

\begin{figure} [h]
	\begin{subfigure}{.245\textwidth}
		\centering
		\includegraphics[width=\linewidth]{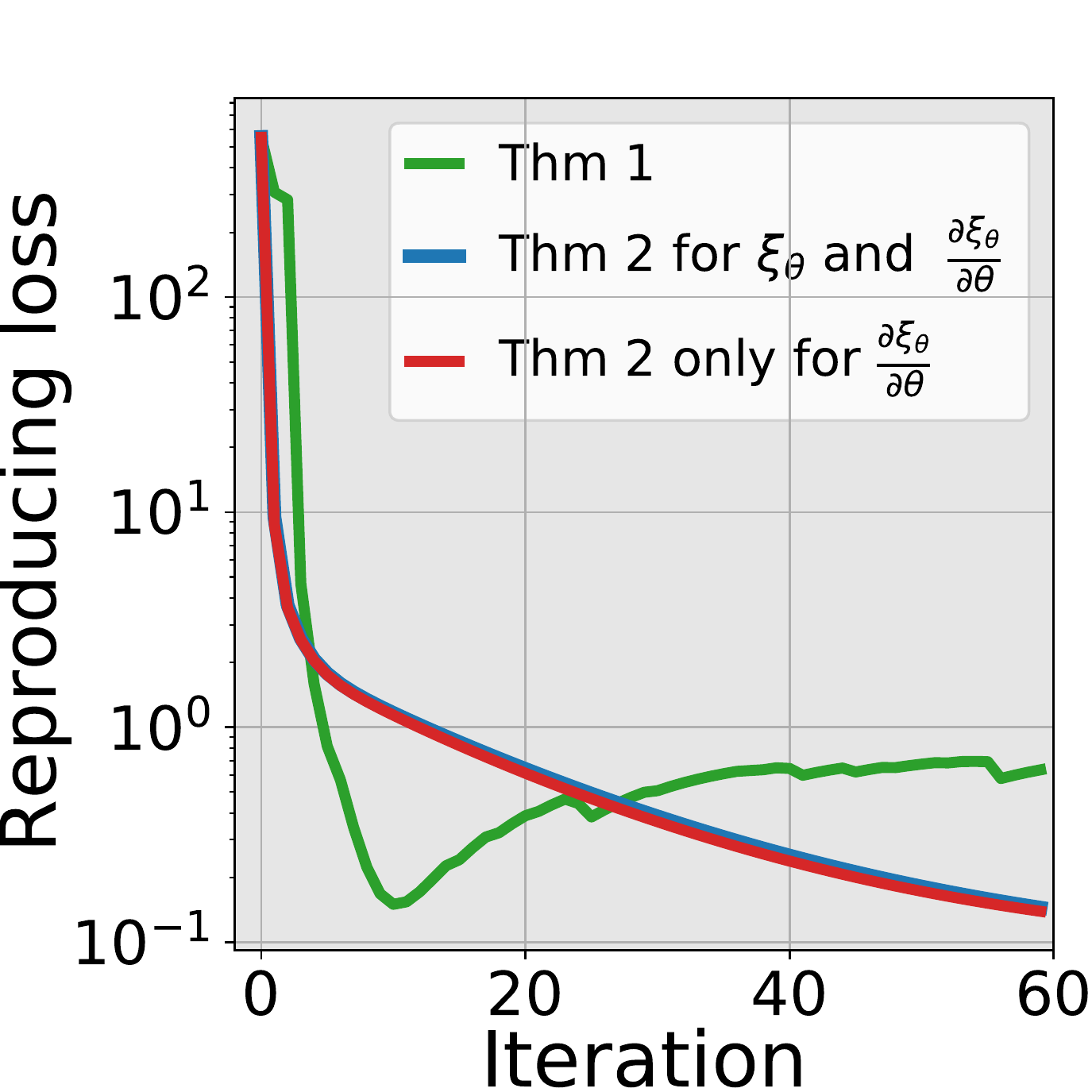}
		\caption{Cartpole (log-y)}
		\label{figsioc2.1}
	\end{subfigure}
	\begin{subfigure}{.245\textwidth}
		\centering
		\includegraphics[width=\linewidth]{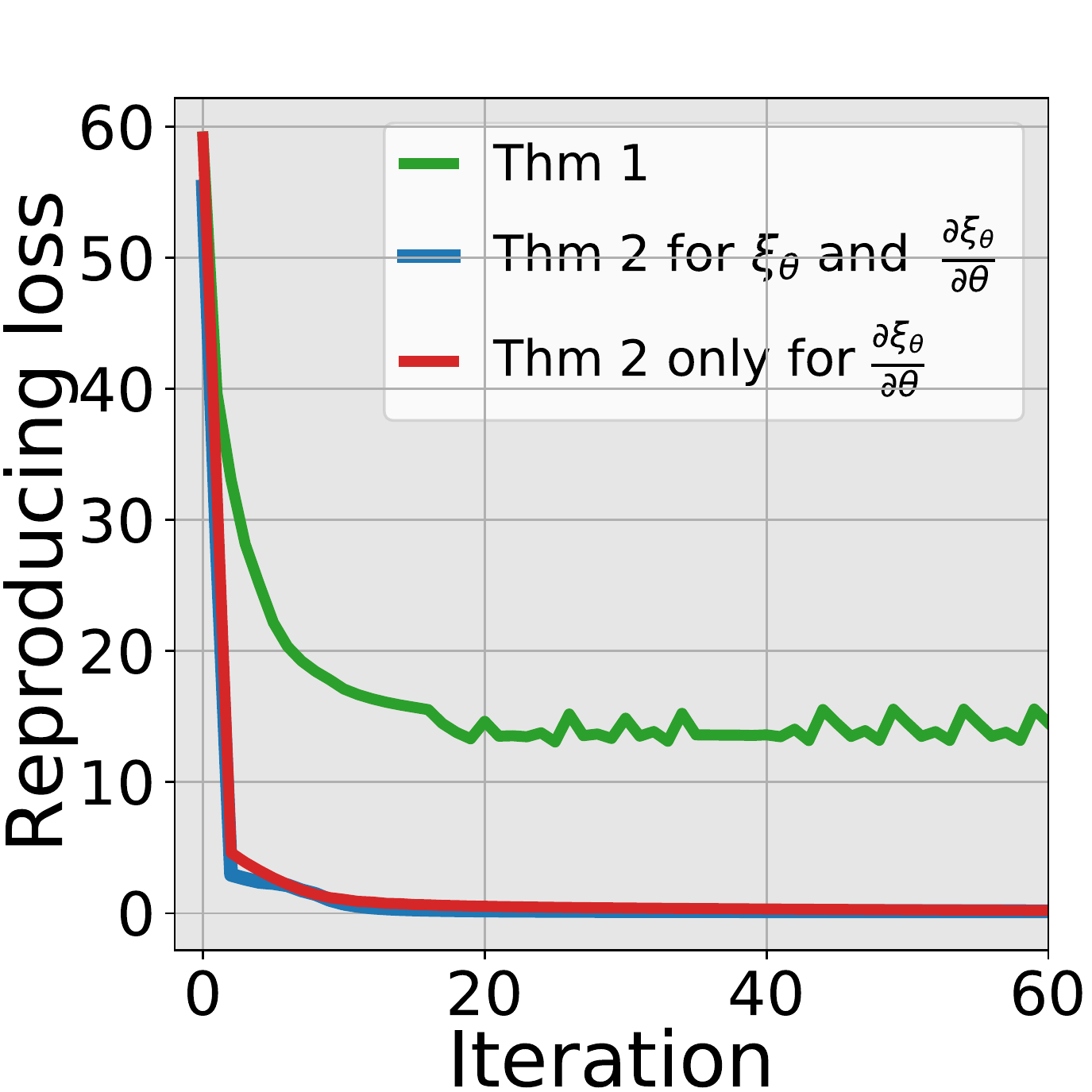}
		\caption{Robot arm}
		\label{figsioc2.2}
	\end{subfigure}
	\begin{subfigure}{.245\textwidth}
		\centering
		\includegraphics[width=\linewidth]{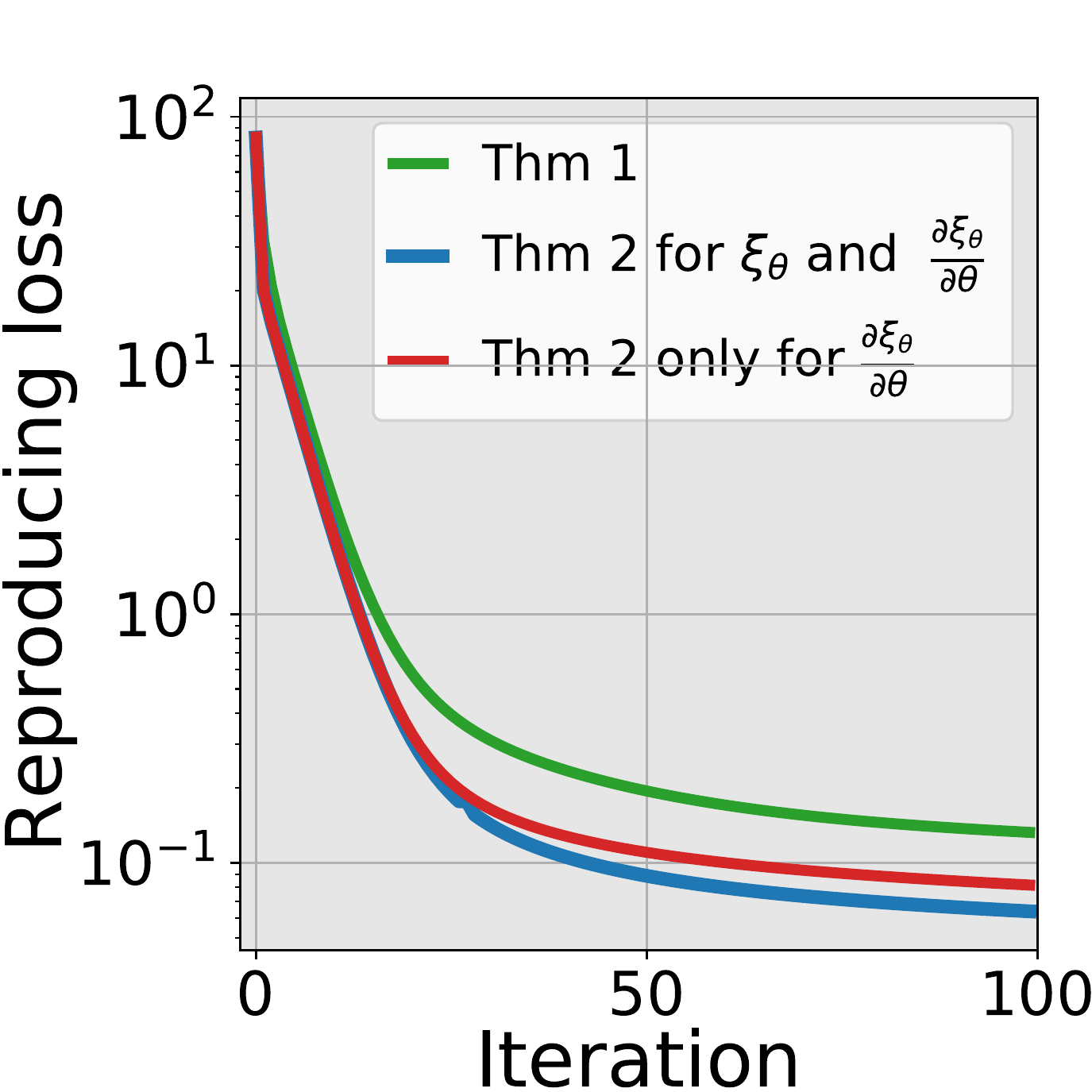}
		\caption{Quadrotor (log-y)}
		\label{figsioc2.3}
	\end{subfigure}
	\begin{subfigure}{.245\textwidth}
		\centering
		\includegraphics[width=\linewidth]{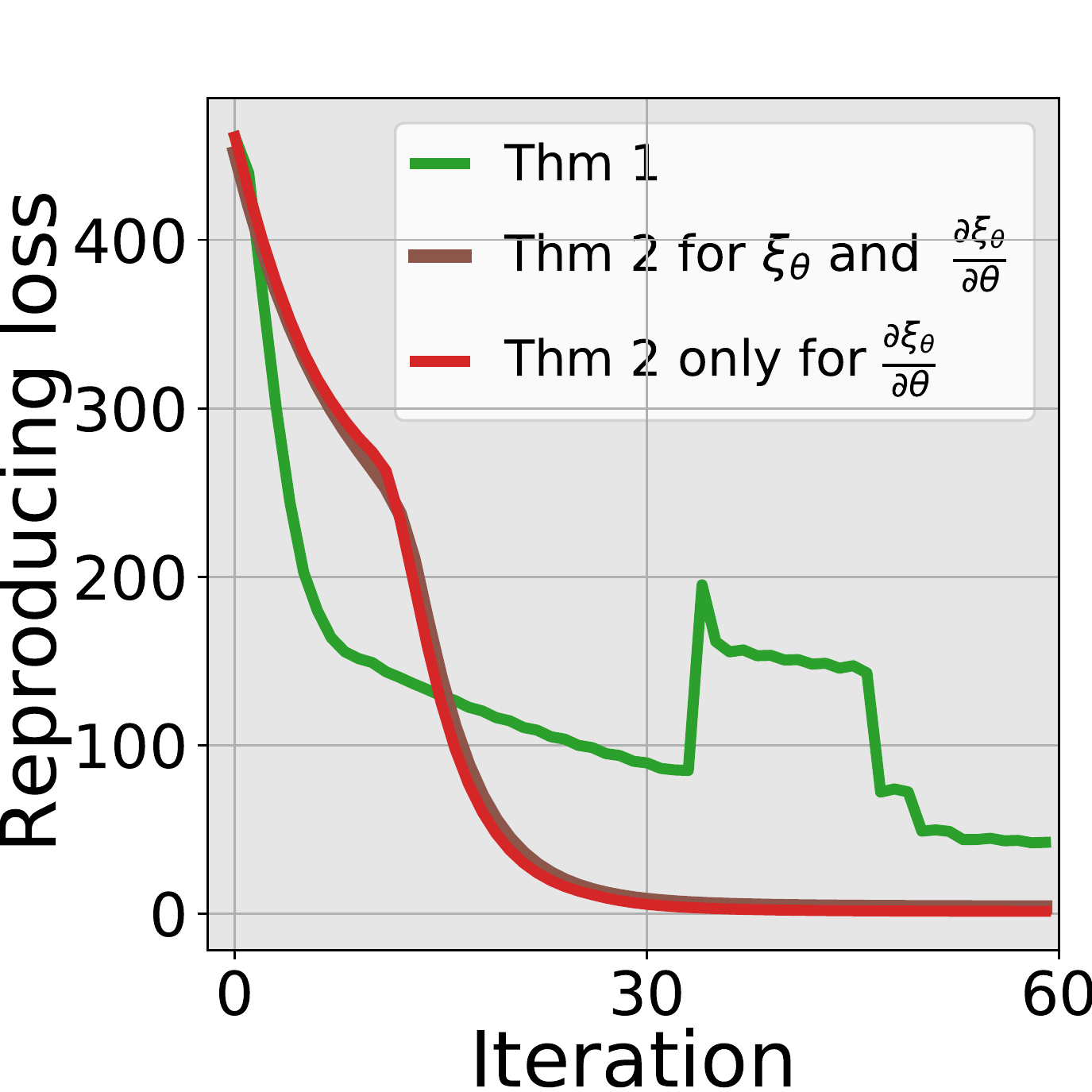}
		\caption{Rocket}
		\label{figsioc2.4}
	\end{subfigure}
	\caption{Jointly learning dynamics,  constraints, and control cost from demonstrations.}
	\label{figsioc2}
\end{figure}

To solve Problem \ref{equ_problem}, since there is no task constraints, we use the vanilla gradient descent to minimize the reproducing loss  (\ref{loss_lfd}) while using    the three strategies as mentioned above to handle the lower-level Problem \ref{equ_traj}. The initial condition for the gradient descent is given randomly, and the learning rate for the gradient-descent is set as $10^{-5}$.
The complete results for all systems in Table \ref{experimenttable} are given in Fig. \ref{figsioc2} (also see Fig. \ref{figsioc.1}-\ref{figsioc.4} in the primary text of the paper).

Fig. \ref{figsioc2.1} - Fig. \ref{figsioc2.4} plot the reproducing loss (\ref{loss_lfd}) versus gradient-descent iteration. The results show that for  Strategies (B) and (C) (in blue and red, respectively), the reproducing loss (\ref{loss_lfd}) is quickly covering to zeros, indicating that the  dynamics, constraints, and control cost function are successfully learned to reproduce the demonstrations.  However, we also note that Strategy (A) (in green) suffers from some numerical instability, which will be discussed below.

We have provided  some videos for the above learning MPCs from demonstrations using Safe PDP. Please visit  the link  \url{https://youtu.be/OBiLYYlWi98}. The codes for all   experiments here can be downloaded at \url{https://github.com/wanxinjin/Safe-PDP}. \footnote{All  experiments in this paper have been performed on a personal computer with 3.5 GHz Dual-Core Intel Core i7 and macOS Big Sur system.}

\bigskip
\bigskip

\noindent
\textbf{Why implementation of Theorem \ref{theorem1} is not numerically stable?} \quad
In both Fig. \ref{figscode} and \ref{figsioc2},
we  have noted that Strategy (A) suffers from some numerical instability, and this is due to the following reasons. First,  Theorem \ref{theorem1}  requires to accurately detect the inactive/active inequalities (i.e., whether an inequality constraint is zero or not), which is always difficult accurately due to  computational error (in our experiments, we detect the active constraints by applying a brutal threshold, as described in Algorithm \ref{alg_theorem1}). Second, although  the  differentiability of $\boldsymbol{\xi}_{\boldsymbol{\theta}}$ holds at the local neighborhood of $\boldsymbol{\theta}$,  $\boldsymbol{\xi}_{\boldsymbol{\theta}}$ might be extremely discontinuous  due to the `jumping switch'  of the active and inactive inequality constraints for the large range of $\boldsymbol{\theta}$; thus, such  non-smoothness  will deteriorate the decrease of loss   between   iterations.

\bigskip

\textbf{Why implementation of Theorem \ref{theorem2} is more numerically stable?} \quad Theorem \ref{theorem2} has perfectly addressed the above numerical issues of Theorem \ref{theorem1}. Specifically, first,  there is no need to distinguish the active and inactive inequality constraints in Theorem \ref{theorem2}; and second, in Theorem \ref{theorem2}, by adding all constraints to the control cost function, it introduces the `softness' of the hard constraints  and potentially eliminates the discontinuous `jumping switch' between  inactive and active inequalities over a large range of $\boldsymbol{\theta}$, enabling a stable decrease of loss when applying gradient descent.
\bigskip

\newpage
\section{Further Discussion} \label{appendix.discussion}
In this section, we will provide further experiments and discussion on the performance of Safe PDP.

\subsection{Comparison Between Safe PDP and PDP}\label{appendix.discussion.pdpcompare}
In this part, we compare Safe PDP and non-safe PDP \cite{jin2019pontryagin} to show the  performance trade-offs between the constraint enforcement of Safe PDP and its resulting computational expense. We use the example of learning MPCs from expert demonstrations for the cartpole system (in Table \ref{experimenttable}) to show this, and the experiment settings are the same with Appendix \ref{appendix.application.mpc}.  The comparison results between Safe PDP and PDP are given in the following Table \ref{discussion.table.pdp}. 

\begin{table}[h]
	\caption{Performance comparison between Safe PDP and PDP}
	\centering
	{%
	\begin{tabular}{@{}cccccc@{}}
		\toprule 
		 \thead{Methods} & \thead{Loss at convergence}  & \thead{Timing for \\ Forward Pass} & \thead{Timing for \\ Backward Pass} & \thead{Learning \\ constraints?} & \thead{Constraint \\ Guaranteed?} \\ 
		\midrule
		PDP      &  524.02                    
		&        \textcolor{red}{0.10s }             &      0.046s                   &           No         &     No                  \\[4pt]
		Safe PDP &  \textcolor{red}{7.42}                  
		&          0.21s              &    0.042s                     &        \textcolor{red}{Yes}            &      \textcolor{red}{Yes}                   \\ 
		\bottomrule
	\end{tabular}
	\label{discussion.table.pdp}
}
\end{table}

Based on the  results in Table \ref{discussion.table.pdp}, we have the following comments and analysis.

(1) We note that Safe PDP achieves lower training loss. This is because compared to PDP, Safe PDP has
	introduced the inductive bias of constraints within its model architecture, making it more suited to learn from
	demonstrations which are the results of a constrained control system (expert). In this sense, Safe PDP architecture
	(with an inductive bias of constraints) can be thought of as having more expressive power than PDP architecture for the above
	experiments.
	
(2)
	For Safe PDP, its ability to learn and guarantee constraints  comes at the cost of lower computational
	efficiency in the forward pass, as shown in  the second column in Table \ref{discussion.table.pdp}. Even though Safe PDP handles constraint enforcement
	by adding them to the control cost using  barrier functions, solving the resulting unconstrained approximation still
	needs more time than solving the unconstrained PDP. This could be because the added log barrier terms can increase the complex/stiff curvature of
	the cost/loss landscape, thus taking longer to find the minimizer. Further discussion about how barrier parameter influences the computational efficiency of the forward pass will be given in Appendix \ref{appendix.discussion.tradeoff}.
	
(3) 
The running time for the backward pass is almost the same for both PDP and Safe PDP because both methods are
	solving an unconstrained LQR   problem  (auxiliary control system) of the same size (see Theorem \ref{theorem2}), which can be very efficient based
	on Riccati equation.

\subsection{Strategies to Accelerate Forward Pass of Safe PDP} \label{appendix.discussion.acceleration}

In the previous  experiments in Appendix \ref{appendix.experimentdetails}, we have used an NLP solver to solve the trajectory optimization (optimal control) in the forward pass. Since
the solver blindly treats an optimal control problem as a general non-linear program without leveraging the
(sparse) structures  in an optimal control problem. Thus, solving the long-horizon
trajectory optimization  is not very  efficient. To accelerate  long-horizon trajectory optimization, one can use  plenty of strategies, as described below.

\begin{itemize}
	\item To solve a long-horizon optimal control problem, one effective method is to scale a 
	(continuous) long time horizon into a smaller one (like a unit) by applying a  time-warping function to 
	the system dynamics and control cost \cite{jin2020sparse}. After discretizing and solving this  short-horizon optimal
	control problem,  re-scale the obtained optimal trajectory back. This  time-scaling  strategy is common in many
	commercial optimal control solvers, such as GPOPS \cite{patterson2014gpops}.
	
	\item There are also the `warm-up’ tricks to accelerate the trajectory optimization in the forward pass of Safe PDP. For example, one
	can initialize the trajectory at the next iteration using the result of the previous iteration.
	
	\item One also can use a coarse-to-fine hierarchical strategy to solve  long-horizon trajectory optimization.
	For example, given a long-time horizon optimal control system, first, discretize the trajectory with larger
	granularity and solve for a coarse-resolution optimal trajectory; then use the coarse trajectory as initial conditions to
	solve the trajectory optimization with fine granular discretization.
	
\end{itemize}

As an additional experiment based on cartpole system (in Table \ref{experimenttable}), we tested and compared the above three strategies for accelerating the forward pass of
Safe PDP. The timing for each strategy is given in the following Table \ref{discussion.table.forwadpass}. Here, $t_f$ is the continuous-time horizon of the cartpole system, $\Delta$ is the discretization interval, and the discrete-time horizon is $T={t_f}/{\Delta}$.

\begin{table}[h]
	\caption{Running time for different strategies in accelerating the forward pass of Safe PDP}
	\centering
	\resizebox{\textwidth}{!}{%
		\begin{tabular}{@{}ccccc@{}}
			\toprule 
			Strategies  & \thead{$t_f=2$s, $\Delta=0.1$s,\\ $T=t_f/\Delta=20$}  &\thead{$t_f=6$s, $\Delta=0.1$s,\\ $T=t_f/\Delta=60$} &\thead{$t_f=10$s, $\Delta=0.1$s,\\ $T=t_f/\Delta=100$} &\thead{$t_f=20$s, $\Delta=0.1$s,\\ $T=t_f/\Delta=200$} \\ 
			\midrule 
			Plain NPL solver      &  0.082s                
			&        0.202s             &      0.491s                   &           1.743s               
			\\[4pt]
			Time scaling &  \textcolor{red}{0.014s}               
			&         \textcolor{red}{0.033s }            &    \textcolor{red}{0.055s}                     &        \textcolor{red}{0.083s}               
			\\ [4pt]
			Warm start &  0.055s                 
			&         0.095s           &    0.108s                    &        0.224s               
			\\ [4pt]
			Hierarchical &  0.021s             
			&         0.055s             &    0.074s                    &       0.133s                 
			\\ 
			\bottomrule
		\end{tabular}
		\label{discussion.table.forwadpass}
	}
\end{table}

From the results in Table \ref{discussion.table.forwadpass}, one can see that time-scaling is the most effective way among others to accelerate long-horizon trajectory
optimization. Of course, one can combine some of the above strategies to further
improve the running performance of the forward pass of Safe PDP.

Additionally, one can also use iLQR \cite{li2004iterative} and DDP \cite{jacobson1970differential} to solve optimal control problems. iLQR can be viewed as the one-and-half-order method---linearizing dynamics and quadratizing cost function. DDP is a second-order method ---
quadractizing both dynamics and cost function. Both methods solve a local bellman equation to generate the
update of the control sequence. But without coding optimization, both methods are slower than the commercial
NPL solver (e.g., CasADi \cite{andersson2019casadi}). Some ongoing works are trying to take advantage of GPUs for accelerating
trajectory optimization, which is also our future research.

\subsection{Trade-off Between Accuracy and  Efficiency using Barrier Penalties}\label{appendix.discussion.tradeoff}

In the paper, we have provided both theoretical  guarantees
(see Theorem \ref{theorem2} and Theorem \ref{theorem3}) and empirical experiments (see Fig. \ref{figthem}, Fig. \ref{figspo.1}
and \ref{figspo.3}, Fig. \ref{figsplan.1} and \ref{figsplan.3}, and Fig. \ref{figsioc}) for the relationship between the accuracy of a solution to an unconstrained approximation and the choice of the barrier parameter. This subsection further investigates  the trade-off between  accuracy and computational efficiency under   different choices of the   barrier parameter. 

In the experiment below (based on the cartpole system in Table \ref{experimenttable}), by choosing different  barrier parameters $\gamma$ in the forward pass of Safe PDP, we show the accuracy of the  trajectory $\boldsymbol{\xi}(\gamma)$ solved from an unconstrained approximation system $\boldsymbol{\Sigma}(\gamma)$ and the corresponding computation time. The results are presented in Table \ref{discussion.table.tradeoff}. 
\begin{table}[h]
	\centering
	\caption{Accuracy of the trajectory $\boldsymbol{\xi}(\gamma)$ from the unconstrained approximation system $\boldsymbol{\Sigma}(\gamma)$  and its computation time with different choices of
		barrier parameter $\gamma$}
	{%
		
		\begin{tabular}{@{}ccccccc@{}}
			\toprule
			 & \multicolumn{6}{c}{choice of $\gamma$}
			 \\
			 \cmidrule(lr){2-7}
			& $1$ & $10^{-1}$ &$10^{-2}$ &$10^{-3}$ &$10^{-4}$ &$10^{-5}$ \\
			\midrule
			\thead{Accuracy of $\boldsymbol{\xi}(\gamma)$ in percentage: \\ 
			$\frac{\norm{\boldsymbol{\xi}(\gamma)-\boldsymbol{\xi}^*}_2}{\norm{\boldsymbol{\xi}^*}_2} \times 100\%$ \textcolor{red}{$^1$ }} & 51.9\% &12.2\% & 1.6\%& 0.18\%& 0.018\% &0.0002\% \\[8pt]
		    Timing for computing $\boldsymbol{\xi}(\gamma)$ &0.023s & 0.033s &0.035s& 0.040s &0.038s& 0.047s \\ 
			\bottomrule
		\end{tabular}
	}
		\label{discussion.table.tradeoff}
		
		\footnotesize{\textcolor{red}{$^1$ }Note that in the above table, $\boldsymbol{\xi}^*$ is the ground-truth solution obtained from solving the original constrained
			trajectory optimization, and the computation time for such a constrained trajectory optimization is 0.062s.}
\end{table}

We have the following comments on the above results in Table \ref{discussion.table.tradeoff}.

\begin{itemize}
	\item First, the results in the first row of  Table \ref{discussion.table.tradeoff} show that a smaller barrier parameter leads to higher accuracy of
	the approximation solution. This again confirms the  theoretical guarantee in Theorem \ref{theorem2}
	(Claim (b)). The results here are also consistent with the ones in Fig. \ref{figthem} in the paper.
	
	\item 
	Second, the second row of Table \ref{discussion.table.tradeoff} shows that a smaller barrier parameter, however,  increases the
	 computation time for solving the unconstrained approximation  optimization. This
	could be because using a small barrier parameter, the added barrier terms can increase the
	complex/stiff curvature of the cost/loss landscape, thus taking Safe PDP longer to find the
	minimizer. Despite this, the time needed for finding a minimizer is still lower than
	directly solving a constrained trajectory optimization in the above experiment. 
	
	\item 
	Third, if one still wants to further increase the computation efficiency of Safe PDP, we have
	provided some strategies to achieve so, including "time scaling," "warm start," and "coarse-to-fine." Please check the Appendix  \ref{appendix.discussion.acceleration}  for more detailed
	descriptions and corresponding experiment results.
\end{itemize}

In summary, we have shown that  higher accuracy of the unconstrained approximation solution can be achieved using
a smaller barrier parameter, while a smaller barrier parameter would increase the computation
time for finding the approximation solution. In practice, one would  likely  choose an appropriate barrier
parameter to balance the trade-off between accuracy and computational efficiency. Also, there are multiple strategies available to increase
the computational efficiency of Safe PDP, as discussed in the Appendix  \ref{appendix.discussion.acceleration}.

\subsection{Learning MPCs from Non-Optimal Demonstrations}\label{appendix.discussion.nonoptimal}
In the application of learning MPCs (including objective, dynamics, constraints), given non-optimal demonstrations, Safe PDP can still learn an MPC   such that the trajectory reproduced by the learned MPC  has the closest discrepancy to the given non-optimal demonstrations (e.g., when  the task loss is defined as $l_2$ norm between the reproduced trajectory and demonstrations). As an illustrative example, the following Table \ref{discussion.table.mpc} shows  learning an  MPC
from a sub-optimal demonstration for the cartpole system.

\begin{table}[h]
	\caption{Safe PDP for learning MPCs from non-optimal demonstrations}
	\resizebox{\textwidth}{!}{%
		\begin{tabular}{@{}ccccccccc@{}}
			\toprule
			& \multicolumn{8}{c}{Number of iterations}
			\\
			\cmidrule(lr){2-9}
			& $0$ & $10$ &$20$ &$50$ &$100$ &$150$ &$200$ &$1000$ \\
			\midrule
			loss with optimal demo &779.986  &2.206  &1.481 &0.832 &0.641 &0.620 &0.611 &0.232 \\[8pt]
	loss with non-optimal demo& 1126.820 &18.975 &17.771 &15.602 &13.690 &12.469 &11.620 &10.923 \\ 
			\bottomrule
		\end{tabular}
	}
	\label{discussion.table.mpc}
\end{table}

As shown  in \ref{discussion.table.mpc}, the only difference between learning from  optimal and non-optimal demonstrations is that the converged
loss for the non-optimal demonstrations is relatively higher than for the optimal ones. This is because, for non-optimal
demonstrations, there might not necessarily exist an exact MPC model in the parameterized model space which perfectly
corresponds to the given demonstration. In such a case, however, Safe PDP can still find the best model in the parametrized model
space such that its reproduced trajectory has a minimal distance to the given non-optimal  demonstrations. For the extended research of the generalization ability of the learned MPCs from the non-optimal demonstrations, please refer to \cite{jin2020sparse}.

\subsection{Limitation of Safe PDP}\label{appendix.discussion.limitation}

Safe PDP requires a safe (feasible) initialization such that the log-barrier-based objectives (cost or task)  are well-defined. While this requirement can be restrictive in some cases, we have the following empirical experiences on how to provide safe initialization for different types of problems.

\begin{itemize}
	\item In safe policy optimization, one could first use supervised learning to learn a safe policy from some safe trajectories/demonstrations (which could not necessarily be optimal). Then, use the learned safe policy to initialize  Safe PDP. We have used this strategy in our previous experiments in Appendix \ref{appendix.application.spo}.
	\item In safe motion planning, one could arbitrarily provide a safe trajectory (not necessarily optimal) to initialize  Safe PDP.  We have used this strategy in the previous experiments in Appendix \ref{appendix.application.splan}.
	\item In learning MPCs from demonstrations (Appendix \ref{appendix.application.mpc}), the goal includes learning  constraint models, and there is no such requirement.
\end{itemize}

Also, Safe PDP cannot apply to robust  learning and control tasks. The goal of robust learning and  control concerns   achieving or maintaining good performance (such as stability or optimality) in the  case of the worst disturbance or attacks to a system. Methods for handling those types of problems, such as robust control and differential game, have been well-developed  in both  control and machine learning communities. On the other hand, Safe PDP only focuses on  guaranteeing  the satisfaction of inequality constraints throughout a learning or control process, and such constraints are defined on the system states and inputs.

\end{document}